\documentclass[11pt]{article}
\usepackage[
left = 0.55in,
right=0.5in,
headheight=15pt
]{geometry}
\usepackage[numbers,sort&compress]{natbib}
\usepackage{tikz}
\usepackage{adjustbox}
\usetikzlibrary{arrows.meta,positioning,fit,backgrounds,calc}
\usepackage{amsfonts, amsmath, amssymb,amsthm}
\usepackage{mathtools}
\usepackage[mathscr]{eucal}
\usepackage{hyperref}
\usepackage[nameinlink]{cleveref}
\usepackage{aliascnt}
\usepackage{authblk}
\usepackage{graphicx}
\usepackage{subcaption}
\usepackage{placeins}
\usepackage{etoolbox}
\usepackage{caption}
\usepackage{booktabs}
\usepackage{float}
\usepackage{bbm}
\usepackage{longtable}
\usepackage{booktabs}
\usepackage{multirow}
\usepackage{array}
\usepackage{lastpage}
\usepackage{bm}
\newtheorem{remark}{Remark}
\crefname{remark}{remark}{remarks}
\Crefname{remark}{Remark}{Remark}

\makeatletter
\pretocmd{\subsection}{\FloatBarrier}{}{}
\pretocmd{\subsubsection}{\FloatBarrier}{}{}
\makeatother

\newcommand{\err}[2]{$#1 \pm #2$}
\newcommand{\besterr}[2]{$\mathbf{#1} \pm #2$}

\DeclareMathOperator{\MLP}{MLP}

\newcommand{\ind}[1]{\mathbb{1}_{#1}}

\newcommand{\Def}{\overset{\textbf{def}}{=}}

\newcommand{\eps}{\varepsilon}

\usepackage{amsopn}

\newcommand{\nnorm}[1]{{\left\vert\kern-0.25ex\left\vert\kern-0.25ex\left\vert #1 \right\vert\kern-0.25ex\right\vert\kern-0.25ex\right\vert}}

\newcommand{\timehor}{\mathbf{T}}

\DeclarePairedDelimiter{\abs}{\lvert}{\rvert}

\DeclarePairedDelimiter{\set}{\{}{\}}

\crefname{appendix}{appendix}{appendices}
\Crefname{appendix}{Appendix}{Appendices}

\title{HypNO: A Graph-Based Neural Operator with Physics-Informed
Message Passing for Hyperbolic Conservation Laws}

\author{
Dimitrije Ždrale\thanks{\texttt{dimitrije.zdrale@universite-paris-saclay.fr}}$^{1}$,
Cassie An Jeng\thanks{\texttt{cassie.jeng@berkeley.edu}}$^{2}$,
Katie Wang\thanks{\texttt{katiewang@berkeley.edu}}$^{2}$,
Sonia Vanier\thanks{\texttt{sonia.vanier@polytechnique.edu}}$^{1}$,
Alexandre Bayen\thanks{\texttt{bayen@berkeley.edu}}$^{2}$, \\
Hossein Nick Zinat Matin\thanks{Corresponding author: \texttt{hossein.matin@polytechnique.edu}}$^{1}$
}

\date{}

\begin{document}
\maketitle
\begin{center}
\small
$^{1}$ Laboratory of Computer Science (LIX), École Polytechnique,
Palaiseau, France

\vspace{0.5em}

$^{2}$ Department of Electrical Engineering and Computer Sciences,
University of California, Berkeley, CA, USA
\end{center}

\vspace{1em}

\begin{abstract}
We introduce HypNO, a graph-based neural operator for scalar hyperbolic conservation laws. HypNO operates directly on a space-time graph of finite-volume cells and uses adjacency-factored, physics-informed message passing to respect upwinding and entropy admissibility near shocks. We benchmark the architecture on the Lighthill-Whitham-Richards (LWR) and Aw-Rascle-Zhang (ARZ) traffic-flow models, a stress test for operator-learning methods because of their simultaneous global transport and shock formation. HypNO predicts solution snapshots accurately across a range of initial conditions while capturing the shocks and discontinuities of the solution.
\end{abstract}

\bigskip



\section{Introduction}
\label{sec:introduction}

Partial differential equations (PDEs) are the basic language of continuum physics and engineering, describing phenomena as diverse as fluid dynamics, electromagnetics, climate, and traffic flow~\citep{karniadakis2021physics}.
Outside a handful of idealized geometries and linear regimes, however, these equations rarely admit closed-form solutions; in the general nonlinear case the solution must instead be approximated by numerical computation~\citep{leveque2002fv,toro2009riemann}.

For this reason, numerical methods have historically been the practical route to solving PDEs.
Finite difference, finite volume, and finite element methods are the standard tools for numerical computations~\citep{leveque2002fv,toro2009riemann}, and for problems with discontinuous solutions, high-order, essentially non-oscillatory schemes such as WENO provide stable, high-resolution approximations~\citep{shu2009weno}.
Numerical solvers are reliable, interpretable, and backed by a mature theory of stability and convergence.
Their cost, however, scales with problem complexity; each new initial condition or parameter setting typically requires a full sequential time integration on a sufficiently resolved grid.

Machine learning has more recently emerged as a complementary (sometimes alternative) route for numerical computations~\citep{karniadakis2021physics,raissi2019pinns}.
Rather than replacing numerical analysis, learned models can amortize computation across families of related inputs, offering fast inference after an upfront training phase.
This is especially attractive for parametric studies, design loops, and real-time prediction, where many instances of the same governing equation must be solved under varying data.

The present work studies \emph{operator learning} for nonlinear hyperbolic conservation laws: learning a map from initial data $u_0(x)$ to the full space-time solution $u(x,t)$.
We propose \textbf{HypNO} (\emph{Hyperbolic Neural Operator}), a graph-based architecture that replaces the global spectral mixing of Fourier operators with causal space-time message passing on a cell graph.
Messages are conditioned on finite volume interface features (flux, characteristic speed, and upwind direction), so the model is biased toward transport along characteristics rather than isotropic local smoothing.
We evaluate HypNO on the scalar Lighthill--Whitham--Richards (LWR) traffic model~\citep{lighthill1955lwr,richards1956traffic} and the Aw--Rascle--Zhang (ARZ) system~\citep{aw2000arz}, benchmarking against FNO, a state-of-the-art neural operator, and against established numerical solvers (WENO, Godunov, and HLL) on shock-dominated initial data.

Our contribution, stated briefly, is a one-shot hyperbolic neural operator that lifts an initial condition onto a causal space-time graph, passes physics-gated messages with finite-volume edge features, and maps directly to the full solution field.
Section~\ref{sec:related_work} contrasts HypNO with numerically structured, physics-informed, operator-based, and generic graph-based learned solvers; Sections~\ref{sec:scheme}--\ref{sec:arz-mark0-eval} develop the architecture and report stratified LWR and ARZ benchmarks; Section~\ref{sec:conclusion} summarizes limitations and future directions.

\section{Background and Related Work}
\label{sec:related_work}

\subsection{Hyperbolic PDEs}
\label{sec:hyperbolic_pdes}

Hyperbolic PDEs and conservation laws describe quantities that are transported but not created or destroyed~\citep{leveque2002fv,toro2009riemann}.
Information propagates along characteristics at speeds determined by the local state, allowing different parts of a profile can move at different rates.
When the flux is nonlinear, characteristics may converge and cross, producing shocks, or spread apart, forming rarefactions; discontinuities can therefore appear in finite time even from smooth initial data~\citep{leveque2002fv}.
Systems such as traffic flow add further structure: multiple wave families can interact, and contact discontinuities may persist alongside genuinely nonlinear shocks~\citep{aw2000arz}.

Classical finite volume and upwind finite difference methods are built around this transport structure~\citep{toro2009riemann}.
Godunov-type fluxes respect upwind information while higher-order reconstructions such as WENO raise accuracy in smooth regions while limiting spurious oscillations near discontinuities~\citep{shu2009weno}.
These schemes remain the reference standard for shock-dominated problems, but their cost scales with the number of time steps and spatial degrees of freedom required for each new instance.

\subsection{Learning-based PDE solvers}
\label{sec:learning_pde}

Recent work on machine learning for PDEs can be organized into four broad classes, distinguished by the component they replace or augment in the numerical solver and by their inference-time computational cost.

\paragraph{Numerically structured schemes.}
The first class embeds neural networks inside time-stepping procedures that resemble classical solvers.
FluxGNN~\citep{horie2024fluxgnn} learns numerical fluxes on mesh interfaces while enforcing conservation and similarity equivariance.
First-order hyperbolic conservation laws of Baba et al.~\citep{baba2026hyperbolic} and the neural finite volume framework of Lichtl\'e et al.~\citep{lichtle2026nfv} replace the classical numerical flux in a finite volume scheme with a trainable network while preserving the conservative update structure, with both supervised and weak-form unsupervised training objectives.
Hybrid FEM--NN models~\citep{mitusch2021hybridfemnn} discretize the PDE in space with classical finite elements and embed neural networks to represent unknown coefficients, operators, or constitutive relations, enforcing the governing equation as a hard constraint in the optimization rather than a soft penalty.
LRNN--DG~\citep{sundong2024lrnn} uses the output fields of the last hidden layer on each subdomain as randomized local basis functions and couples them through a discontinuous-Galerkin weak formulation, with only the output weights obtained by least squares.
These approaches are interpretable and can inherit stability mechanisms from numerical analysis, but they generally advance the solution sequentially in time, so inference complexity remains comparable to conventional solvers.

\paragraph{Physics-informed neural networks (PINNs).}
The second class trains a neural network to represent the solution field itself, penalizing violations of the governing equation, boundary conditions, and initial data~\citep{raissi2019pinns,karniadakis2021physics,sirignano2018dgm}.
The Deep Galerkin Method~\citep{sirignano2018dgm} is an early mesh-free representative: a deep network approximates the solution globally and is trained on batches of randomly sampled space-time points to minimize the PDE residual together with boundary and initial data.
Variational and weak-form variants replace collocation on strong-form residuals with integration against test functions~\citep{kharazmi2019vpinn}.
Training difficulties are now well documented: Wang, Yu, and Perdikaris~\citep{wang2022pinnsfail} analyze failure modes of PINNs through neural tangent kernel theory, showing mismatched convergence rates across loss components.
For hyperbolic conservation laws, standard PINNs assume sufficient smoothness and can fail near shocks and contacts, where strong-form derivatives are not meaningful~\citep{deryck2024wpinns}.
Mishra and Molinaro~\citep{mishra2023generalization} bound the generalization error of PINNs for forward problems in terms of training error and quadrature, and De Ryck, Mishra, and Molinaro~\citep{deryck2024wpinns} introduce weak PINNs (wPINNs) that target entropy solutions through a min-max formulation based on Kru\v{z}kov entropies.
For hyperbolic systems, Patel et al.~\citep{patel2022tcpinn} adopt a space-time control-volume discretization that enforces conservation and thermodynamic consistency more directly than collocation-based PINNs.
These developments show that physics-informed learning can be made rigorous, but one-shot inference across parametric families of initial data remains outside the usual PINN setup.

\paragraph{Operator-based schemes.}
The third class learns mappings between infinite-dimensional function spaces directly so that a trained model transfers across discretizations rather than being tied to a fixed grid~\citep{kovachki2021neuraloperator}.
\textit{DeepONet}~\citep{lu2021deeponet} encodes the input function and the query coordinates $(x,t)$ through separate branch and trunk networks whose inner product defines the predicted field.
The \textit{Fourier Neural Operator} (FNO)~\citep{li2021fno} lifts the input to a latent space and applies global convolutions in Fourier space, giving fast global mixing and resolution-invariant parameter sharing; its physics-informed extension, PINO~\citep{li2023pino}, adds a PDE-residual regularizer.
Related architectures include multipole graph operators~\citep{li2020mgno}, spatio-spectral graph operators~\citep{sarkar2025sp2gno}, and deeper U-shaped operator networks~\citep{rahman2023uno}.
After training, these models predict full space-time fields for a new initial condition in a single forward pass, which is attractive for parametric studies and real-time use~\citep{karniadakis2021physics}.
Operator learning has been most successful on relatively smooth PDE families; for hyperbolic problems, global spectral mixing is biased toward smooth functions and tends to blur sharp fronts~\citep{khodakarami2026spectralbias}.
Hybrid flux-learning schemes within the FNO framework~\citep{kim2024fluxfno} and conservation-encoded operators~\citep{liu2024clawno} have started to address this gap, although shock-forming scalar laws remain underexplored.

\paragraph{Graph-based solvers.}
The fourth class replaces both classical time-stepping and global spectral mixing with message passing on a graph~\citep{anandkumar2020gkn,brandstetter2022mpde,pfaff2021meshgraphnets,horie2022penn}. Graph neural networks (GNNs) iteratively refine node representations by aggregating learned messages from neighboring nodes and combining them with each node’s existing features through a learned update mechanism. The graph-kernel neural operator~\citep{anandkumar2020gkn} formulates operator learning as message passing on a domain discretization. Physics-embedded networks~\citep{horie2022penn} embed boundary conditions and implicit time stepping into an equivariant GNN. MeshGraphNets~\citep{pfaff2021meshgraphnets} and message-passing PDE solvers~\citep{brandstetter2022mpde} learn local interactions on unstructured meshes and across resolutions.
Compared with spectral operators, message passing emphasizes neighborhood structure over global mixing.
The critical design question is therefore what information the messages should carry.
Most existing GNN-based PDE solvers rely on generic edge features (coordinate differences, state jumps, or latent inner products) and do not explicitly encode flux, characteristic speed, upwind direction, or Rankine--Hugoniot structure.
For hyperbolic problems, this is a fundamental limitation: the relevant stencil is not symmetric and the direction of information flow depends on the solution itself.
FluxGNN~\citep{horie2024fluxgnn} moves toward conservation-aware flux learning but retains sequential time marching, and one-shot graph operators with explicit hyperbolic edge structure have not yet been established.

\paragraph{Our contribution.}
HypNO combines operator learning with physics-gated message passing on a causal space-time cell graph (\Cref{sec:scheme}).
After training, a single forward pass maps an initial condition to the full space-time field, so evaluation complexity scales with network depth rather than with the number of time steps required by a classical or hybrid integrator.
This is the central operational advantage over numerically structured learned solvers~\citep{horie2024fluxgnn,baba2026hyperbolic,lichtle2026nfv,brandstetter2022mpde}, which embed networks inside sequential time-stepping loops.

Relative to physics-informed networks~\citep{raissi2019pinns,deryck2024wpinns,patel2022tcpinn}, HypNO does not minimize strong-form residuals at collocation points, where derivatives fail at shocks and contacts.
Instead, it learns the solution operator directly and biases predictions through finite-volume interface quantities (flux, characteristic speed, Rankine--Hugoniot speed, upwind direction, and entropy- and CFL-based gates) computed from decoded physical probes at each layer.
On stratified LWR and ARZ benchmarks with exact or wave-front-tracking ground truth (\Cref{sec:arz-mark0-eval}), HypNO attains substantially lower density error on discontinuity-dominated initial data than PINN-type approaches would permit and than global-operator baselines achieve in practice.

Relative to operator-based methods such as FNO and DeepONet~\citep{li2021fno,lu2021deeponet}, which apply global spectral mixing, HypNO replaces isotropic long-range coupling with causal, characteristic-aware message passing.
Adjacent edges carry interface quantities that mirror finite-volume stencils; upwind, entropy, and temporal gates suppress entropy-violating and out-of-cone messages rather than relying on a smooth global basis to represent sharp fronts~\citep{khodakarami2026spectralbias,kim2024fluxfno}.
The architecture is therefore physics-aware by construction: messages encode the same local transport information that classical schemes use, composed into a learned operator that generalizes across resolutions and initial-condition families.

Relative to generic graph-based solvers~\citep{pfaff2021meshgraphnets,brandstetter2022mpde,horie2022penn}, HypNO specifies \emph{what} should be communicated along edges: not only coordinates or state jumps, but fluxes, eigenvalues, upwind flags, and CFL-like ratios; it also sizes the stencil so that the receptive field covers the hyperbolic domain of dependence (\Cref{app:reach}).
By design, the model preserves the structural ingredients of entropy-solution evolution: causal time marching, selective updates near discontinuities, and distinct treatment of upstream versus downstream neighbors.
Numerical experiments confirm that this combination yields dominant accuracy: on LWR, HypNO beats WENO5, Godunov, and FNO at every in- and out-of-distribution segment count (\Cref{tab:paper_eval_by_ic}); on ARZ, it reduces pooled density MAE by three- to fourfold over every classical and learned baseline while retaining a tighter error spread (\Cref{sec:arz-mark0-eval}).

\section{Mathematical Modeling}
A crucial component of hyperbolic PDEs is the solution to the Riemann problem. Riemann problems are defined by initial conditions with a single jump discontinuity, and produce discontinuities that propagate through the solution. Riemann problems enable solutions to more complex piecewise-constant initial conditions. Generic complex initial conditions of this form appear like sequential Riemann problems at different jump discontinuity locations $x$. Iteratively solving the Riemann problems as discontinuities propagate and collide throughout the solution, using methods like \textit{wavefront tracking}, is one numerical approach to solving these higher complexity initial conditions. However, in the numerical formulation of the problem, important choices, such as the choice of $\Delta t$ and $\Delta x$, impact stability of the final solution and how well the solution models the exact behavior of the system. Laws like the \textit{Courant-Friedrichs-Lewy} (CFL) stability condition govern choices of these parameters to ensure physically feasible and stable final solutions (\cite{front2015holden}).

The existence of discontinuities causes the solutions of hyperbolic PDEs to be interpreted in the weak sense. The entropy condition specifies the physically-relevant weak solution among the set of mathematically admissible weak solutions, ensuring the correct propagation of information through the system. The propagation of information in hyperbolic PDE systems is described by characteristic curves, and these characteristics and their speeds determine the formulation of shocks and rarefactions in scalar conservation laws, and shocks, rarefactions, and contact discontinuities in systems of conservation laws.


\subsection{Scalar conservation laws} 
In the general form, the scalar conservation laws equation can be presented by 
\begin{equation}\label{E:main-conservation}
\begin{split}
    \partial_t U(\bm x,t) + \operatorname{div}_{\bm x} f(\bm x, t,U(\bm x,t))&  = 0,
    \qquad \bm x \in \mathbb{R}^n,\quad t \in [0,T], \\
    U(\bm x, 0) & = U_0(\bm x). 
\end{split}
\end{equation}
Here, \(U(\bm x,t)\) denotes the conserved quantity, and \(f( \bm x, t, U)\) is the flux function. 

In this paper, our main goal is to introduce a Graph-based neural operator which can preserve the structure and physical properties of the model, and in particular, propagation of discontinuities. Therefore, we will focus on the two main subclasses of such hyperbolic PDEs: first-order and second-order conservation laws. In theory, the architecture would be adapted for any other type of PDEs. 

\subsubsection{LWR Equation} \label{sec:lwr}
For evaluation of our proposed HypNO architecture, we use the \textit{Lighthill-Whitham-Richards} (LWR) traffic flow model. This model describes traffic flow using fluid dynamics and treats the collection of vehicles like a compressible fluid rather than discrete entities. The first-order scheme offers simplicity and an exact numerical solution, allowing it to be widely used in fluid approximations of traffic, commonly used in transportation. It combines global transport, local steepening, and discontinuities, and exposes whether a model can generalize from smooth training signals to shock-dominated regimes, therefore serving as a good benchmark test for learning-based solvers. 

The LWR model describes the conservation of density $U(x, t) = \rho(x,t)$ (see \Cref{E:main-conservation}) as:
\begin{equation}
\partial_t \rho + \partial_x f(\rho) = 0,
\end{equation}
with a concave flux function. There are several options for defining the flux function $f$ in the literature from a practical point of view. In this work, we consider the Greenshields flux, defined as:
\begin{equation}
f(\rho) = \rho(1-\rho),
\end{equation}
which implies a density-dependent wave speed $f'(\rho)=1-2\rho$. The nonlinearity of $f(\rho)$ causes characteristics to intersect and shocks to form even from smooth initial conditions. 

\paragraph{Rankine-Hugoniot Condition} The propagation speed of information depends directly on the local density, with higher densities moving more slowly. When characteristics intersect, weak solutions develop discontinuities with propagation speed governed by the Rankine-Hugoniot (R-H) condition. For a shock, separating left and right density states $\rho_L$ and $\rho_R$, the shock speed $s$ is given by
\[
s = \frac{f(\rho_R) - f(\rho_L)}{\rho_R - \rho_L}
\]

\paragraph{Riemann problem}
For the scalar LWR model, the Riemann problem has two initial densities $\rho_0 = (\rho_L,\rho_R)$. Solving this fundamental problem with piecewise constant initial conditions produces shocks ($\rho_L < \rho_R$), with a slope determined by the R-H condition, and rarefactions ($\rho_L > \rho_R$).

\begin{figure}
    \centering
    \includegraphics[scale=0.4]{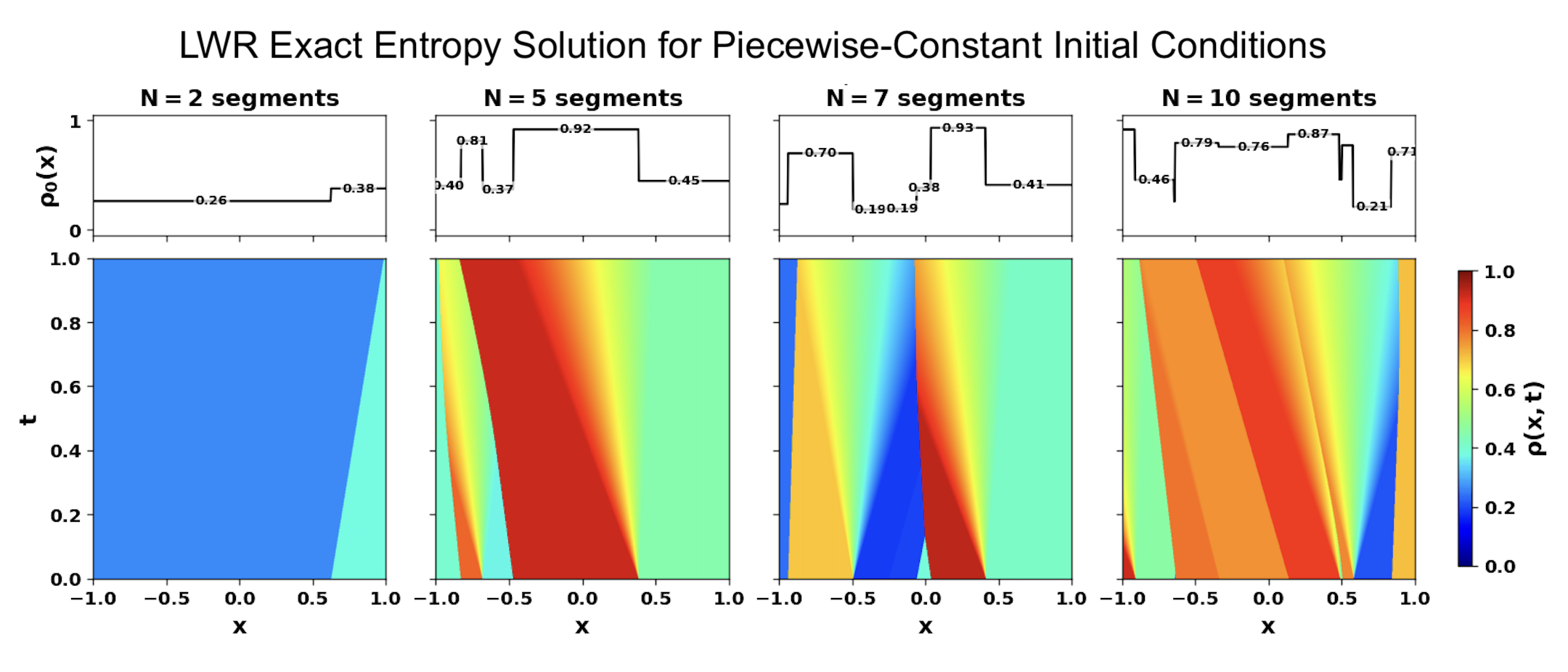}
    \caption{Exact LWR solutions for piecewise-constant initial data of increasing complexity. Each column corresponds to a different number of initial segments
  $N\in\{2,5,7,10\}$. The top row shows the initial density profile
  $\rho_0(x)$ with the constant value of each segment annotated; the row below
  shows the resulting space-time density field $\rho(x,t)$ from the exact
  Lax-Hopf entropy solution of
  $\partial_t\rho+\partial_x[\rho(1-\rho)]=0$ on $x\in[-1,1]$, $t\in[0,1]$.
  Shocks appear as sharp convergent interfaces and rarefactions as smooth fans;
  the density color scale is shared across all columns.}
    \label{fig:lwr_1d}
\end{figure}

\paragraph{Illustration of the solution}
Fig. \ref{fig:lwr_1d} shows example LWR solutions for $\rho(x,t)$ for piecewise-constant initial conditions with 2, 5, 7, and 10 initial segments, showing how these discontinuities evolve and collide in the numerical solution.

\subsubsection{ARZ equation} \label{sec:arz}
To evaluate the HypNO architecture's capability to generalize over systems of equations and more complex wave dynamics, we extend the HypNO network to support second-order ARZ-type systems of PDE equations, which describe both conservation of mass and conservation of momentum for out-of-equilibrium traffic flow. These systems introduce the concept of a pressure function to allow for modeling of driver anticipation and response to surrounding vehicle behaviors.

We use the \textit{Aw-Rascle-Zhang} (ARZ) model as defined in \cite{aw2000arz}. This model also describes the conservation of vehicle density $\rho(x,t)$, and here, the velocity $v(x,t)$ is an independent dynamical variable. 

In particular, letting 
\[
U = \begin{pmatrix}
    \rho \\ y  
\end{pmatrix}, \qquad y = \rho \omega = \rho v + \rho p(\rho), \quad f(\bm x, t, U) = \begin{pmatrix} \rho v \\ y v \end{pmatrix} 
\]
the conservation form of \Cref{E:main-conservation}, can be written in the explicit form as 
\[
\partial_t \rho + \partial_x(\rho v) = 0,
\qquad
\partial_t y + \partial_x(y v) =0,
\]
where $p(\rho) = \rho$ is the traffic pressure encoding drivers' anticipation of downstream conditions. Typically, pressure functions take the general form $p(\rho) = \rho^\gamma$, where $\gamma \geq 1$. In the HypNO evaluations, we take $\gamma = 1$.

Unlike the LWR model, the ARZ system carries two distinct wave families.
The genuinely nonlinear first field has characteristic speed
\begin{equation}
\label{eq:arz_eigenvalues}
\lambda_1 = v - \rho\,p'(\rho),
\end{equation}
and supports shocks and rarefactions; the linearly degenerate second field has speed $\lambda_2 = v$ and supports contact discontinuities.
With $p(\rho)=\rho$ we have $p'(\rho)=1$ and hence $\lambda_1 = v-\rho$.

\paragraph{Riemann invariants}
The ARZ system admits two Riemann invariants associated with these fields~\citep{aw2000arz}.
Along first-characteristic curves ($\lambda_1$), the quantity
\begin{equation}
\label{eq:arz_riemann_invariant_w}
\omega = v + p(\rho)
\end{equation}
is constant; this is the Lagrangian marker used throughout our model and conserved variable $y=\rho\omega$.
The genuinely nonlinear \emph{1-wave} (shock or rarefaction) preserves $\omega$ while $\rho$ and $v$ change.
Along second-characteristic curves ($\lambda_2=v$), the velocity $v$ itself is constant; the \emph{2-contact} therefore propagates at speed $v$ while density may jump across it.

\paragraph{Riemann problem}
For the ARZ system, the Riemann problem connects left and right primitive states,
\[
(\rho,v)(x,0) = \begin{cases}
    (\rho_L,v_L), & x < 0,\\
    (\rho_R,v_R), & x > 0,
\end{cases}
\]
with corresponding markers $\omega_L=v_L+p(\rho_L)$ and $\omega_R=v_R+p(\rho_R)$.
The entropy solution is a \emph{1-wave} (on $\lambda_1$, with $\omega$ preserved) followed by a \emph{2-contact} (on $\lambda_2$, with $v$ preserved).
Writing $v_L = \omega_L - p(\rho_L)$ and $v_R = \omega_R - p(\rho_R)$, the intermediate state satisfies $\omega_*=\omega_L$, $v_*=v_R$, and $\rho_*$ is determined implicitly by $p(\rho_*)=\omega_L-v_R$.
The \emph{1-wave} is a shock when first-characteristics converge into it,
$\lambda_1(U_L)>\lambda_1(U_*)$, and a rarefaction when they spread apart,
$\lambda_1(U_L)<\lambda_1(U_*)$.
The subsequent \emph{2-contact} has speed $v_R$ and separates the intermediate state from the right state; across it $v$ is continuous while $\rho$ and $\omega$ generally differ.
Unlike a scalar Riemann problem, the wave pattern therefore depends on both $(\rho,v)$ on each side, not on density alone.

\begin{figure}
    \centering
    \includegraphics[scale=0.4]{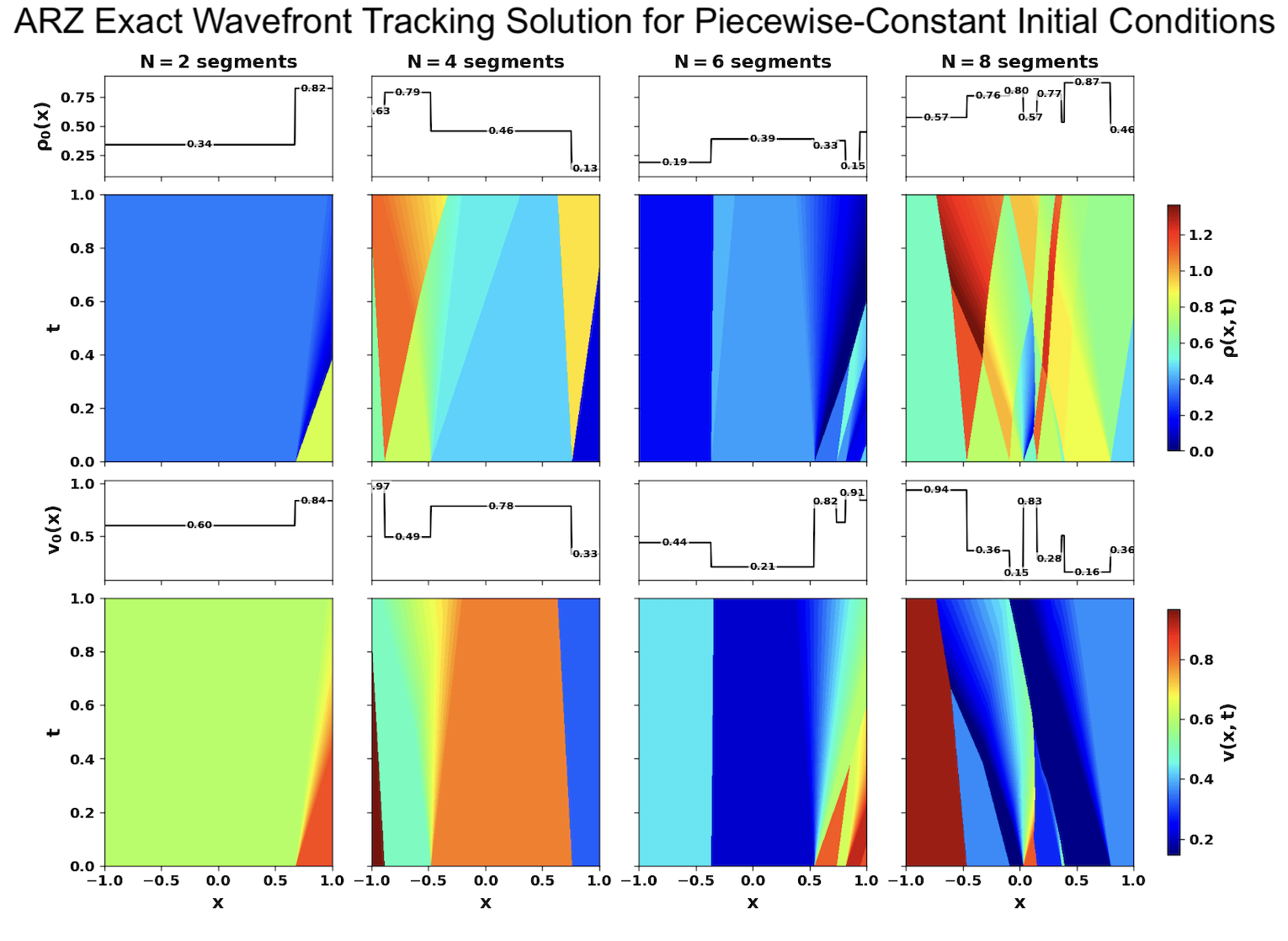}
    \caption{Exact ARZ solutions for piecewise-constant initial data of increasing
  complexity, with pressure $p(\rho)=\rho$. Each column corresponds to a
  different number of initial segments $N\in\{2,4,6,8\}$. For every column the
  two labeled rows give the initial density $\rho_0(x)$ and velocity $v_0(x)$
  profiles, with each segment's constant value annotated, above the
  corresponding space-time fields $\rho(x,t)$ and $v(x,t)$. Solutions are
  computed by wave-front tracking, which is exact to machine precision on shocks
  and contact discontinuities (the only approximation being the rarefaction-fan
  resolution). The genuinely nonlinear $1$-wave produces the shocks and
  rarefactions seen in both channels, while the linearly degenerate $2$-contact
  appears as a density jump across which $v$ stays continuous. Domain
  $x\in[-1,1]$, $t\in[0,1]$. The color scale is shared across columns within
  each field.}
    \label{fig:arz_ex}
\end{figure}

\paragraph{Illustration of the solution}
Fig. \ref{fig:arz_ex} shows example ARZ solutions for both $\rho(x,t)$ and $v(x,t)$ for piecewise-constant initial conditions with 2, 4, 6, and 8 initial segments, showing how these discontinuities evolve and collide in the numerical solution.




\section{HypNo scheme}
\label{sec:scheme}
GNNs provide a natural architecture for PDE surrogate modeling since their message-passing structure resembles the local update rules used in classical numerical schemes. In the finite volume type approaches, the state of a cell is updated through fluxes exchanged with neighboring cells. Similarly, in a message-passing layer, each node updates its latent state by aggregating information from neighboring nodes through learned edge functions. This makes GNNs particularly suitable when the dynamics are governed by local interactions, geometric structure, or mesh-dependent neighborhoods.

In addition, operator learning presents a promising direction for learning PDE representations, as it allows one to extend from a scheme that works for a single initial condition problem to a broader collection of more complex initial conditions. Learning the mappings between initial conditions (ICs) and their corresponding solutions is what makes neural operators desirable for various physical applications in which the ICs might dynamically change.

Thus, the proposed HypNO sets to combine these two methods into a message passing graph neural operator architecture, allowing us to learn a mapping between ICs and solutions of hyperbolic PDEs while capturing the irregular behavior of such PDEs more accurately.
\subsection{Network schematic}
We start by depicting the architecture of the proposed scheme in \Cref{fig:hypno-arch} The concepts and notations are elaborated in \Cref{app:design}.

\begin{figure}
    \centering
    \resizebox{1\textwidth}{!}{%

\def\Nlayers{L}

\begin{tikzpicture}[
  font=\small,
  >=Stealth,
  box/.style={
    rounded corners=2pt,
    draw=black!72,
    line width=0.6pt,
    fill=blue!7,
    align=center,
    inner sep=4pt,
    minimum height=12mm,
    minimum width=21mm
  },
  mpbox/.style={
    rounded corners=2pt,
    draw=black!72,
    line width=0.6pt,
    fill=blue!12,
    align=center,
    inner sep=4pt,
    minimum height=12mm,
    minimum width=21mm
  },
  widebox/.style={
    rounded corners=2pt,
    draw=black!72,
    line width=0.6pt,
    fill=blue!7,
    align=center,
    inner sep=4pt,
    minimum height=12mm,
    minimum width=30mm
  },
  gate/.style={
    rounded corners=2pt,
    draw=orange!75!black,
    fill=orange!13,
    line width=0.6pt,
    align=center,
    font=\scriptsize,
    inner sep=3pt,
    minimum height=8mm,
    minimum width=17mm
  },
  smallbox/.style={
    rounded corners=2pt,
    draw=black!60,
    line width=0.55pt,
    fill=blue!5,
    align=center,
    font=\scriptsize,
    inner sep=4pt,
    minimum height=11mm,
    minimum width=18mm
  },
  op/.style={
    circle,
    draw=black!70,
    line width=0.6pt,
    fill=white,
    inner sep=0pt,
    minimum size=6mm
  },
  thumb/.style={
    draw=black!60,
    line width=0.5pt,
    minimum size=10.5mm,
    inner sep=0pt
  },
  arr/.style={->, line width=0.9pt, draw=black!78, shorten >=2pt, shorten <=2pt},
  msg/.style={->, line width=0.75pt, draw=black!58, shorten >=2pt, shorten <=2pt},
  aux/.style={->, line width=0.7pt, draw=orange!75!black, dashed, shorten >=2pt, shorten <=2pt},
  lbl/.style={font=\scriptsize, text=black!65, align=center},
  panel/.style={font=\bfseries}
]

\path[use as bounding box] (-7.0,1.85) rectangle (7.0,-6.45);


\node[panel] at (0,1.30)
  {(a)\ Space--time operator: $u_0(x) \mapsto u(t,x)$};

\node[thumb, fill=blue!10] (uin) at (-6,0) {};
\foreach \k in {1,...,4}{
  \draw[black!35, line width=0.3pt]
    ([xshift=\k*2.1mm]uin.south west) -- ([xshift=\k*2.1mm]uin.north west);
  \draw[black!35, line width=0.3pt]
    ([yshift=\k*2.1mm]uin.south west) -- ([yshift=\k*2.1mm]uin.south east);
}
\node[lbl, below=1mm of uin] {$u_0(x)$};

\node[box]   (lift) at (-3,0) {\textbf{Lifting}\\[-1pt]\scriptsize node encoder};
\node[mpbox] (mp)   at (0,0) {\textbf{MP}$\times \Nlayers$\\[-1pt]\scriptsize physics-gated};
\node[box]   (dec)  at (3,0) {\textbf{Shared}\\[-1pt]\textbf{Decoder}};

\node[thumb, left color=blue!15, right color=red!28] (uout) at (6,0) {};
\node[lbl, below=1mm of uout] {$u(x,t)$};

\draw[arr] (uin.east) -- (lift.west);
\draw[arr] (lift.east) -- (mp.west);
\draw[arr] (mp.east) -- (dec.west);
\draw[arr] (dec.east) -- (uout.west);

\draw[aux] (mp.south) to[out=-70,in=-110]
  node[lbl, below=1.5mm, text=orange!60!black]
  {deep supervision: $\hat u^{(\ell)}=\mathrm{Decoder}(h^{(\ell)})$}
  (dec.south);


\node[panel] at (0,-2.30)
  {(b)\ One physics-gated message-passing layer};

\node[smallbox] (neigh) at (-6,-4.35)
  {candidate\\neighbors\\[-1pt]$\mathcal{N}_{i,n}^{(k_x, k_t)}$};

\node[widebox] (emlp) at (-3.50,-4.35)
  {\textbf{Edge MLP}\\[-1pt]
   \scriptsize adj / non-adj\\[-1pt]
   \scriptsize $m_k=\mathrm{MLP}([h_{i,n},h_{j,m},e_{ij}])$};

\node[gate] (gup)   at (-0.25,-3.50) {$g_{\mathrm{up}}$\\[-1pt]\scriptsize upwind};
\node[gate] (gent)  at (-0.25,-4.35) {$g_{\mathrm{ent}}$\\[-1pt]\scriptsize entropy};
\node[gate] (gtime) at (-0.25,-5.20) {$g_{\mathrm{time}}$\\[-1pt]\scriptsize distance};

\node[op] (prod) at (1.25,-4.35) {$\times$};
\node[lbl, above=1mm of prod] {$g_k$};

\node[op] (agg) at (2.5,-4.35) {$\sum$};
\node[lbl, above=1mm of agg]
  {$\displaystyle M_{i,n}=\sum_k\frac{g_k}{\sum_j g_j+\varepsilon}\,m_k$};

\node[widebox] (upd) at (5,-4.35)
  {\textbf{Residual update}\\[-1pt]
   \scriptsize $h_{i,n}^{\ell+1}=\sigma(\mathrm{MLP}([h^{\ell}_{i,n},M_{i,n}])+Wh^{\ell}_{i,n})$};

\draw[msg] (neigh.east) -- (emlp.west);

\draw[msg] (emlp.east) -- ++(0.35,0) |- (gup.west);
\draw[msg] (emlp.east) -- ++(0.35,0) -- (gent.west);
\draw[msg] (emlp.east) -- ++(0.35,0) |- (gtime.west);

\draw[msg] (gup.east) -- (prod.north west);
\draw[msg] (gent.east) -- (prod.west);
\draw[msg] (gtime.east) -- (prod.south west);

\draw[msg] (emlp.south) to[out=-90,in=-90]
  node[lbl, below] {$m_k$}
  (agg.south west);

\draw[arr] (prod.east) -- (agg.west);
\draw[arr] (agg.east) -- (upd.west);

\node[lbl, text=orange!65!black] at (-0.75,-6.0)
  {physics gate: $g_k=g_{\mathrm{up}}\,g_{\mathrm{ent}}\,g_{\mathrm{time}}$};

\begin{scope}[on background layer]
  \node[rounded corners=5pt, fill=gray!4, draw=black!7,
        fit=(uin)(lift)(mp)(dec)(uout), inner xsep=14pt, inner ysep=14pt] {};
  \node[rounded corners=5pt, fill=gray!4, draw=black!7,
        fit=(neigh)(emlp)(gup)(gtime)(upd), inner xsep=14pt, inner ysep=14pt] {};
\end{scope}

\end{tikzpicture}%
      }
    \vspace{2pt}
    \caption{HypNO. (a) The model lifts $u_0(x)$ onto a
          space-time grid and predicts the full field $u$ over the $(t,x)$-coordinates through $L$ physics-gated message-passing layers. (b) One layer: messages on a
         stencil are weighted by a physics gate
          $g=g_{\mathrm{up}}g_{\mathrm{ent}}g_{\mathrm{time}}$.}
    \label{fig:hypno-arch}
\end{figure}

\subsection{Space-time grid and operator learning}
\label{sec:st-grid}
HypNO is a neural operator. Rather than advancing the solution regressively in time over a space grid, it maps an initial condition directly to the solution field. More precisely, let $n_x$ and $n_t$ represent the number of spatial and temporal cells, respectively. Then, given an initial state $u_0(x)$ sampled on a spatial grid $\{x_i\}_{i=1}^{n_x}$, the model starts by stacking the initial condition onto the entire space-time grid
$\{(x_i,t_n)\}_{i=1,\ldots,n_x}^{\,n=1,\ldots,n_t}$ which provides a snapshot of the field
$u(x_i,t_n)$ at every node in a single forward pass.

To respect the finite domain of dependence of hyperbolic problems, \cite{leveque2002fv}, a node $(i,n)$ may only receive messages from nodes at the same or earlier time levels---i.e. the node can communicate with the nodes $(i, n-q)$, for $q\in\{0, \cdots, k_t\}$, where $k_t$ is the temporal neighboring index (see \Cref{neighborhood}).

\subsection{Lifting layer}
\label{sec:lifting}
Before the processor, the \emph{lifting layer} maps every space-time node to a latent representation $h^{(0)}_{(i,n)}\in\mathbb{R}^{d}$. 
Given the lifted (time-stacked) initial condition, the nodes go through a single gated message-passing layer in order to produce a better initialization. The detailed encoder design is given in \Cref{app:lifting}, and the per-PDE node features in \Cref{app:lwr-design} (LWR) and \Cref{sec:arz-design} (ARZ). The
resulting $h^{(0)}$ is the input to the processor of message-passing layers.

\subsection{Message passing GNN}
For updating the state of our nodes, we choose the message passing mechanism, as it resembles the behavior of numerical schemes in preserving the physical structure of the problem.

Message passing is defined as the process of collecting and aggregating messages from nodes within a \emph{neighborhood} $\mathcal{N}_{i, n}^{(k_x, k_t)}$ of the source node $(i, n)$ with $i \in \set{0,\cdots, n_x}$ and $n \in \set{0, \cdots, n_t}$ (see \Cref{app:neighborhood} for more details) and then updating the latent state of the node using the aggregate, most often performed via an update MLP. This process is repeated for $\ell = 1, \cdots, L$, allowing each node to gather information from outside of its immediate neighborhood.

\subsubsection{Neighbor messages}
A message is the information being passed from a neighbor $(j,m)$ to a source node $(i,n)$, defined as the output of a function.
\begin{equation}
\label{message-def}
m^{(\ell)}_{(j,m),(i,n)} = \phi_{\theta}\Bigl(h^{_(\ell)}_{(i,n)}, h^{(\ell)}_{(j,m)}, e^{(\ell)}_{(j,m),(i,n)} \Bigr)
\end{equation}
With $h_{(i,n)}^{(\ell)}, h_{(j,m)}^{(\ell)}$ being the hidden latent states of nodes $(i,n), (j,m)$, and $e_{(j,m),(i,n)}^{(\ell)}$ is the edge feature vector between the source and the neighbor node, defined in Section~\ref{edge-features}. The parameter $\theta$ is a usual MLP.

\subsubsection{Message aggregate}
For each source node, we denote the incoming message to be the aggregate of all the messages $m^{(\ell)}_{(j,m),(i,n)}$ for all $(j,m) \in \mathcal{N}_{(i,n)}^{(k_x, k_t)} \setminus \set{(i, n)}$
\begin{equation}
\label{message_aggregate}
M^{(\ell)}_{(i,n)}
=
\operatorname{AGG}_{(j,m)}
\left(
m^{(\ell)}_{(j,m),(i,n)}
\right),
\end{equation}
and the aggregation function $AGG$ is defined in \Cref{E:aggregation}.
\subsubsection{Latent update}
After obtaining the incoming message aggregate, the source node's latent representation $h_{(i,n)}^{(\ell)}$ is updated via an operation on its current state at layer $\ell$ and the message aggregate. 
 \begin{equation}
      {h}^{(\ell+1)}_{(i,n)} = \sigma\!\left(
          \mathrm{MLP}_{\mathrm{upd}}\!\left(
              [{h}^{(\ell)}_{(i,n)},\, M^{(\ell)}_{(i,n)}]
          \right)
          + \mathbf{W} {h}^{(\ell)}_{(i,n)}
      \right)
  \end{equation}

The latent update procedure is repeated $L$ times, until the final output is obtained. Here $\mathbf{W}$ is a residual linear map parameter. In this work we consider $\sigma$ to be a GELU (\cite{DBLP:journals/corr/HendrycksG16}) activation function.  
\subsection{Physics-based gating}

The message passing mechanism constructs a set of candidate neighbors $\mathcal{N}_{(i,n)}^{(k_x, k_t)}$. However, not all of these candidates should contribute to the update equally. Moreover, it has been noted in \cite{rusch2023oversmoothing} that unscaled message passing leads to oversmoothing of graphs as the depth $L$ increases. Works such as GAT (\cite{velickovic2018gat}) introduce learnable message weights in the form of attention.

For the purpose of explainability and incorporation of physics, we opt for a design of structured edge weight mechanisms, aiming to provide relevant physical information to the network.
To this end, we introduce scalar edge gates. In particular, given a message (\Cref{message-def})
from node $(j,m)$ to $(i,n)$, a gate is a function with $g_{(j,m),(i,n)}^{(\ell)} \in [0,1]$, that act as a weight of the messages between the nodes.
More precisely, from \Cref{message_aggregate}, we have
\begin{equation}\label{E:aggregation}
M_{(i,n)}^{(\ell)}
=
\sum_{(j,m)\in\mathcal{N}_{(i,n)}^{(k_x,k_t)}}
\frac{
g_{(j,m),(i,n)}^{(\ell)}
}{
\sum_{(j',m')\in\mathcal{N}_{(i,n)}^{(k_x,k_t)}} g_{(j',m'),(i,n)}^{(\ell)}+\varepsilon
}
m_{(j,m),(i,n)}^{(\ell)}.
\end{equation}
for sufficiently small $\eps$, which chosen to be $10^{-6}$ in this paper. 
The physics gate is a combination of the following:
\begin{itemize}
\item \textbf{Upwind gate} -- encoding the direction in which information travels
\item \textbf{Entropy gate} -- filters out messages from entropy-violating edges
\item \textbf{Temporal gate} -- acts as a physics-informed weighing mechanism for messages coming from the past.
\end{itemize}
While the general idea remains the same, the specific design of each of these gates depends on the type of the target PDE; the precise per-PDE definitions are
given in \Cref{app:lwr-design} (for LWR) and \Cref{sec:arz-design} (for ARZ).
\subsection{Edge feature vectors}
\label{edge-features}
A central component of message passing is the edge feature vector \(e^{(\ell)}_{(j,m),(i,n)}\).
For each directed edge \((j,m)\to (i,n)\), the message MLP
receives the latent states \(h^{(\ell)}_{(i,n)},h^{(\ell)}_{(j,m)}\) together with a set of features describing their pairwise relation: relative position, wave propagation direction, and speed. Since the edge features are derived from physical values, they are computed from the decoded state probe (\Cref{probed-solution}), i.e.\ \(\hat\rho_{(i,n)}\) for LWR and \((\hat\rho_{(i,n)},\hat\omega_{(i,n)})\) for ARZ, rather than on the latent vectors directly.

We distinguish two edge classes. \emph{Adjacent} edges connect immediate spatial
neighbors at the same time level (\(m=n\), \(|j-i|=1\)); they are designed to
mimic finite volume interfaces and therefore carry interface quantities such as
characteristic and Rankine-Hugoniot speeds, upwind indicators, and conserved-value jumps. All remaining edges---temporal, diagonal, and non-local space-time edges selected by the neighborhood rule---are treated as
\emph{non-adjacent}; they do not correspond to an immediate finite-volume interface, so quantities such as the Rankine-Hugoniot speed or the upwind indicator are undefined, and only topological features are employed. A separate message MLP is used for each adjacency class. The exact per-PDE feature vectors are listed in \Cref{app:lwr-design} (for LWR model, and specifically \Cref{lwr-edge-feats})
and \Cref{sec:arz-design} (for ARZ model, and specifically \Cref{arz-edge-feats}).
\subsection{Decoder and output}
\label{sec:decoder}
A single shared decoder $\mathrm{MLP}_{\mathrm{dec}}$ maps a latent vector back to
the physical state, and it plays two roles. Inside every message-passing layer it
produces the intermediate probe
\begin{equation}
\label{probed-solution}
\hat\rho^{(\ell)}_{(i,n)}=\mathrm{MLP}_{\mathrm{dec}}(h^{(\ell)}_{(i,n)}),
\end{equation}
from which the physical edge features and gates are computed (for ARZ, $\mathrm{MLP}_{\mathrm{dec}}$ outputs the pair $(\hat\rho,\hat\omega)$; see \Cref{sec:arz-design}). After the final layer $L$, it produces the model output. The per-equation output specifics are detailed in \Cref{app:lwr-design} (for LWR model) and \Cref{sec:arz-design} (for ARZ model).

As the shared decoder yields a physical prediction at every layer, the intermediate probes $\hat\rho^{(\ell)}$ (and $\hat\omega^{(\ell)}$ for ARZ) are additionally supervised during training (deep supervision), encouraging each layer to refine a physically meaningful
state.
\section{Numerical Experiments}

 We compare the proposed HypNO architecture with existing numerical methods and operator-based approximations on the 1D-LWR and 1D-ARZ equations. The data generation procedure is discussed in Section~\ref{sec:data-construction}.







\subsection{LWR Evaluation}
We evaluate the performance of our proposed HypNO model over a varying number of discontinuities and varying initialization schemes. The evaluations presented contain configurations both seen and unseen during training, showcasing the model's zero-shot capability on different numbers of discontinuities in the input initial conditions. We benchmark the HypNO performance against FNO, a baseline operator learning method, and two numerical schemes, Godunov and a positivity-limited WENO5 reconstruction. We use the accepted LWR exact Lax-Hopf solution as ground truth in training and compare each benchmark to this result during evaluation. We stratify the evaluation by initial condition type and segment count (see Section~\Cref{sec:data-construction}). 

\texttt{piecewise\_constant} initial conditions are trained on $\{2,3,5,7,10\}$ initial segments; these cells are treated as in-distribution (ID). Tests for the \texttt{piecewise\_constant} ICs with $\{8,20,25,30\}$ segments are treated as out-of-distribution (OOD); these segment counts are not seen during training in order to test zero-shot prediction. \texttt{riemann} ICs always have 2 segments so all evaluations are treated as ID. For both ID and OOD evaluations across all IC types, we calculate mean absolute error (MAE) of the density $\rho$ prediction against the Lax-Hopf ground truth, as shown in Table \ref{tab:paper_eval_by_ic}.

\begin{figure*}[!htbp]
\centering

\begin{subfigure}[!htbp]{0.95\textwidth}
    \centering
    \includegraphics[width=0.85\linewidth,height=0.24\textheight,keepaspectratio]{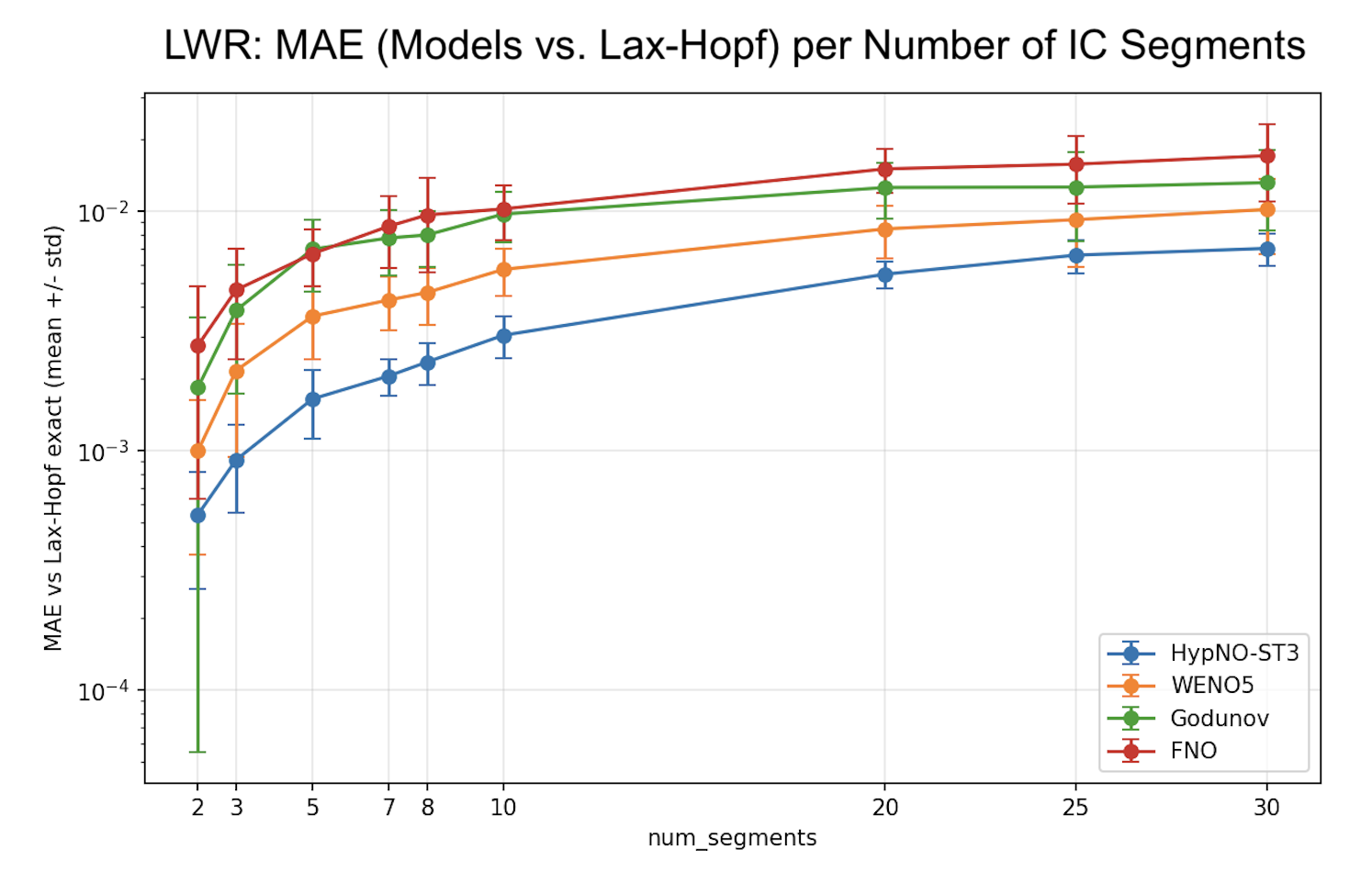}
    \caption{LWR density error versus initial-condition complexity. Mean absolute
  error on $\rho$ against the exact Lax-Hopf solution is plotted against the
  number of initial segments on a logarithmic scale; markers and error bars are
  the mean $\pm$ one standard deviation over $20$ samples per setting. Points 2, 3, 5, 7, 10 mark segment counts seen during training (ID), the
  remainder being out-of-distribution (OOD). HypNO-LWR is compared against WENO5,
  Godunov, and the learned FNO baseline, attaining the lowest error across both
  the ID and OOD complexity ranges.}
    \label{fig:mae_vs_segments}
\end{subfigure}

\vspace{0.25em}

\begin{subfigure}[!htbp]{0.95\textwidth}
    \centering
    \includegraphics[width=0.85\linewidth,height=0.24\textheight,keepaspectratio]{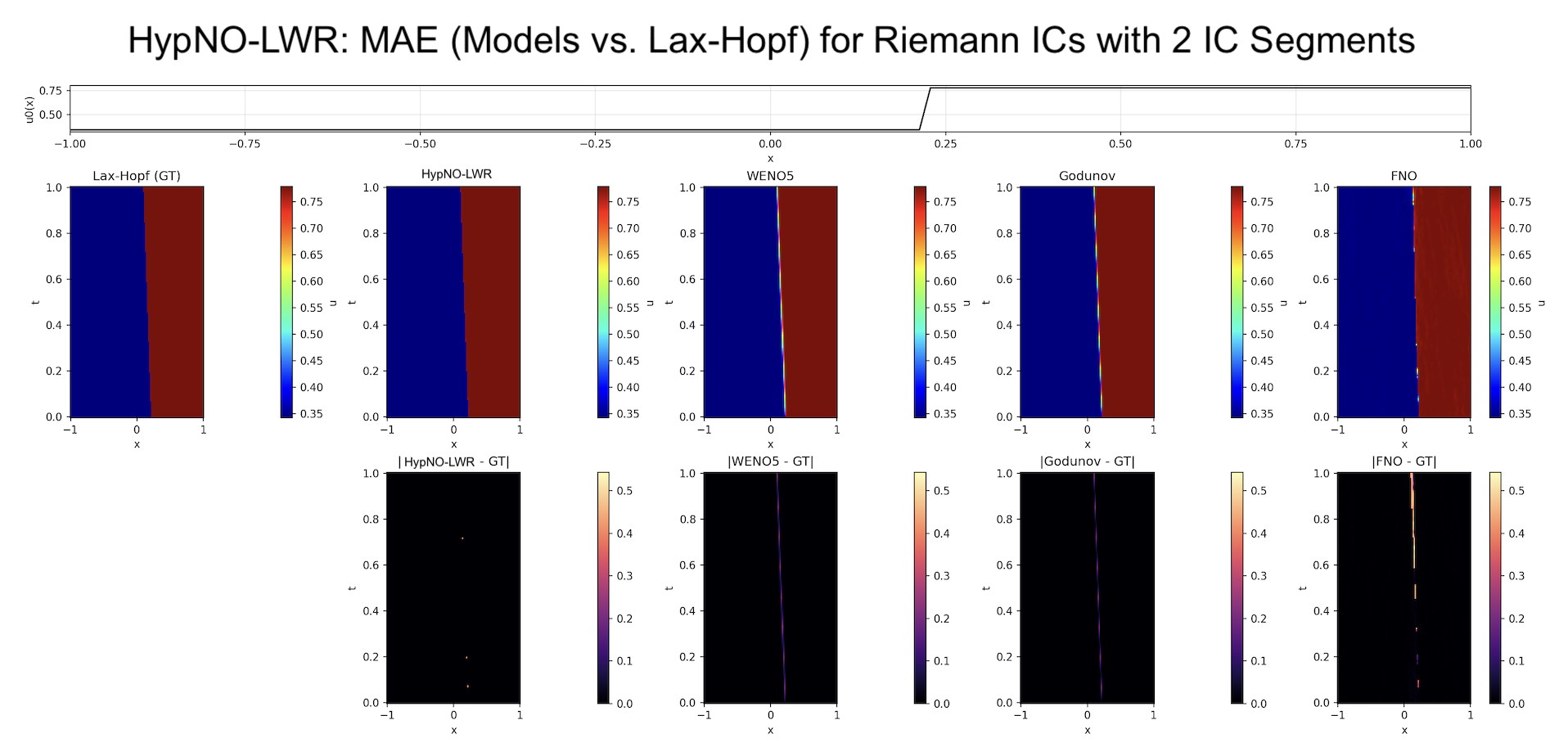}
    \caption{LWR: Representative ID example: a Riemann initial condition
  (single discontinuity). Top row: the space-time density field $\rho(x,t)$ for
  the exact Lax-Hopf ground truth and for each method (HypNO-LWR, WENO5, Godunov,
  and FNO); bottom row: the corresponding absolute error
  $|\rho-\rho_{\mathrm{GT}}|$ on a shared colour scale. HypNO-LWR resolves the
  shock as a sharp interface with near-zero error, whereas the numerical and FNO
  baselines tend to smear it into a diffused band.}
    \label{fig:qual_id_piecewise}
\end{subfigure}

\vspace{0.25em}

\begin{subfigure}[!htbp]{0.95\textwidth}
    \centering
    \includegraphics[width=0.85\linewidth,height=0.24\textheight,keepaspectratio]{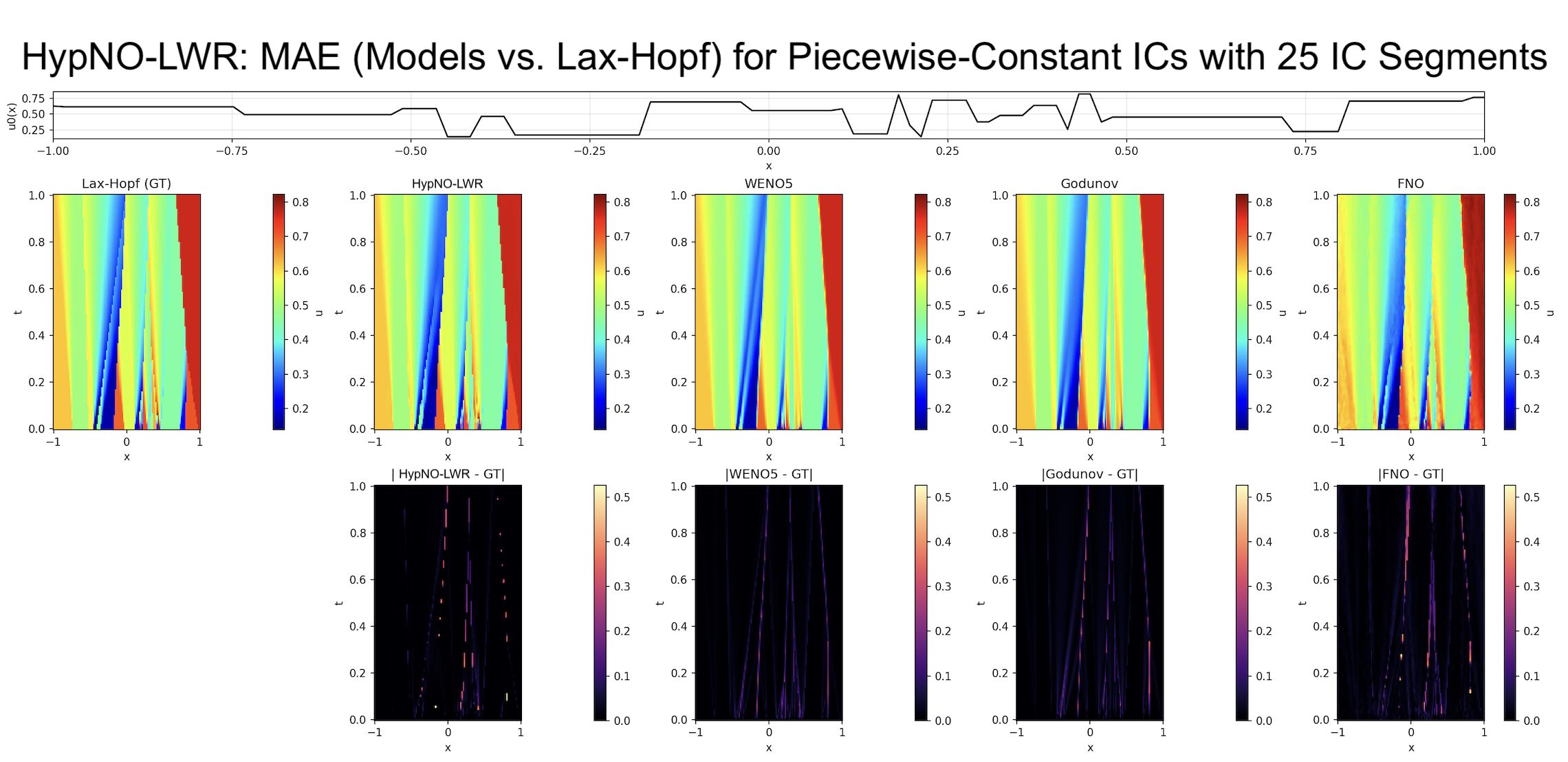}
    \caption{LWR: Representative OOD example: a piecewise-constant initial
  condition with $25$ segments, a complexity not seen during training. Top row: density fields against the Lax-Hopf ground truth; bottom row:
  absolute error on a shared color scale. Despite the many interacting shocks and
  rarefactions, HypNO-LWR keeps the interfaces sharp and the error low, while the
  numerical and FNO baselines accumulate visible diffusion.}
    \label{fig:qual_ood_const}
\end{subfigure}

\caption{Quantitative and qualitative evaluation of HypNO-LWR against classical finite volume baselines and FNO.}
\label{fig:paper_eval_summary}
\end{figure*}

\paragraph{Accuracy relative to numerical and learned baselines.}
As shown in Table \ref{tab:paper_eval_by_ic}, in both IC types, across all ID and OOD segment counts, our HypNO-LWR model attains the best mean MAE over 20 samples against the numerical schemes WENO5 and Godunov, as well as the learned FNO scheme. On average, across all ID and OOD numbers of segments, HypNO-LWR with \texttt{piecewise\_constant} initial conditions achieved a mean MAE of $3.289 \times 10^{-3}$ compared to the average WENO5 MAE ($5.474 \times 10^{-3}$), Godunov MAE ($8.500 \times 10^{-3}$), and FNO MAE ($1.007 \times 10^{-2}$). HypNO-LWR evaluated on \texttt{riemann} initial conditions attained a mean MAE of $5.481 \times 10^{-4}$, in comparison to WENO5 ($8.800 \times 10^{-4}$), Godunov ($1.504 \times 10^{-3}$), and FNO ($2.122 \times 10^{-3}$). In both IC families, HypNO-LWR showed a significant reduction in MAE against classical and learned baselines, performing the best on the \texttt{riemann} ICs for all evaluated ID segment counts. Figure \ref{fig:mae_vs_segments} depicts the MAE from the Lax-Hopf ground truth for each evaluated number of initial segments and each numerical or learned scheme, across both IC families. Figure \ref{fig:qual_id_piecewise} shows an example ID evaluation on the \texttt{riemann} IC family for our model and each of the baselines.

\paragraph{Out-of-distribution generalization.}
The OOD segment counts in this LWR evaluation, $\{8,20,25,30\}$, were held out of training to allow for testing of the model's zero-shot capability for \texttt{piecewise\_constant} ICs. All OOD tests with HypNO-LWR attained a MAE consistent with ID evaluation, and performed better than OOD evaluation with WENO5, Godunov, and FNO. HypNO-LWR remains the best performing model against all baselines, even OOD, as seen by the attained MAE in each IC family in Table \ref{tab:paper_eval_by_ic}. This result is vital in proving the model's ability to extrapolate and achieve competitive results with complex initial conditions unseen during training and significantly larger than some ID segment counts.

\paragraph{Error consistency.}
In addition to smaller MAE across segment count and IC family, our HypNO-LWR model saw smaller error standard deviations, demonstrating a tighter error spread compared to all baseline schemes. For \texttt{riemann} ICs, HypNO-LWR saw an average error standard deviation of $\pm1.986 \times 10^{-4}$ compared to WENO5 ($\pm4.974 \times 10^{-4}$), Godunov ($\pm1.166 \times 10^{-3}$), and FNO ($\pm1.153 \times 10^{-3}$). For \texttt{piecewise\_constant} ICs, HypNO-LWR saw an average error standard deviation of $\pm6.073 \times 10^{-4}$ compared to WENO5 ($\pm1.737 \times 10^{-3}$), Godunov ($\pm2.918 \times 10^{-3}$), and FNO ($\pm3.327 \times 10^{-3}$). The smaller error spread attained by HypNO-LWR shows consistency in it's accuracy across initial conditions, for both ID and OOD evaluations, and initial condition families. It also proves better precision than all baseline numerical and learned schemes. Fig. \ref{fig:qual_ood_const} depicts an example OOD evaluation with \texttt{piecewise\_constant} initial conditions and 25 segments, showing smaller MAE from the Lax-Hopf ground truth for HypNO-LWR compared to WENO5, Godunov, and FNO. 

\begin{table*}[t]
\centering
\small
\setlength{\tabcolsep}{5pt}
\caption{LWR: mean absolute error against the exact Lax-Hopf solution,
stratified by initial-condition family and number of segments and reported as
mean $\pm$ standard deviation over $20$ samples per setting. HypNO-LWR is
compared against the numerical WENO5 and Godunov schemes and the learned FNO
baseline; the lowest mean MAE in each row is highlighted in bold. Segment counts $\{2,3,5,7,10\}$
seen during training are in-distribution, while larger counts $\{8,20,25,30\}$ are
out-of-distribution and probe zero-shot generalization to more complex initial
data. HypNO-LWR attains the lowest error across all families and complexities,
with a consistently tighter error spread than every baseline.}
\label{tab:paper_eval_by_ic}
\begin{tabular}{llcccc}
\toprule
IC type & N segs & HypNO-LWR & WENO5 & Godunov & FNO \\
\midrule

\multirow{8}{*}{\texttt{pcw\_const}}
& 2  & \besterr{5.41{\times}10^{-4}}{2.75{\times}10^{-4}} & \err{9.98{\times}10^{-4}}{6.30{\times}10^{-4}} & \err{1.84{\times}10^{-3}}{1.78{\times}10^{-3}} & \err{2.75{\times}10^{-3}}{2.12{\times}10^{-3}} \\
& 3  & \besterr{9.16{\times}10^{-4}}{3.65{\times}10^{-4}} & \err{2.16{\times}10^{-3}}{1.22{\times}10^{-3}} & \err{3.86{\times}10^{-3}}{2.13{\times}10^{-3}} & \err{4.70{\times}10^{-3}}{2.28{\times}10^{-3}} \\
& 5  & \besterr{1.65{\times}10^{-3}}{5.24{\times}10^{-4}} & \err{3.65{\times}10^{-3}}{1.23{\times}10^{-3}} & \err{6.95{\times}10^{-3}}{2.32{\times}10^{-3}} & \err{6.66{\times}10^{-3}}{1.77{\times}10^{-3}} \\
& 7  & \besterr{2.05{\times}10^{-3}}{3.64{\times}10^{-4}} & \err{4.27{\times}10^{-3}}{1.07{\times}10^{-3}} & \err{7.75{\times}10^{-3}}{2.34{\times}10^{-3}} & \err{8.68{\times}10^{-3}}{2.89{\times}10^{-3}} \\
& 8  & \besterr{2.35{\times}10^{-3}}{4.60{\times}10^{-4}} & \err{4.58{\times}10^{-3}}{1.22{\times}10^{-3}} & \err{7.97{\times}10^{-3}}{2.10{\times}10^{-3}} & \err{9.67{\times}10^{-3}}{4.08{\times}10^{-3}} \\
& 10 & \besterr{3.04{\times}10^{-3}}{6.13{\times}10^{-4}} & \err{5.72{\times}10^{-3}}{1.27{\times}10^{-3}} & \err{9.75{\times}10^{-3}}{2.28{\times}10^{-3}} & \err{1.02{\times}10^{-2}}{2.64{\times}10^{-3}} \\
& 20 & \besterr{5.47{\times}10^{-3}}{7.15{\times}10^{-4}} & \err{8.45{\times}10^{-3}}{2.10{\times}10^{-3}} & \err{1.26{\times}10^{-2}}{3.31{\times}10^{-3}} & \err{1.51{\times}10^{-2}}{3.12{\times}10^{-3}} \\
& 25 & \besterr{6.57{\times}10^{-3}}{1.06{\times}10^{-3}} & \err{9.24{\times}10^{-3}}{3.37{\times}10^{-3}} & \err{1.26{\times}10^{-2}}{5.14{\times}10^{-3}} & \err{1.58{\times}10^{-2}}{4.98{\times}10^{-3}} \\
& 30 & \besterr{7.01{\times}10^{-3}}{1.09{\times}10^{-3}} & \err{1.02{\times}10^{-2}}{3.52{\times}10^{-3}} & \err{1.32{\times}10^{-2}}{4.86{\times}10^{-3}} & \err{1.71{\times}10^{-2}}{6.06{\times}10^{-3}} \\

\midrule

\multirow{8}{*}{\texttt{riemann}}
& 2  & \besterr{5.26{\times}10^{-4}}{1.49{\times}10^{-4}} & \err{7.41{\times}10^{-4}}{4.64{\times}10^{-4}} & \err{1.24{\times}10^{-3}}{8.01{\times}10^{-4}} & \err{1.87{\times}10^{-3}}{8.76{\times}10^{-4}} \\
& 2  & \besterr{4.98{\times}10^{-4}}{1.69{\times}10^{-4}} & \err{8.25{\times}10^{-4}}{5.40{\times}10^{-4}} & \err{1.40{\times}10^{-3}}{1.04{\times}10^{-3}} & \err{2.04{\times}10^{-3}}{1.04{\times}10^{-3}} \\
& 2  & \besterr{5.37{\times}10^{-4}}{2.09{\times}10^{-4}} & \err{9.04{\times}10^{-4}}{5.44{\times}10^{-4}} & \err{1.51{\times}10^{-3}}{1.35{\times}10^{-3}} & \err{2.35{\times}10^{-3}}{1.26{\times}10^{-3}} \\
& 2  & \besterr{4.39{\times}10^{-4}}{1.13{\times}10^{-4}} & \err{6.65{\times}10^{-4}}{3.87{\times}10^{-4}} & \err{1.02{\times}10^{-3}}{5.59{\times}10^{-4}} & \err{2.00{\times}10^{-3}}{7.72{\times}10^{-4}} \\
& 2  & \besterr{6.70{\times}10^{-4}}{2.94{\times}10^{-4}} & \err{9.97{\times}10^{-4}}{4.29{\times}10^{-4}} & \err{2.64{\times}10^{-3}}{1.77{\times}10^{-3}} & \err{1.86{\times}10^{-3}}{6.80{\times}10^{-4}} \\
& 2 & \besterr{5.99{\times}10^{-4}}{2.60{\times}10^{-4}} & \err{9.36{\times}10^{-4}}{5.10{\times}10^{-4}} & \err{2.04{\times}10^{-3}}{1.49{\times}10^{-3}} & \err{2.23{\times}10^{-3}}{1.34{\times}10^{-3}} \\
& 2 & \besterr{5.99{\times}10^{-4}}{2.31{\times}10^{-4}} & \err{9.36{\times}10^{-4}}{5.53{\times}10^{-4}} & \err{1.79{\times}10^{-3}}{1.21{\times}10^{-3}} & \err{2.19{\times}10^{-3}}{1.92{\times}10^{-3}} \\
& 2 & \besterr{5.57{\times}10^{-4}}{1.83{\times}10^{-4}} & \err{9.69{\times}10^{-4}}{5.21{\times}10^{-4}} & \err{1.72{\times}10^{-3}}{1.28{\times}10^{-3}} & \err{2.17{\times}10^{-3}}{9.33{\times}10^{-4}} \\
& 2 & \besterr{5.08{\times}10^{-4}}{1.79{\times}10^{-4}} & \err{9.47{\times}10^{-4}}{5.29{\times}10^{-4}} & \err{1.72{\times}10^{-3}}{9.96{\times}10^{-4}} & \err{2.39{\times}10^{-3}}{1.56{\times}10^{-3}} \\

\bottomrule
\end{tabular}
\end{table*}

\FloatBarrier

\FloatBarrier

\paragraph{Shock neighborhood evaluation (LWR)}
Given the nature of the problem and the purpose of the model, we also choose to report performance comparison around shocks. We score every method on a single \emph{shock band} detected on the Lax-Hopf
ground truth and reused, unchanged, across methods. The detector---a signed Lax
entropy test, a local total-variation gate, and a dilation into a fixed-width
neighborhood---is defined in \Cref{app:shock-band}.
\begin{center}
\centering
\captionof{table}{LWR accuracy on shock neighborhoods, as a function of initial-condition
complexity. For each number of initial segments, \emph{full} is the mean
absolute error against the exact Lax-Hopf solution over the whole space-time
domain, and \emph{shock} restricts the same error to a shock-band mask detected
on the ground truth via the signed Lax entropy condition $\rho_L<\rho_R$ (threshold
$\tau=0.06$) and dilated to a fixed-width neighborhood; the identical mask is
reused across all methods. HypNO-LWR is compared against WENO5, Godunov, and the
learned FNO baseline, with the lowest mean MAE in each row in bold. HypNO-LWR
attains the smallest error in every setting, and the margin widens on the shock
band, where the numerical and learned baselines smear the discontinuity.}
\label{tab:shock-mae}
\small
\begin{tabular}{ccccccccc}
\toprule
num.\ seg. & \multicolumn{2}{c}{HypNO-LWR} & \multicolumn{2}{c}{WENO5} & \multicolumn{2}{c}{Godunov} & \multicolumn{2}{c}{FNO} \\
\cmidrule(lr){2-3} \cmidrule(lr){4-5} \cmidrule(lr){6-7} \cmidrule(lr){8-9}
 & full & shock & full & shock & full & shock & full & shock \\
\midrule
2  & $\mathbf{4.98{\times}10^{-4}}$ & $\mathbf{3.24{\times}10^{-3}}$ & $7.27{\times}10^{-4}$ & $2.37{\times}10^{-2}$ & $1.33{\times}10^{-3}$ & $2.49{\times}10^{-2}$ & $3.92{\times}10^{-3}$ & $6.37{\times}10^{-2}$ \\
3  & $\mathbf{8.25{\times}10^{-4}}$ & $\mathbf{4.73{\times}10^{-3}}$ & $3.27{\times}10^{-3}$ & $4.83{\times}10^{-2}$ & $3.16{\times}10^{-3}$ & $2.61{\times}10^{-2}$ & $6.99{\times}10^{-3}$ & $6.47{\times}10^{-2}$ \\
5  & $\mathbf{1.49{\times}10^{-3}}$ & $\mathbf{8.10{\times}10^{-3}}$ & $3.23{\times}10^{-3}$ & $2.80{\times}10^{-2}$ & $6.25{\times}10^{-3}$ & $3.22{\times}10^{-2}$ & $6.19{\times}10^{-3}$ & $3.35{\times}10^{-2}$ \\
7  & $\mathbf{1.80{\times}10^{-3}}$ & $\mathbf{5.55{\times}10^{-3}}$ & $4.11{\times}10^{-3}$ & $2.51{\times}10^{-2}$ & $7.27{\times}10^{-3}$ & $2.50{\times}10^{-2}$ & $8.29{\times}10^{-3}$ & $3.31{\times}10^{-2}$ \\
8  & $\mathbf{2.16{\times}10^{-3}}$ & $\mathbf{9.67{\times}10^{-3}}$ & $4.63{\times}10^{-3}$ & $3.11{\times}10^{-2}$ & $6.43{\times}10^{-3}$ & $2.95{\times}10^{-2}$ & $1.29{\times}10^{-2}$ & $6.50{\times}10^{-2}$ \\
10 & $\mathbf{2.58{\times}10^{-3}}$ & $\mathbf{8.46{\times}10^{-3}}$ & $5.72{\times}10^{-3}$ & $2.84{\times}10^{-2}$ & $9.86{\times}10^{-3}$ & $3.30{\times}10^{-2}$ & $9.17{\times}10^{-3}$ & $3.57{\times}10^{-2}$ \\
20 & $\mathbf{4.93{\times}10^{-3}}$ & $\mathbf{1.20{\times}10^{-2}}$ & $9.42{\times}10^{-3}$ & $2.81{\times}10^{-2}$ & $1.40{\times}10^{-2}$ & $2.91{\times}10^{-2}$ & $1.53{\times}10^{-2}$ & $3.59{\times}10^{-2}$ \\
25 & $\mathbf{6.91{\times}10^{-3}}$ & $\mathbf{1.59{\times}10^{-2}}$ & $1.15{\times}10^{-2}$ & $3.18{\times}10^{-2}$ & $1.78{\times}10^{-2}$ & $3.47{\times}10^{-2}$ & $2.14{\times}10^{-2}$ & $4.86{\times}10^{-2}$ \\
30 & $\mathbf{7.52{\times}10^{-3}}$ & $\mathbf{1.49{\times}10^{-2}}$ & $1.29{\times}10^{-2}$ & $3.20{\times}10^{-2}$ & $1.82{\times}10^{-2}$ & $3.23{\times}10^{-2}$ & $1.96{\times}10^{-2}$ & $4.32{\times}10^{-2}$ \\
\bottomrule
\end{tabular}
\end{center}

\begin{center}
  \includegraphics[width=0.8\linewidth]{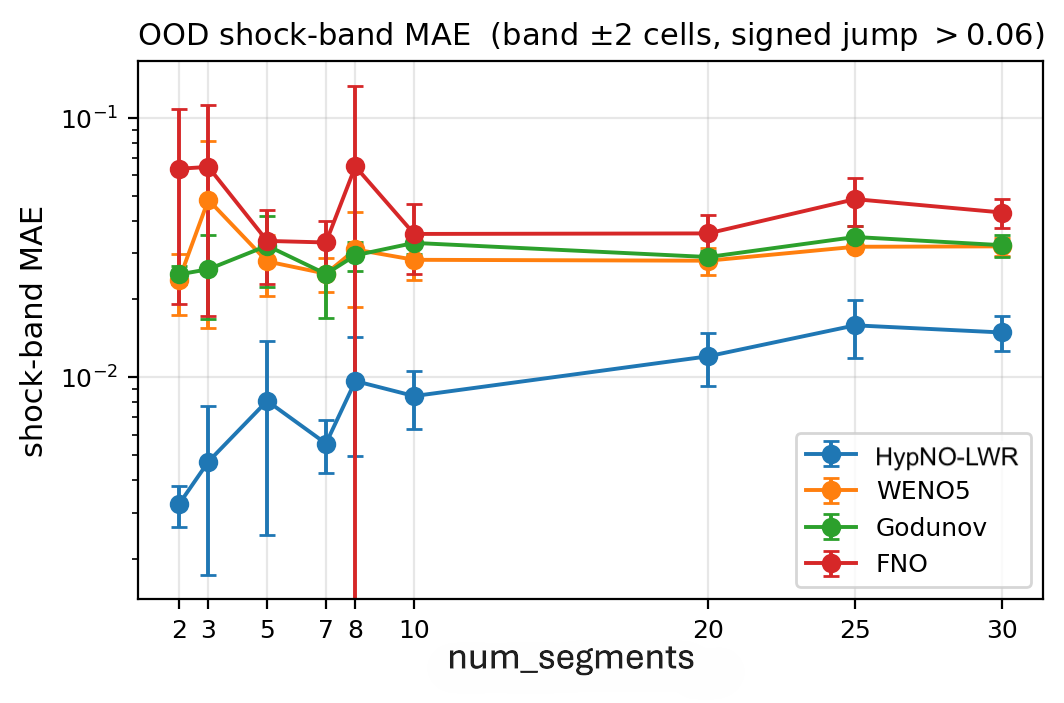}
  \captionof{figure}{LWR: MAE restricted to the shock band, as a function of the number of
    discontinuities in the initial condition. HypNO-LWR is compared against WENO5, Godunov, and the FNO
    baseline, all scored on the same Lax-Hopf ground truth. HypNO-LWR keeps the
    in-band error well below every baseline across the full complexity range.}
  \label{fig:shock-mae-vs-seg}
\end{center}

\begin{figure}[t]
  \centering
  \includegraphics[width=0.7\linewidth]{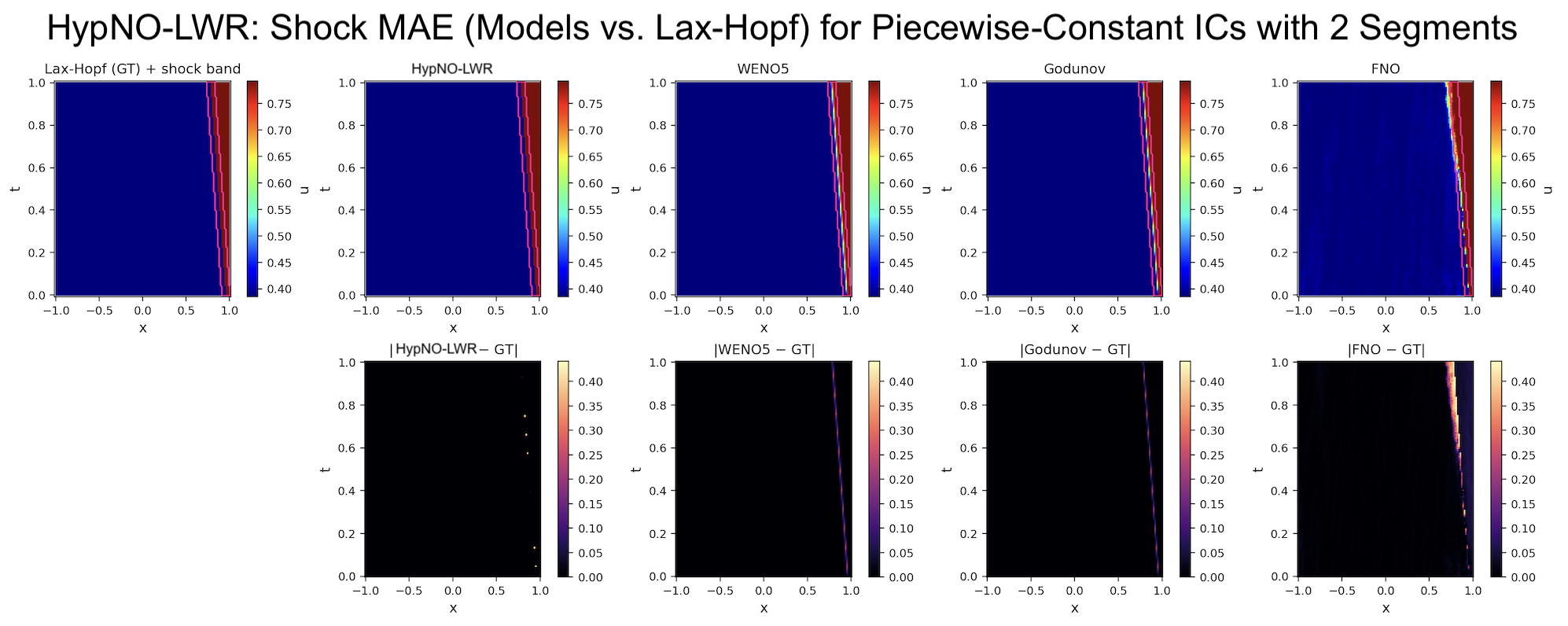}
  \caption{LWR: Representative sample for an initial condition partitioned into $N=2$ segments. Top row: the space-time
    density fields for the Lax-Hopf ground truth and each method (HypNO-LWR,
    WENO5, Godunov, FNO), with the detected shock band outlined in pink; bottom
    row: the absolute error to the Lax-Hopf ground truth. The baselines'
    error concentrates inside the band, while HypNO-LWR remains sharp.}
    \label{fig:shock-compare-2}
\end{figure}
\begin{figure}[t]
  \centering
  \includegraphics[width=0.7\linewidth]{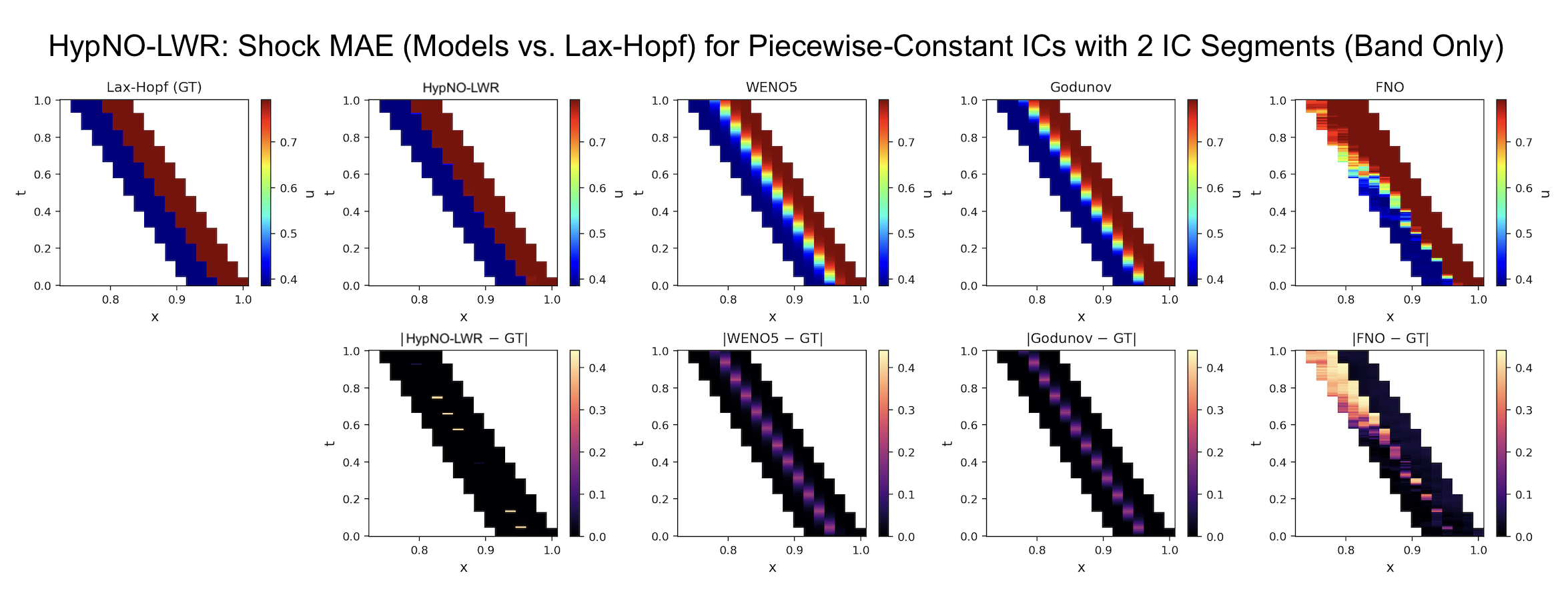}
  \caption{Same LWR representative sample as Fig.~\ref{fig:shock-compare-2}, zoomed to the
    shock-band bounding box with non-band cells blanked. Color scales are computed
    inside the band, so within-band differences are not washed out by the
    full-domain range. }
  \label{fig:shock-zoom-2}
\end{figure}
\begin{figure}[t]
  \centering
  \includegraphics[width=0.8\linewidth]{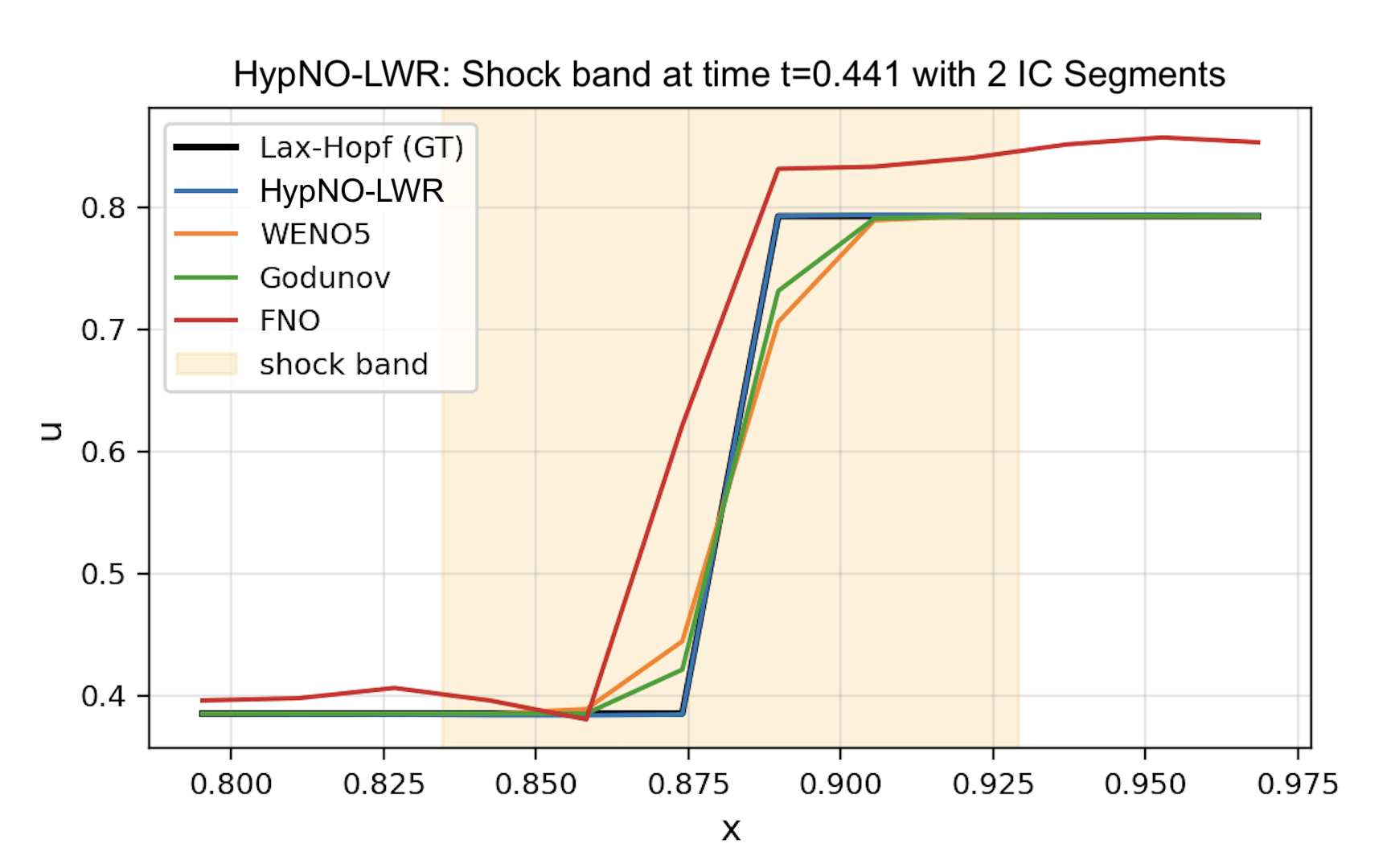}
  \caption{$\rho(x,t)$ at $t=0.441$ for the widest-band row with an initial condition consisting of $N=2$ segments. The shaded
    region marks the shock band; HypNO-LWR tracks the sharp ground-truth front
    while the baselines round corners.}
  \label{fig:shock-slice-2}
\end{figure}
\begin{figure}[t]
  \centering
  \includegraphics[width=0.7\linewidth]{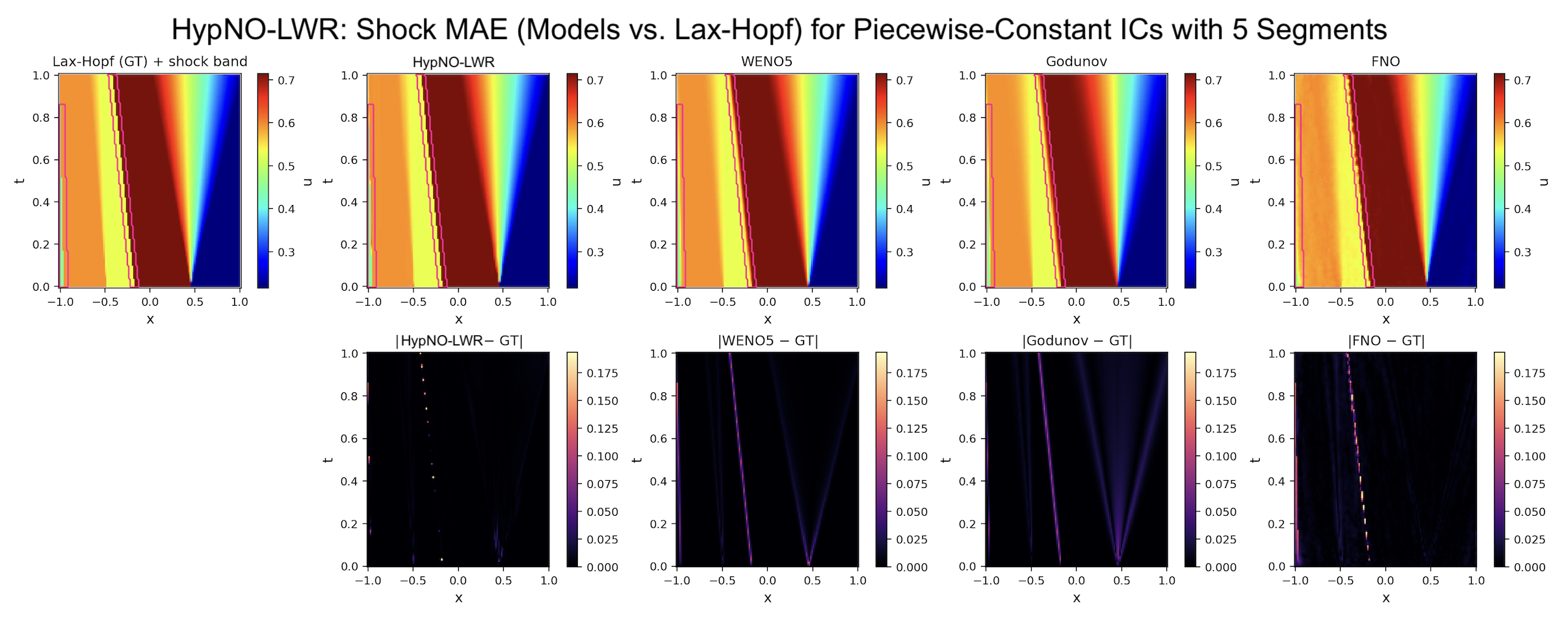}
  \caption{LWR: Representative sample for an initial condition partitioned into $N=5$ segments. Top row: the space-time
    density fields for the Lax-Hopf ground truth and each method (HypNO-LWR,
    WENO5, Godunov, FNO), with the detected shock band outlined in pink; bottom
    row: the absolute error to the Lax-Hopf ground truth. The baselines'
    error concentrates inside the band, while HypNO-LWR stays sharp.}
  \label{fig:shock-compare-5}
\end{figure}
\begin{figure}[t]
  \centering
  \includegraphics[width=0.7\linewidth]{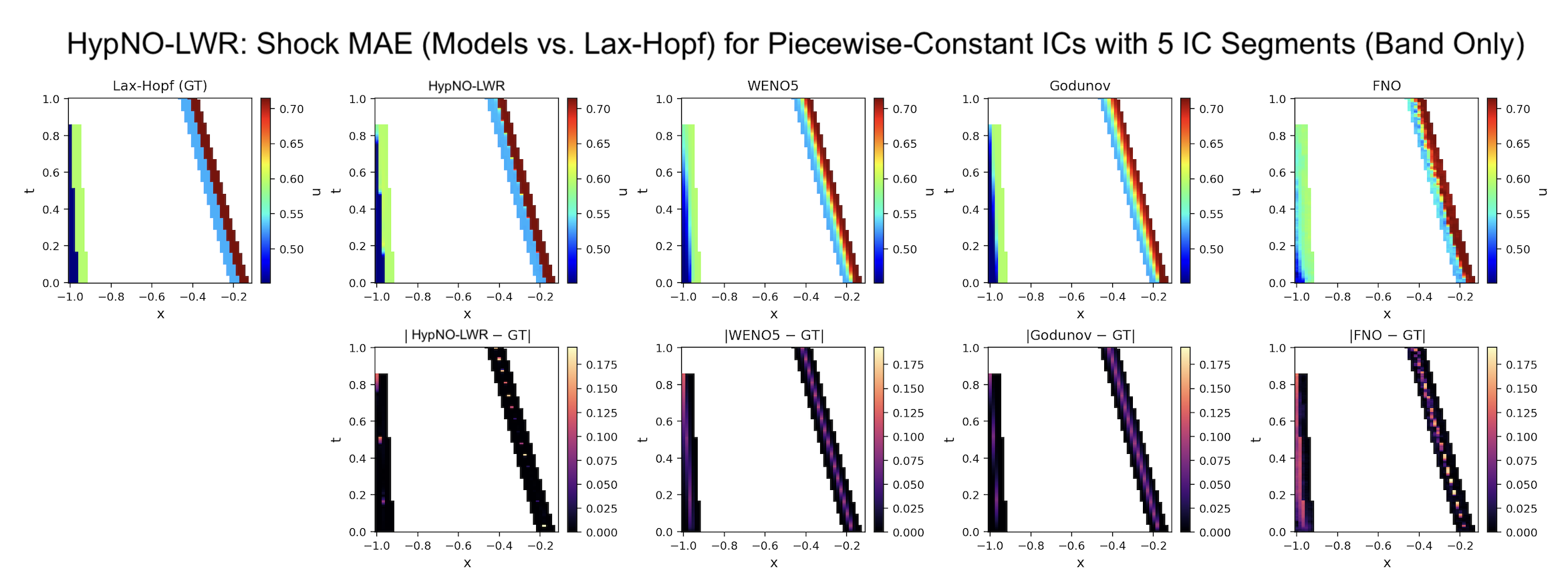}
  \caption{Same LWR representative sample as Fig.~\ref{fig:shock-compare-5}, zoomed to the
    shock-band bounding box with non-band cells blanked. Color scales are computed
    inside the band, so within-band differences are not washed out by the
    full-domain range.}
  \label{fig:shock-zoom-5}
\end{figure}
\begin{figure}[t]
  \centering
  \includegraphics[width=0.8\linewidth]{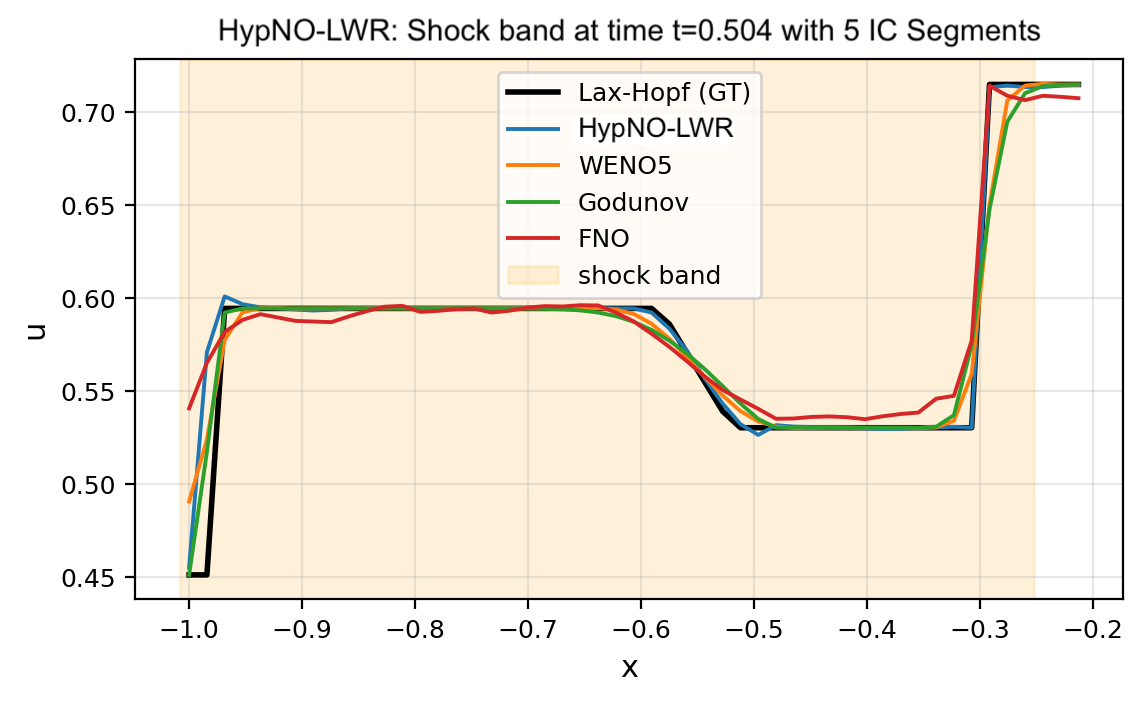}
  \caption{$\rho(x,t)$ at $t=0.504$ through the widest band row at $N=5$ segments. The shaded
    region marks the shock band; HypNO-LWR tracks the sharp ground-truth front
    while the baselines round corners.}
  \label{fig:shock-slice-5}
\end{figure}

\FloatBarrier
\subsection{ARZ Evaluation}
\label{sec:arz-mark0-eval}

We evaluate HypNO on the ARZ model, training HypNO-ARZ on a stratified dataset with multiple discontinuities, as with LWR. HypNO-ARZ-HLL, a model variant pretrained on HLL reference solutions, is also evaluated with the same stratified dataset, as discussed in further detail in \Cref{app:arz-hll-eval}. For both ARZ models, we benchmark against an operator baseline (FNO) and
three classical numerical schemes (Godunov, HLL, and a positivity-limited WENO5 reconstruction),
varying number of discontinuities, and several ways of initializing them.

The evaluation is stratified by initial condition family and segment count.
Both initial condition families and the segment counts $\{2,3,5,7,10\}$ are represented in
training; we, therefore, treat these cells as in-distribution (ID) and evaluate on
fresh, held-out samples drawn with a different seed. The segment counts
$\{8,20,30\}$ not seen in training and constitute the
out-of-distribution (OOD) band; they probe the model's ability to zero-shot a
number of discontinuities beyond its training support. Ground truth is
computed to machine precision (exact Riemann solver for
\texttt{riemann\_stratified}; wave-front tracking for the \texttt{piecewise-constant} family), so the reported errors reflect each solver's true deviation
from the entropy solution rather than a discretized reference. Following standard practice for the ARZ model, we report errors in density $\rho$, the Lagrangian marker $
\omega=v+p(\rho)$, where $p(\rho)$ denotes the traffic pressure function, and the velocity $v$.

\paragraph{Accuracy relative to numerical baselines.}
Pooled across all initial condition families and segment counts, HypNO-ARZ attains a density
MAE of $1.57\times10^{-2}$, versus $5.35\times10^{-2}$ (FNO),
$5.30\times10^{-2}$ (Godunov), $5.61\times10^{-2}$ (HLL), and
$5.79\times10^{-2}$ (WENO5), demonstrating a three- to fourfold reduction in error over
\emph{every} baseline, learned and classical alike. The margin is largest on
the \texttt{riemann\_stratified} family, where the model is essentially
insensitive to the number of interfaces (density MAE between
$6.16\times10^{-3}$ and $6.80\times10^{-3}$ across all eight segment counts)
while the numerical schemes sit three- to fourfold higher
($1.9\!-\!3.0\times10^{-2}$). On \texttt{piecewise\_constant} initial conditions the
model leads by a similar factor at every segment count, e.g.\ $7.45\times
10^{-2}$ versus $2.03\!-\!2.28\times10^{-1}$ for the numerical schemes at the
hardest cell ($30$ segments). The model is the most accurate method in every
single $(\text{family},\,\text{segment})$ cell of
Tables~\ref{tab:arz_mark0_paper_eval_piecewise_constant_stratified}--%
\ref{tab:arz_mark0_paper_eval_riemann_stratified}.

\paragraph{Out-of-distribution generalization.}
The held-out segment counts $\{8,20,30\}$ test extrapolation in the number of
discontinuities. Moving from the ID band to the OOD band, the model's density
MAE rises only from $1.03\times10^{-2}$ to $2.46\times10^{-2}$ (a $2.4\times$
increase), and the OOD error remains $3.0\times$ below the best numerical scheme
on that band (Godunov, $7.35\times10^{-2}$) and $3.0\times$ below FNO
($7.37\times10^{-2}$). The same three-to-fourfold advantage that holds
in-distribution is thus retained out-of-distribution; the model does not collapse
on segment counts beyond its training support, and crucially it remains the most
accurate method in every OOD cell ($8$, $20$, and $30$ segments) of the per-family
tables.

\paragraph{Error consistency.}
Beyond the mean, the \emph{spread} of the per-sample error is markedly tighter
for HypNO-ARZ. On \texttt{riemann\_stratified} the model's standard deviation
stays near $2.8\times10^{-3}$ across all segment counts, roughly an order of
magnitude smaller than the $3.9\!-\!5.6\times10^{-2}$ spreads of the numerical
schemes on the same cells; for \texttt{piecewise-constant} data the model's per-cell
standard deviation is consistently $2\!-\!3\times$ tighter than Godunov, HLL, and WENO5.
The numerical baselines exhibit standard deviations comparable to their own means (e.g.\ Godunov at $30$ segments,
$8.94\times10^{-2}\pm9.87\times10^{-2}$ pooled), reflecting occasional large
failures on strong or closely spaced shocks. The model's tighter distribution
shows that its accuracy is uniform across initial conditions rather than
an average of lucky and unlucky cases.


\begin{table}[!htbp]
  \centering
\caption{ARZ density error on the \texttt{piecewise\_constant}
family: mean absolute error on $\rho$ against the wave-front-tracking ground
truth, reported as mean $\pm$ standard deviation per cell. Rows are the number of
initial segments; HypNO-ARZ (\emph{model}) is compared against the learned FNO
baseline and the Godunov, HLL, and WENO5 numerical schemes, with the lowest mean
MAE in each row in bold. Segment counts $\{2,3,5,7,10\}$ are in-distribution and
$\{8,20,30\}$ are out-of-distribution. HypNO-ARZ is the most accurate method in
every cell, including the out-of-distribution complexities.}
\label{tab:arz_mark0_paper_eval_piecewise_constant_stratified}
\resizebox{\textwidth}{!}{%
\begin{tabular}{lccccc}
\hline
\# segs & HypNO-ARZ & FNO & Godunov & HLL & WENO5 \\
\hline
2 & \besterr{1.98{\times}10^{-3}}{6.69{\times}10^{-4}} & \err{2.26{\times}10^{-2}}{2.75{\times}10^{-2}} & \err{2.30{\times}10^{-2}}{3.71{\times}10^{-2}} & \err{2.42{\times}10^{-2}}{3.94{\times}10^{-2}} & \err{2.11{\times}10^{-2}}{3.43{\times}10^{-2}} \\
3 & \besterr{3.21{\times}10^{-3}}{1.25{\times}10^{-3}} & \err{3.50{\times}10^{-2}}{3.17{\times}10^{-2}} & \err{3.87{\times}10^{-2}}{4.35{\times}10^{-2}} & \err{4.11{\times}10^{-2}}{4.73{\times}10^{-2}} & \err{3.98{\times}10^{-2}}{4.73{\times}10^{-2}} \\
5 & \besterr{7.29{\times}10^{-3}}{5.06{\times}10^{-3}} & \err{5.31{\times}10^{-2}}{2.71{\times}10^{-2}} & \err{5.30{\times}10^{-2}}{3.19{\times}10^{-2}} & \err{5.61{\times}10^{-2}}{3.43{\times}10^{-2}} & \err{5.87{\times}10^{-2}}{4.25{\times}10^{-2}} \\
7 & \besterr{1.08{\times}10^{-2}}{5.60{\times}10^{-3}} & \err{6.10{\times}10^{-2}}{2.24{\times}10^{-2}} & \err{6.25{\times}10^{-2}}{2.92{\times}10^{-2}} & \err{6.57{\times}10^{-2}}{3.08{\times}10^{-2}} & \err{6.90{\times}10^{-2}}{3.89{\times}10^{-2}} \\
8 & \besterr{1.21{\times}10^{-2}}{6.69{\times}10^{-3}} & \err{6.50{\times}10^{-2}}{2.70{\times}10^{-2}} & \err{6.78{\times}10^{-2}}{3.41{\times}10^{-2}} & \err{7.13{\times}10^{-2}}{3.64{\times}10^{-2}} & \err{7.03{\times}10^{-2}}{4.12{\times}10^{-2}} \\
10 & \besterr{1.68{\times}10^{-2}}{9.07{\times}10^{-3}} & \err{7.54{\times}10^{-2}}{3.25{\times}10^{-2}} & \err{7.48{\times}10^{-2}}{4.32{\times}10^{-2}} & \err{7.87{\times}10^{-2}}{4.49{\times}10^{-2}} & \err{8.48{\times}10^{-2}}{4.46{\times}10^{-2}} \\
20 & \besterr{3.97{\times}10^{-2}}{2.16{\times}10^{-2}} & \err{1.04{\times}10^{-1}}{5.97{\times}10^{-2}} & \err{1.06{\times}10^{-1}}{7.61{\times}10^{-2}} & \err{1.11{\times}10^{-1}}{7.98{\times}10^{-2}} & \err{1.11{\times}10^{-1}}{6.41{\times}10^{-2}} \\
30 & \besterr{5.66{\times}10^{-2}}{3.38{\times}10^{-2}} & \err{1.25{\times}10^{-1}}{8.40{\times}10^{-2}} & \err{1.25{\times}10^{-1}}{1.01{\times}10^{-1}} & \err{1.30{\times}10^{-1}}{1.05{\times}10^{-1}} & \err{1.38{\times}10^{-1}}{8.83{\times}10^{-2}} \\
\hline
\end{tabular}}
\bigskip

\caption{ARZ density error on the \texttt{riemann} family: mean absolute error on
$\rho$ against the exact Riemann-solver ground truth, reported as mean $\pm$
standard deviation per cell. Because \texttt{riemann} initial conditions always
contain a single discontinuity (two segments), rows index the stratified
evaluation bins rather than segment counts, so all cells are in-distribution.
HypNO-ARZ (\emph{model}) is compared against the learned FNO baseline and the
Godunov, HLL, and WENO5 numerical schemes, with the lowest mean MAE in each row
in bold. HypNO-ARZ is the most accurate method in every cell, with an error
spread roughly an order of magnitude tighter than any baseline.}
\label{tab:arz_mark0_paper_eval_riemann_stratified}
\resizebox{\textwidth}{!}{%
\begin{tabular}{lccccc}
\hline
Bin & HypNO-ARZ & FNO & Godunov & HLL & WENO5 \\
\hline
1 & \besterr{1.75{\times}10^{-3}}{5.16{\times}10^{-4}} & \err{2.66{\times}10^{-2}}{3.94{\times}10^{-2}} & \err{2.33{\times}10^{-2}}{5.20{\times}10^{-2}} & \err{2.61{\times}10^{-2}}{5.60{\times}10^{-2}} & \err{2.47{\times}10^{-2}}{4.94{\times}10^{-2}} \\
2 & \besterr{1.72{\times}10^{-3}}{4.58{\times}10^{-4}} & \err{2.58{\times}10^{-2}}{2.53{\times}10^{-2}} & \err{1.97{\times}10^{-2}}{3.24{\times}10^{-2}} & \err{2.18{\times}10^{-2}}{3.43{\times}10^{-2}} & \err{2.31{\times}10^{-2}}{3.75{\times}10^{-2}} \\
3 & \besterr{1.74{\times}10^{-3}}{6.08{\times}10^{-4}} & \err{2.98{\times}10^{-2}}{3.78{\times}10^{-2}} & \err{2.58{\times}10^{-2}}{4.36{\times}10^{-2}} & \err{3.03{\times}10^{-2}}{5.26{\times}10^{-2}} & \err{3.50{\times}10^{-2}}{5.49{\times}10^{-2}} \\
4 & \besterr{1.75{\times}10^{-3}}{5.48{\times}10^{-4}} & \err{2.50{\times}10^{-2}}{2.69{\times}10^{-2}} & \err{2.20{\times}10^{-2}}{3.33{\times}10^{-2}} & \err{2.38{\times}10^{-2}}{3.85{\times}10^{-2}} & \err{2.46{\times}10^{-2}}{4.64{\times}10^{-2}} \\
5 & \besterr{1.72{\times}10^{-3}}{4.48{\times}10^{-4}} & \err{2.43{\times}10^{-2}}{3.15{\times}10^{-2}} & \err{2.06{\times}10^{-2}}{3.48{\times}10^{-2}} & \err{2.39{\times}10^{-2}}{4.21{\times}10^{-2}} & \err{2.41{\times}10^{-2}}{4.37{\times}10^{-2}} \\
6 & \besterr{1.76{\times}10^{-3}}{5.07{\times}10^{-4}} & \err{1.99{\times}10^{-2}}{1.78{\times}10^{-2}} & \err{1.50{\times}10^{-2}}{1.92{\times}10^{-2}} & \err{1.66{\times}10^{-2}}{2.34{\times}10^{-2}} & \err{2.11{\times}10^{-2}}{3.72{\times}10^{-2}} \\
7 & \besterr{1.71{\times}10^{-3}}{5.14{\times}10^{-4}} & \err{2.90{\times}10^{-2}}{3.40{\times}10^{-2}} & \err{2.60{\times}10^{-2}}{4.68{\times}10^{-2}} & \err{2.85{\times}10^{-2}}{4.96{\times}10^{-2}} & \err{3.11{\times}10^{-2}}{4.98{\times}10^{-2}} \\
8 & \besterr{1.73{\times}10^{-3}}{4.99{\times}10^{-4}} & \err{2.17{\times}10^{-2}}{2.57{\times}10^{-2}} & \err{1.90{\times}10^{-2}}{3.63{\times}10^{-2}} & \err{2.02{\times}10^{-2}}{3.78{\times}10^{-2}} & \err{2.25{\times}10^{-2}}{4.50{\times}10^{-2}} \\
\hline
\end{tabular}}
\end{table}

\FloatBarrier

\begin{figure*}[!htbp]
\centering

\begin{subfigure}[t]{0.5\textwidth}
    \centering
    \includegraphics[width=\linewidth]{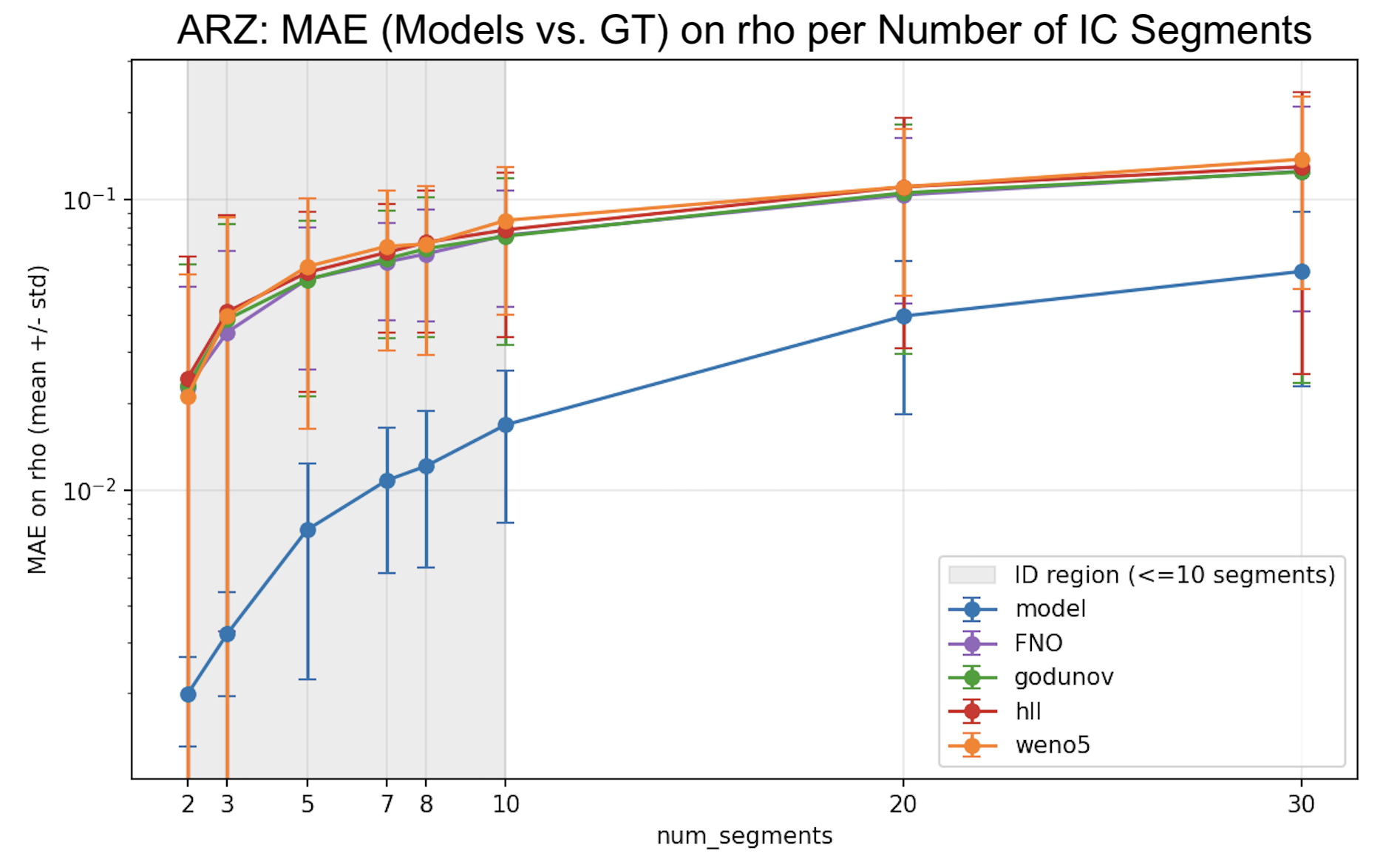}
    \caption{\texttt{Density $\rho$}}
    \label{fig:arz_mae_pwc_rho}
\end{subfigure}
\begin{subfigure}[t]{0.5\textwidth}
    \centering
    \includegraphics[width=\linewidth]{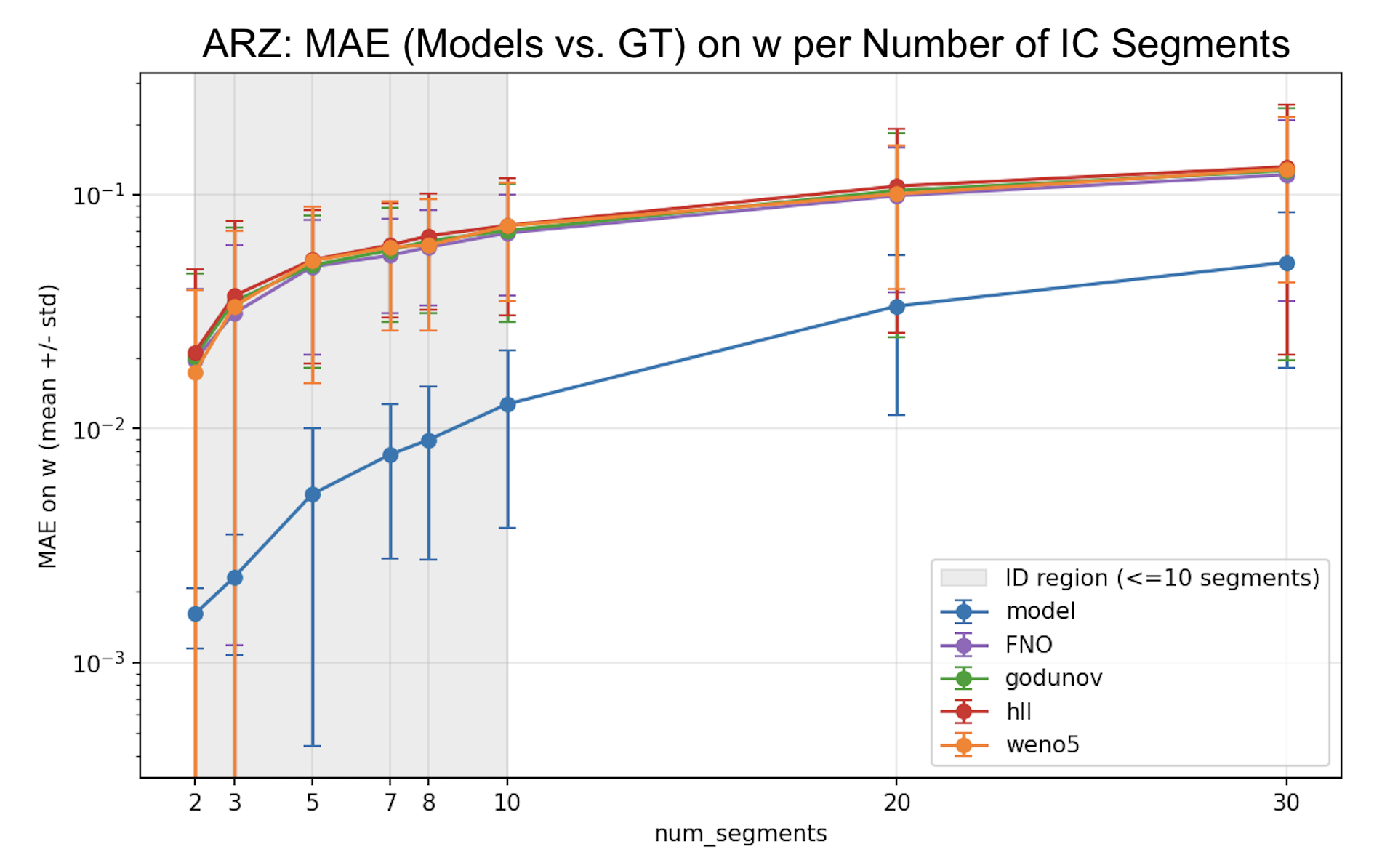}
    \caption{\texttt{Lagrangian marker $\omega$}}
    \label{fig:arz_mae_pwc_w}
\end{subfigure}

\caption{ARZ: $\rho$ and $\omega$ - mean absolute error on the $\rho$ and $\omega$ fields is plotted agains the initial condition complexity, averaged over the number of samples per bin, for our model, FNO, WENO, HLL and Godunov schemes. Gray area marks segment numbers present in the training set. HypNO-ARZ accuracy stays well below the baseline across the whole complexity range, and it's error grows more slowly in the out-of-distribution region.}
\label{fig:arz_mark0_mae_vs_seg}
\end{figure*}

\FloatBarrier

\begin{figure*}[!htbp]
  \centering

\begin{subfigure}[t]{0.95\textwidth}
  \centering
  \includegraphics[width=0.85\linewidth,height=0.24\textheight,keepaspectratio]{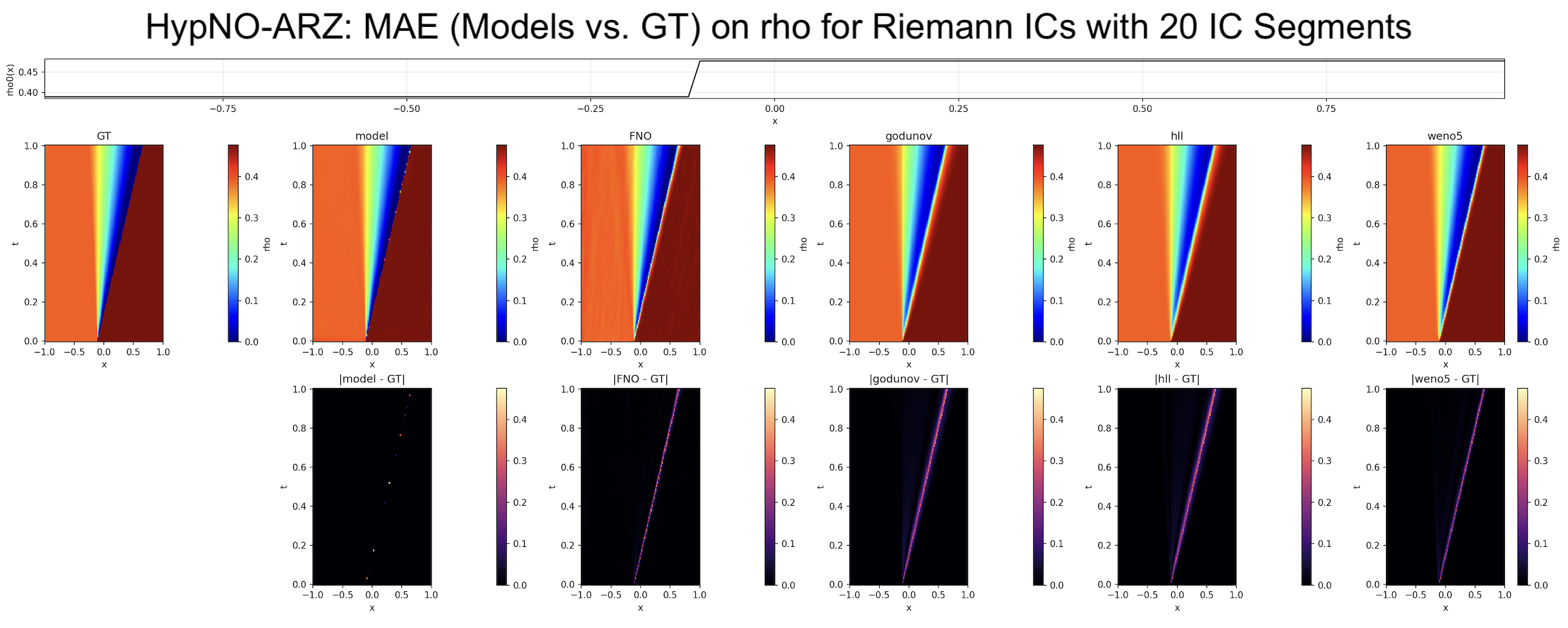}
  \caption{Density $\rho$: predicted fields (top) and mean absolute error compared to the ground truth (bottom).}
  \label{fig:arz_wft_riemann_rho}
\end{subfigure}

\vspace{0.25em}

\begin{subfigure}[t]{0.95\textwidth}
  \centering
  \includegraphics[width=0.85\linewidth,height=0.24\textheight,keepaspectratio]{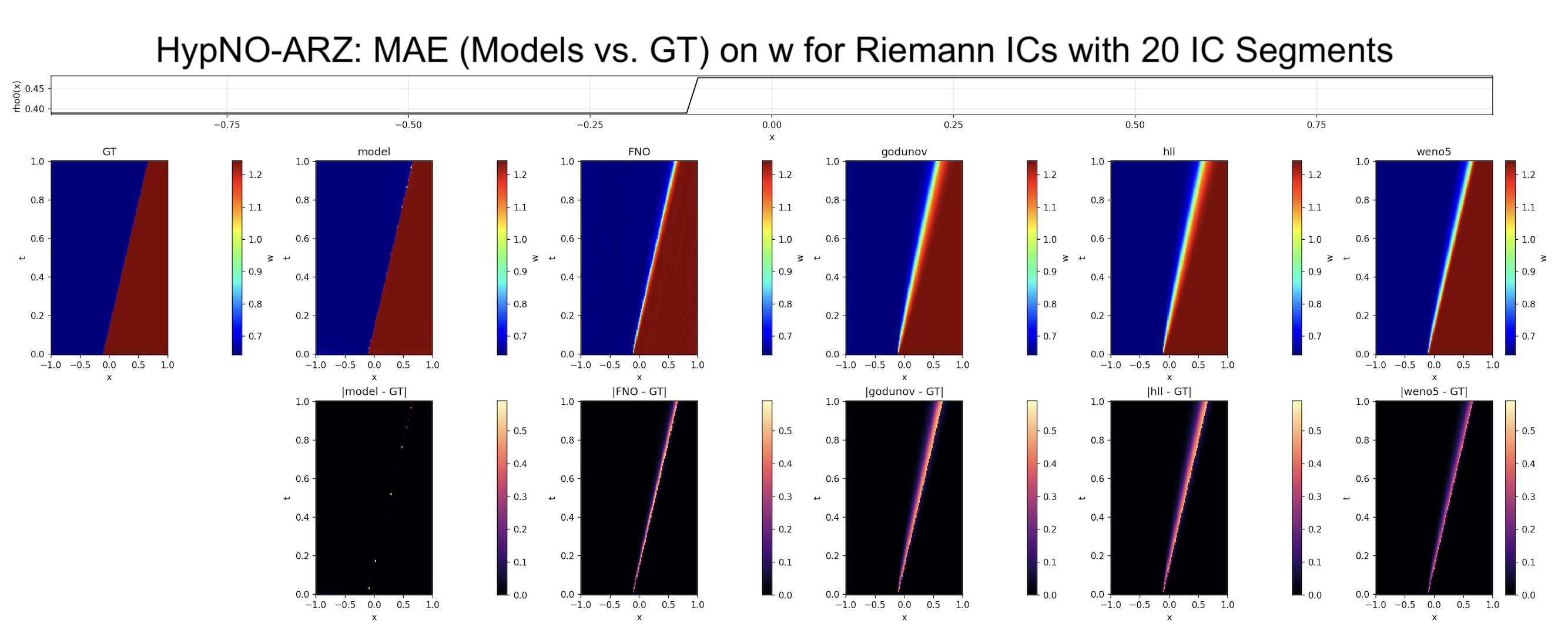}
  \caption{Lagrangian marker $\omega$: predicted fields (top) and mean absolute error compared to ground truth (bottom).}
  \label{fig:arz_wft_riemann_w}
\end{subfigure}

\caption{In-distribution \texttt{riemann\_stratified} sample (initial condition partitioned into 2 segments) on the
WFT reference: HypNO-ARZ vs.\ FNO, Godunov and HLL. The learned model keeps the
shock and contact sharp, whereas the finite volume baselines diffuse the
discontinuities into a multi-cell band.}
\label{fig:arz_wft_qual_riemann}
\end{figure*}

\begin{figure*}[!htbp]
\centering

\begin{subfigure}[t]{0.95\textwidth}
  \centering
  \includegraphics[width=0.85\linewidth,height=0.24\textheight,keepaspectratio]{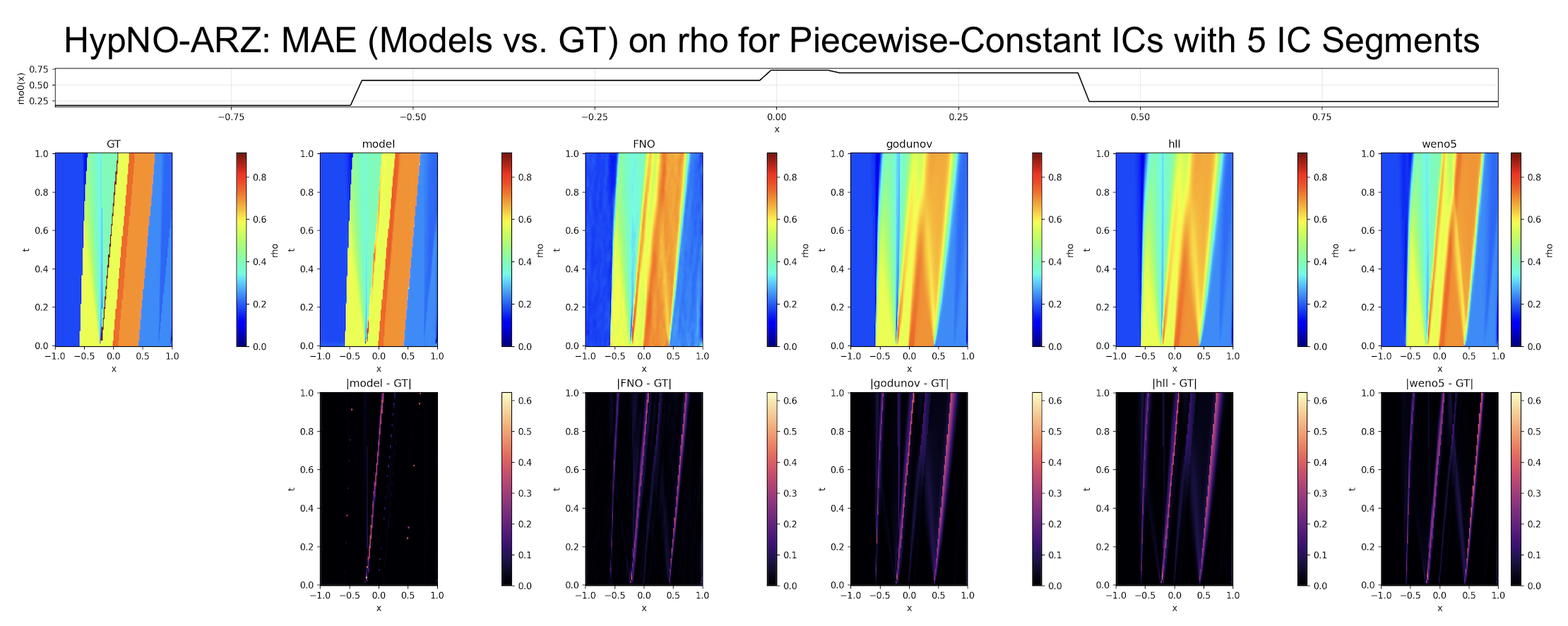}
  \caption{Density $\rho$: predicted fields (top) and mean absolute error compared to the ground truth (bottom).}
  \label{fig:arz_wft_pwc_rho}
\end{subfigure}

\vspace{0.25em}

\begin{subfigure}[t]{0.95\textwidth}
  \centering
  \includegraphics[width=0.85\linewidth,height=0.24\textheight,keepaspectratio]{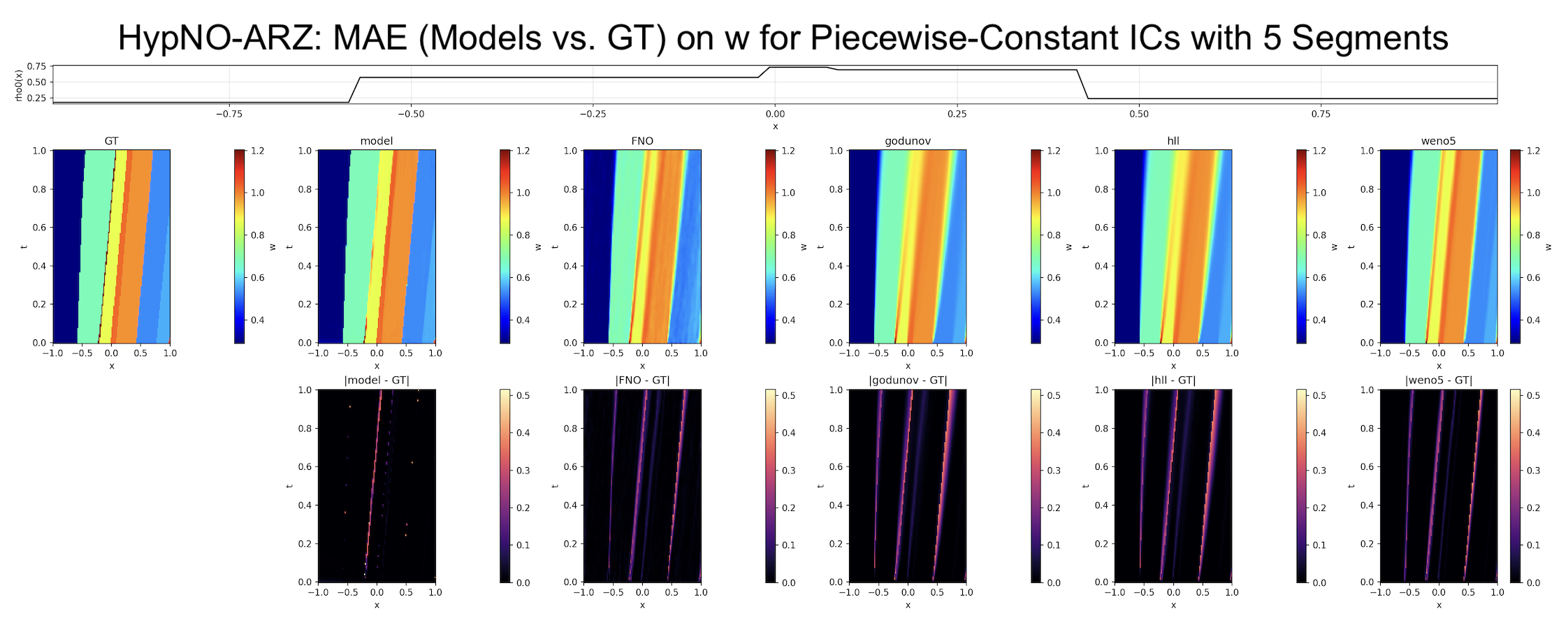}
  \caption{Lagrangian marker $\omega$: predicted fields (top) and mean absolute error compared to ground truth (bottom).}
  \label{fig:arz_wft_pwc_w}
\end{subfigure}

\caption{In-distribution \texttt{piecewise\_constant} sample
(initial condition partitioned into 5 segments): multiple interacting shocks and contacts. HypNO-ARZ resolves the
fronts more sharply than the diffusive Godunov and HLL references, and its error
($\mathrm{MAE}_w = 8.5\times10^{-3}$) concentrates on a few thin interfaces
rather than spreading across the wave structure.}
\label{fig:arz_wft_qual_pwc}
\end{figure*}

\begin{figure*}[!htbp]
\centering

\begin{subfigure}[t]{0.95\textwidth}
  \centering
  \includegraphics[width=0.85\linewidth,height=0.24\textheight,keepaspectratio]{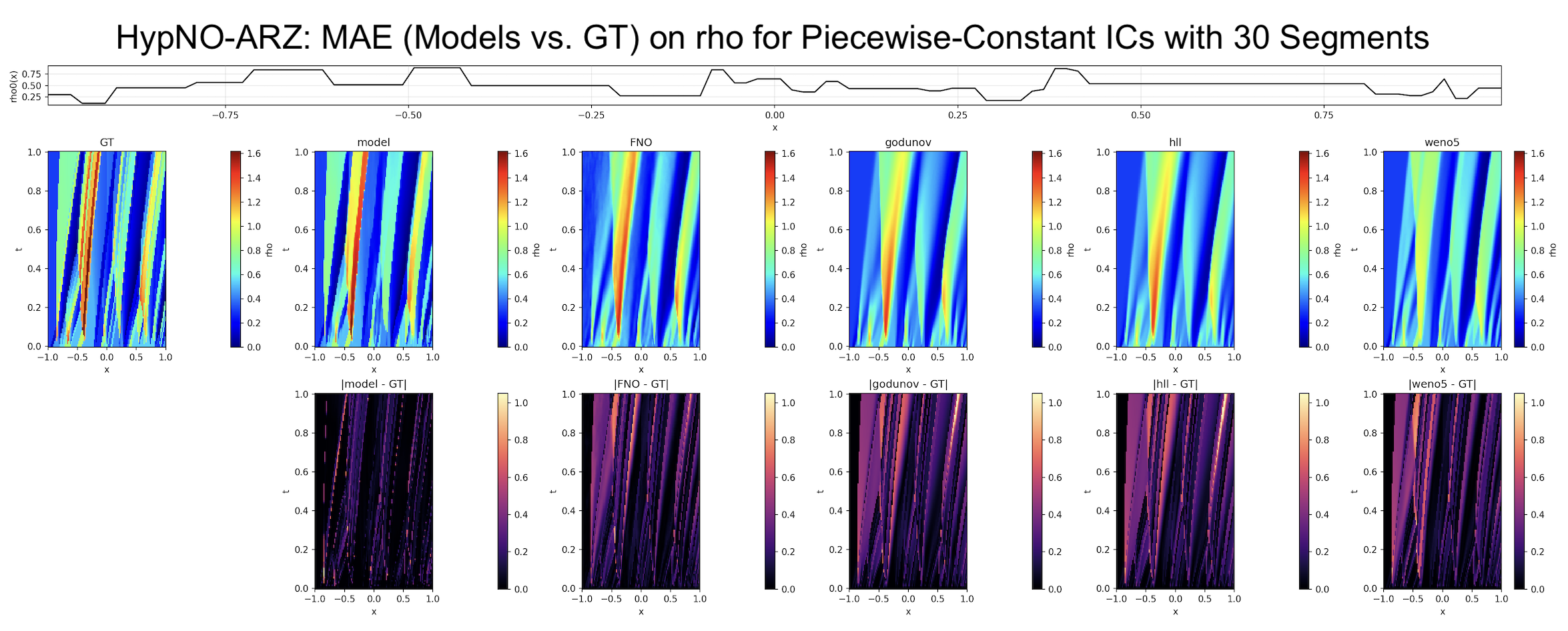}
  \caption{Density $\rho$: predicted fields (top) and mean absolute error compared to the ground truth (bottom).}
  \label{fig:arz_wft_const_rho}
\end{subfigure}

\vspace{0.25em}

\begin{subfigure}[t]{0.95\textwidth}
  \centering
  \includegraphics[width=0.85\linewidth,height=0.24\textheight,keepaspectratio]{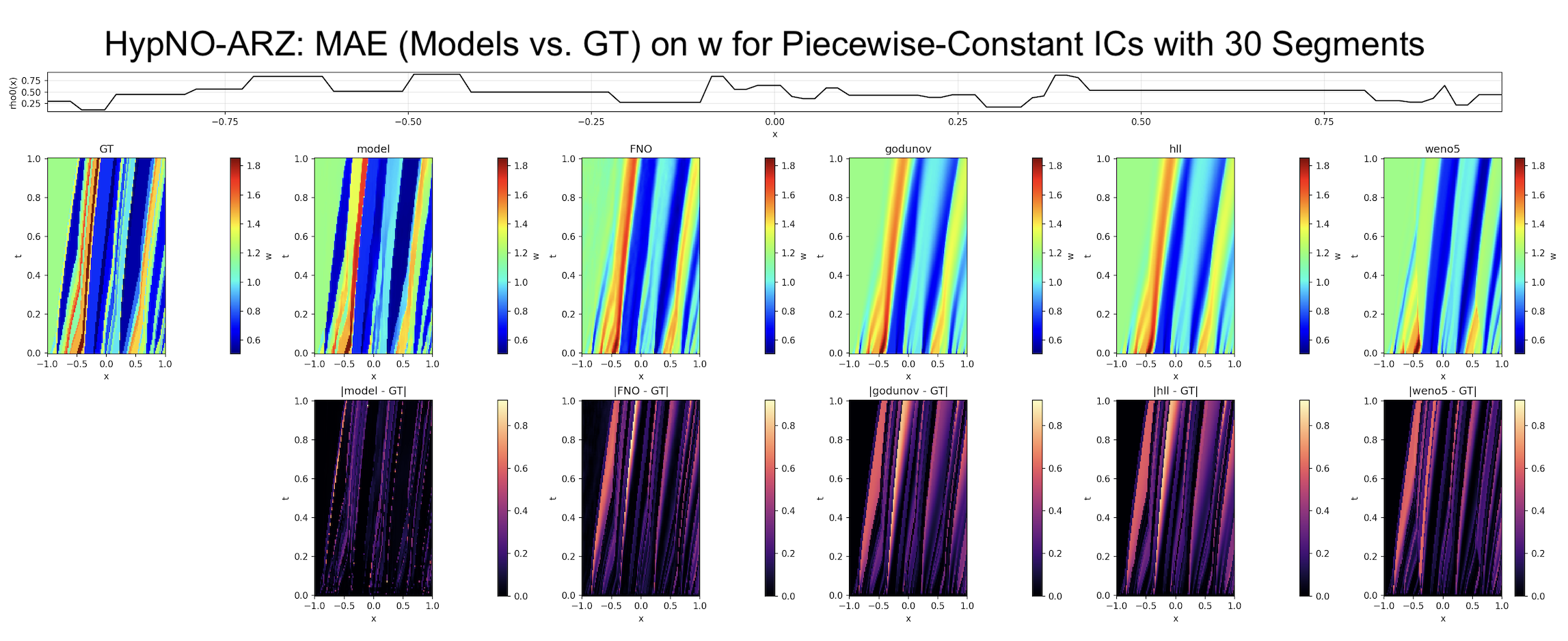}
  \caption{Lagrangian marker $\omega$: predicted fields (top) and mean absolute error compared to ground truth (bottom).}
  \label{fig:arz_wft_const_w}
\end{subfigure}
\caption{ARZ: Out-of-distribution \texttt{piecewise\_constant} sample (initial condition partitioned into 30 segments, well beyond the training range), evaluated against the wave-front-tracking ground truth. Panels show the density $\rho$ (a) and the Lagrangian marker $\omega$ (b), each with predicted space-time fields (top) and mean absolute error to the ground truth (bottom). Despite the interaction density, our model captures the shocks and resolves wave interactions way more sharply compared to the other methods.}.
\label{fig:arz_wft_qual_const}
\end{figure*}
\FloatBarrier

\paragraph{Shock neighborhood evaluation (ARZ)}
\label{sec:arz-shock-comparison}

To isolate accuracy on the discontinuous portion of the ARZ solution, we report MAE on $\rho$ restricted to two discontinuity bands detected on the ground truth and reused across all methods: a genuinely-nonlinear \emph{1-shock} band and a linearly-degenerate \emph{2-contact} band. Their detection is defined in \Cref{app:shock-band}.
\begin{center}
\captionof{table}{ARZ accuracy on shock neighborhoods ($\rho$ channel). \emph{full} is MAE over the whole space-time domain; \emph{1shock} and \emph{contact} restrict to the genuinely-nonlinear $1$-shock band (detected on the $\lambda_1$/$v$ field via the Lax condition $\lambda_{1,L}>\lambda_{1,R}$) and the $2$-contact band (detected on $\rho$ with $v$ continuity), both dilated to a fixed-width neighborhood. The same masks are reused across methods.}
\label{tab:arz-shock-mae}
\small
\resizebox{\textwidth}{!}{%
\begin{tabular}{cccccccccccccccc}
\toprule
num.\ seg. & \multicolumn{3}{c}{HypNO-ARZ} & \multicolumn{3}{c}{FNO} & \multicolumn{3}{c}{Godunov} & \multicolumn{3}{c}{HLL} & \multicolumn{3}{c}{WENO5} \\
\cmidrule(lr){2-4} \cmidrule(lr){5-7} \cmidrule(lr){8-10} \cmidrule(lr){11-13} \cmidrule(lr){14-16}
 & full & 1shock & contact & full & 1shock & contact & full & 1shock & contact & full & 1shock & contact & full & 1shock & contact \\
\midrule
2  & $\mathbf{1.91{\times}10^{-3}}$ & $\mathbf{9.98{\times}10^{-3}}$ & $\mathbf{8.43{\times}10^{-3}}$ & $2.39{\times}10^{-2}$ & $1.12{\times}10^{-1}$ & $1.25{\times}10^{-1}$ & $2.47{\times}10^{-2}$ & $1.17{\times}10^{-1}$ & $1.34{\times}10^{-1}$ & $2.63{\times}10^{-2}$ & $1.26{\times}10^{-1}$ & $1.42{\times}10^{-1}$ & $2.23{\times}10^{-2}$ & $1.12{\times}10^{-1}$ & $1.24{\times}10^{-1}$ \\
3  & $\mathbf{2.71{\times}10^{-3}}$ & $\mathbf{1.27{\times}10^{-2}}$ & $\mathbf{1.01{\times}10^{-2}}$ & $3.19{\times}10^{-2}$ & $1.24{\times}10^{-1}$ & $1.37{\times}10^{-1}$ & $3.44{\times}10^{-2}$ & $1.29{\times}10^{-1}$ & $1.46{\times}10^{-1}$ & $3.65{\times}10^{-2}$ & $1.39{\times}10^{-1}$ & $1.56{\times}10^{-1}$ & $3.43{\times}10^{-2}$ & $1.29{\times}10^{-1}$ & $1.42{\times}10^{-1}$ \\
5  & $\mathbf{5.44{\times}10^{-3}}$ & $\mathbf{2.06{\times}10^{-2}}$ & $\mathbf{1.55{\times}10^{-2}}$ & $4.54{\times}10^{-2}$ & $1.25{\times}10^{-1}$ & $1.39{\times}10^{-1}$ & $4.63{\times}10^{-2}$ & $1.27{\times}10^{-1}$ & $1.44{\times}10^{-1}$ & $4.96{\times}10^{-2}$ & $1.37{\times}10^{-1}$ & $1.53{\times}10^{-1}$ & $5.09{\times}10^{-2}$ & $1.35{\times}10^{-1}$ & $1.49{\times}10^{-1}$ \\
7  & $\mathbf{7.78{\times}10^{-3}}$ & $\mathbf{2.37{\times}10^{-2}}$ & $\mathbf{1.89{\times}10^{-2}}$ & $4.90{\times}10^{-2}$ & $1.22{\times}10^{-1}$ & $1.31{\times}10^{-1}$ & $5.13{\times}10^{-2}$ & $1.27{\times}10^{-1}$ & $1.37{\times}10^{-1}$ & $5.39{\times}10^{-2}$ & $1.35{\times}10^{-1}$ & $1.45{\times}10^{-1}$ & $5.43{\times}10^{-2}$ & $1.33{\times}10^{-1}$ & $1.41{\times}10^{-1}$ \\
8  & $\mathbf{8.65{\times}10^{-3}}$ & $\mathbf{2.36{\times}10^{-2}}$ & $\mathbf{1.94{\times}10^{-2}}$ & $5.15{\times}10^{-2}$ & $1.21{\times}10^{-1}$ & $1.30{\times}10^{-1}$ & $5.45{\times}10^{-2}$ & $1.29{\times}10^{-1}$ & $1.38{\times}10^{-1}$ & $5.77{\times}10^{-2}$ & $1.39{\times}10^{-1}$ & $1.48{\times}10^{-1}$ & $5.49{\times}10^{-2}$ & $1.31{\times}10^{-1}$ & $1.41{\times}10^{-1}$ \\
10 & $\mathbf{1.18{\times}10^{-2}}$ & $\mathbf{2.82{\times}10^{-2}}$ & $\mathbf{2.32{\times}10^{-2}}$ & $5.69{\times}10^{-2}$ & $1.17{\times}10^{-1}$ & $1.23{\times}10^{-1}$ & $5.70{\times}10^{-2}$ & $1.18{\times}10^{-1}$ & $1.26{\times}10^{-1}$ & $6.00{\times}10^{-2}$ & $1.27{\times}10^{-1}$ & $1.35{\times}10^{-1}$ & $6.36{\times}10^{-2}$ & $1.32{\times}10^{-1}$ & $1.39{\times}10^{-1}$ \\
20 & $\mathbf{2.71{\times}10^{-2}}$ & $\mathbf{4.60{\times}10^{-2}}$ & $\mathbf{4.07{\times}10^{-2}}$ & $7.88{\times}10^{-2}$ & $1.34{\times}10^{-1}$ & $1.43{\times}10^{-1}$ & $8.11{\times}10^{-2}$ & $1.35{\times}10^{-1}$ & $1.43{\times}10^{-1}$ & $8.51{\times}10^{-2}$ & $1.43{\times}10^{-1}$ & $1.51{\times}10^{-1}$ & $8.42{\times}10^{-2}$ & $1.44{\times}10^{-1}$ & $1.52{\times}10^{-1}$ \\
30 & $\mathbf{3.83{\times}10^{-2}}$ & $\mathbf{5.91{\times}10^{-2}}$ & $\mathbf{5.38{\times}10^{-2}}$ & $9.07{\times}10^{-2}$ & $1.41{\times}10^{-1}$ & $1.45{\times}10^{-1}$ & $9.09{\times}10^{-2}$ & $1.40{\times}10^{-1}$ & $1.44{\times}10^{-1}$ & $9.45{\times}10^{-2}$ & $1.47{\times}10^{-1}$ & $1.51{\times}10^{-1}$ & $9.93{\times}10^{-2}$ & $1.52{\times}10^{-1}$ & $1.55{\times}10^{-1}$ \\
\bottomrule
\end{tabular}%
}
\end{center}

\begin{center}
  \includegraphics[width=\linewidth]{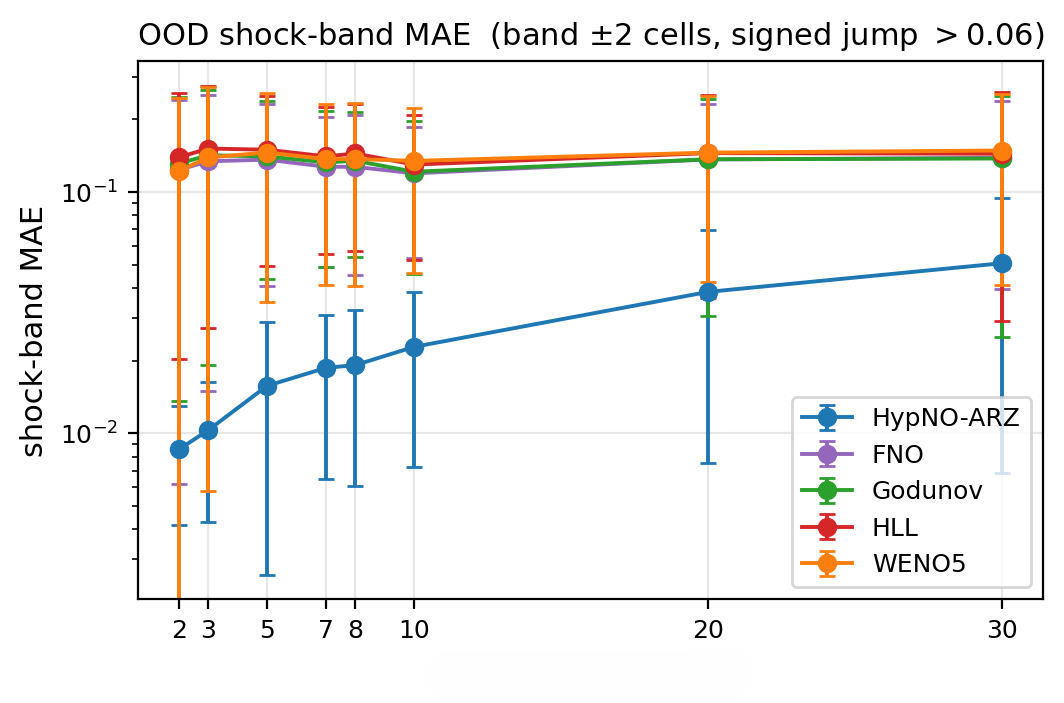}
  \captionof{figure}{ARZ: MAE restricted to the shock band, as a function of the number of
    discontinuities in the initial condition. HypNO-ARZ is compared against WENO5, Godunov, HLL, and the FNO
    baseline, all scored on the same WFT ground truth. HypNO-ARZ keeps the
    in-band error well below every baseline across the full complexity range.}
  \label{fig:shock-mae-vs-seg-arz}
\end{center}
\FloatBarrier
\begin{figure}[!htbp]
  \centering
  \includegraphics[width=0.9\linewidth]{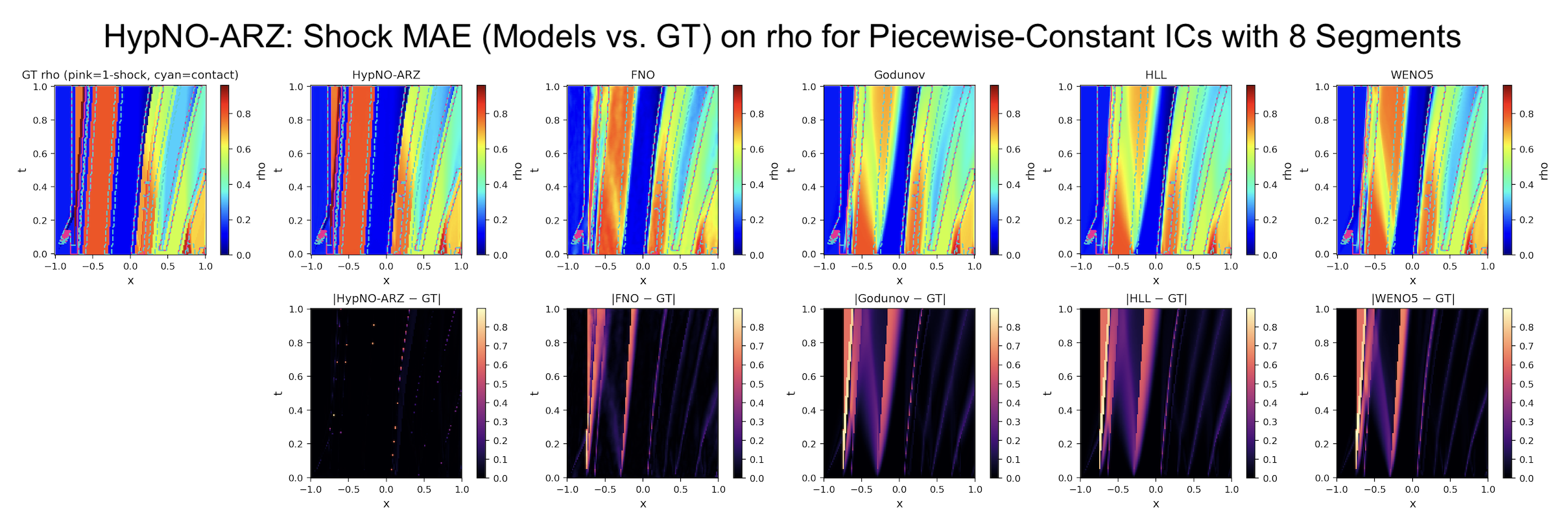}
  \caption{ARZ: Representative sample for an initial condition partitioned into $N=8$ segments.  Top row: $\rho$ fields with the 1-shock band (pink) and contact band (cyan dashed) outlined; bottom row: absolute error to the ground truth.}
  \label{fig:arz-shock-compare-8}
\end{figure}
\begin{figure}[!htbp]
  \centering
  \includegraphics[width=0.9\linewidth]{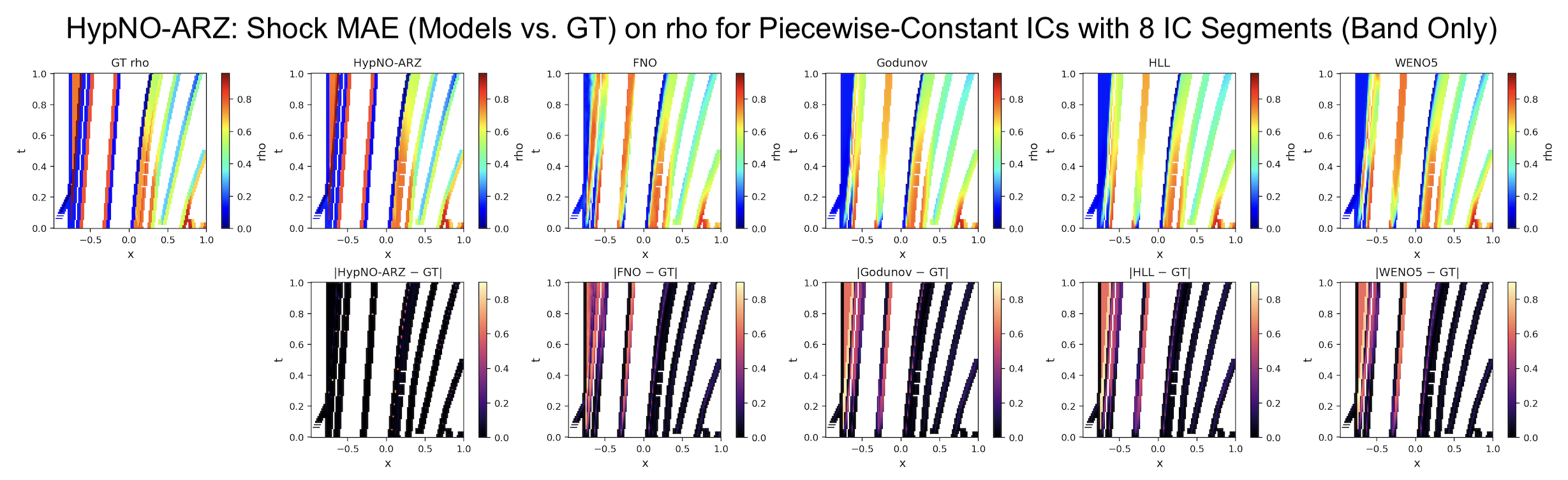}
  \caption{Same ARZ representative sample as Fig.~\ref{fig:arz-shock-compare-8}, zoomed to the combined-band bounding box with non-band cells blanked.}
  \label{fig:arz-shock-zoom-8}
\end{figure}
\begin{figure}[!htbp]
  \centering
  \includegraphics[width=0.8\linewidth]{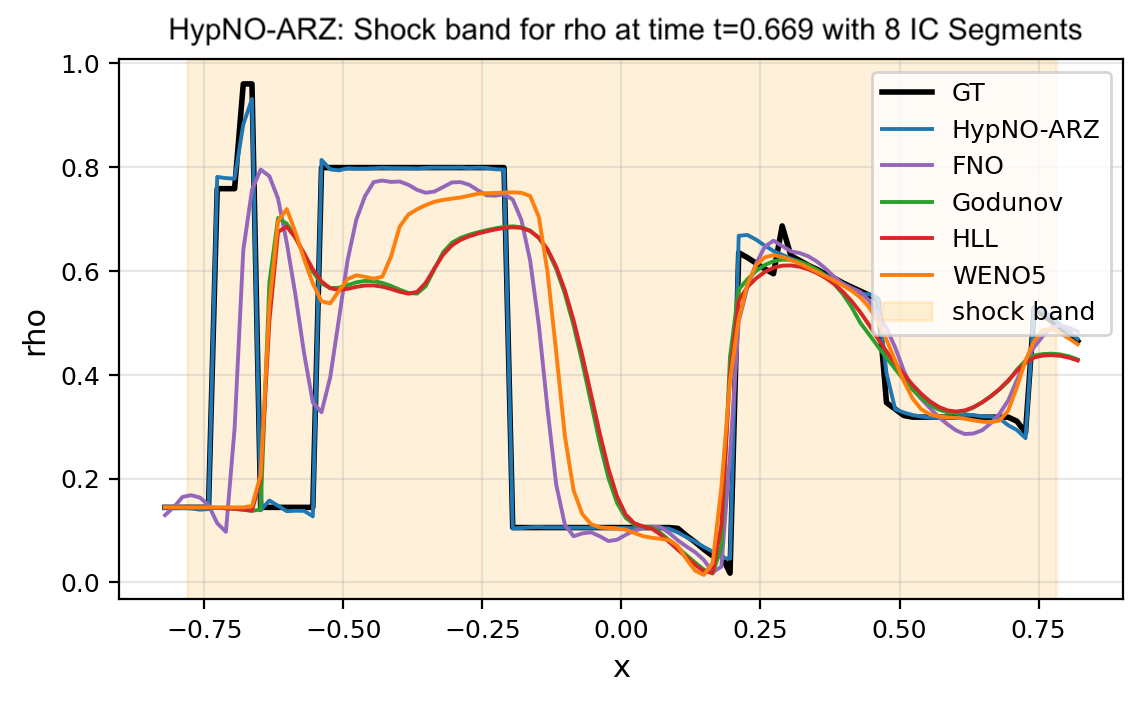}
  \caption{$\rho(x)$ at $t=0.669$ through a band row for ARZ model with num\_segments=8 initial discontinuity segments. The shaded region marks the combined shock band.}
  \label{fig:arz-shock-slice-8}
\end{figure}
\begin{figure}[!htbp]
  \centering
  \includegraphics[width=0.9\linewidth]{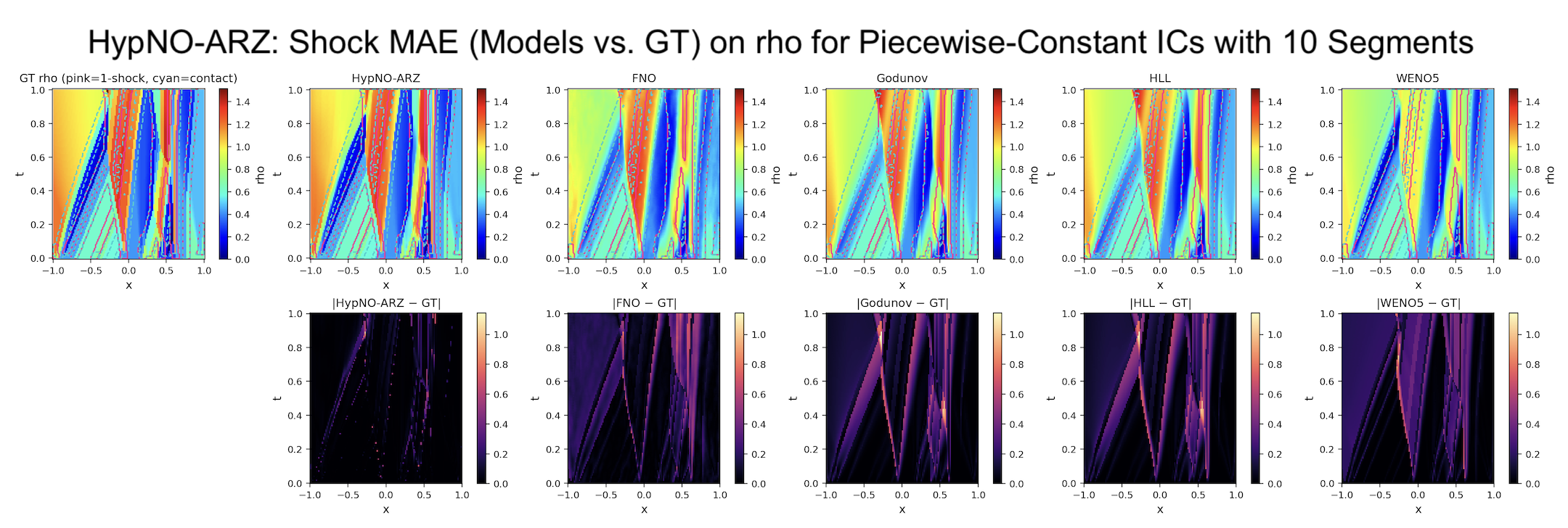}
  \caption{ARZ: Representative sample for an initial condition partitioned into $N=10$ segments. Top row: $\rho$ fields with the 1-shock band (pink) and contact band (cyan dashed) outlined; bottom row: absolute error to the ground truth.}
  \label{fig:arz-shock-compare-10}
\end{figure}
\begin{figure}[!htbp]
  \centering
  \includegraphics[width=0.9\linewidth]{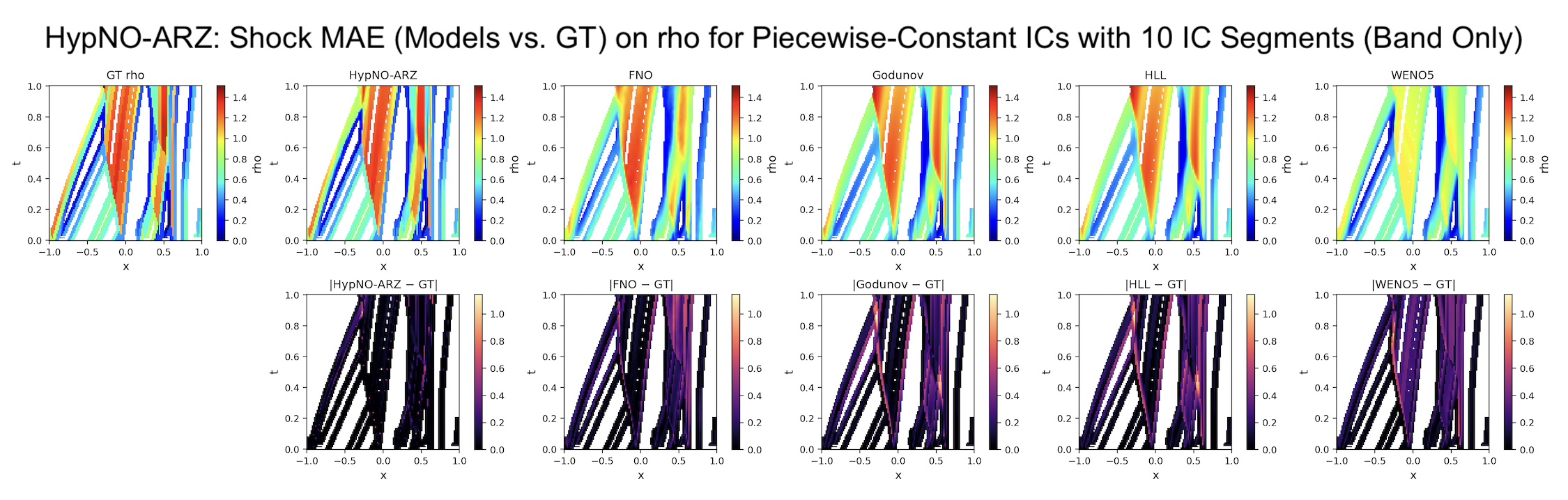}
  \caption{Same ARZ representative sample as Fig.~\ref{fig:arz-shock-compare-10}, zoomed to the combined-band bounding box with non-band cells blanked.}
  \label{fig:arz-shock-zoom-10}
\end{figure}
\begin{figure}[!htbp]
  \centering
  \includegraphics[width=0.8\linewidth]{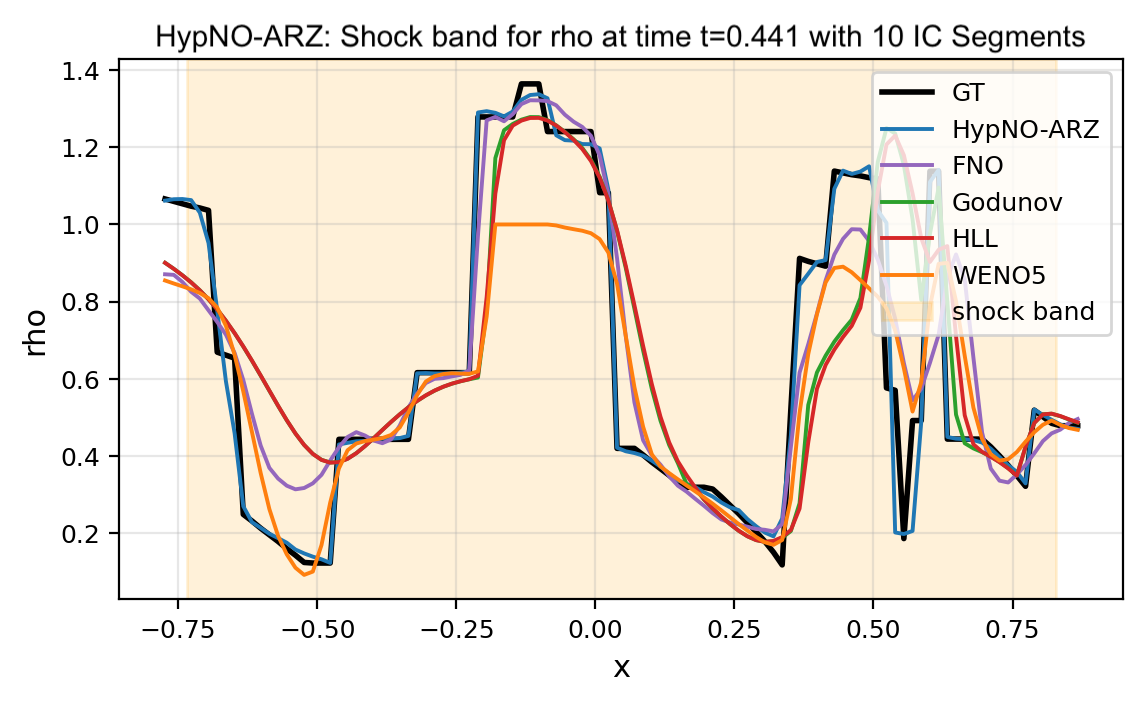}
  \caption{$\rho(x)$ at $t=0.441$ through a band row for ARZ model with num\_segments=10 initial discontinuity segments. The shaded region marks the combined shock band.}
  \label{fig:arz-shock-slice-10}
\end{figure}


\section{Conclusion}
\label{sec:conclusion}

We introduced HypNO, a hyperbolic neural operator for one-dimensional conservation laws that learns a map from initial data to the full space-time solution in a single forward pass.
The architecture lifts the initial condition onto a causal space-time grid, refines latent states through physics-gated message passing with finite-volume edge features, and decodes a physical field at every layer under deep supervision.
Stratified benchmarks on the LWR scalar model and the ARZ system, against exact Lax--Hopf and wave-front-tracking references, classical finite-volume solvers (WENO5, Godunov, HLL), and the FNO operator baseline, show that this design preserves shock and contact structure and delivers the best reported accuracy on discontinuity-dominated initial data, with favorable inference cost relative to sequential numerically structured learned integrators.

Several directions remain open.
First, computational complexity must be improved for higher-dimensional problems: the present one-dimensional cell graph and fixed stencil width scale favorably in amortized inference, but extending HypNO to two- and three-dimensional domains will require careful control of message-passing depth, edge sparsity, and memory as the number of mesh entities grows.
Second, the neighborhood strategy can be refined using the domain of dependence of hyperbolic solutions: although causal time connectivity and the receptive-field sizing in \Cref{app:reach} already encode a necessary coverage condition, future work should adapt stencils dynamically to local wave speeds and domains of influence rather than relying on a fixed symmetric spatial offset $k_x$. Finally, coupling HypNO with adaptive mesh refinement, richer multi-family training data, and hybrid reference generation (exact Riemann solvers and wave-front tracking on challenging initial conditions) offers a path toward robust operators for engineering-scale hyperbolic systems beyond the traffic-flow models studied in this work.

\clearpage
\appendix

\section{Notation}
\label{app:notation}

\begin{longtable}{p{3.4cm} p{9cm}}
\toprule
\textbf{Symbol} & \textbf{Description} \\
\midrule
\endhead

\multicolumn{2}{l}{\textit{Domain and grids}} \\[2pt]
$\Omega = [x_{\min}, x_{\max}]$ & Bounded one-dimensional spatial domain \\
$T$ & Final time; solutions are computed on $t\in[0,T]$ \\
$n_x$, $n_t$ & Number of spatial cells and temporal snapshots \\
$\Delta x$, $\Delta t$ & Uniform spatial and temporal grid spacings \\
$x_i$ & Spatial grid point, $x_i = x_{\min} + i\,\Delta x$ \\
$t_n$ & Temporal grid point, $t_n = n\,\Delta t$ \\
$(i,n)$ & Space-time node at position $x_i$ and time $t_n$ \\[6pt]

\multicolumn{2}{l}{\textit{LWR: solution and initial data}} \\[2pt]
$\rho(x,t) \in [0,1]$ & Traffic density (LWR conserved variable) \\
$\rho_{0,i}=\rho_0(x_i)$ & Initial density at cell $i$ (time-stacked to all
$t_n$ before encoding) \\
$\hat\rho_{(i,n)}^{(\ell)}$ & Decoded density probe at layer $\ell$, node
$(i,n)$ \\
$\hat\rho_{(i,n)}$ & Final decoded density after layer $L$ \\[6pt]

\multicolumn{2}{l}{\textit{LWR: flux and characteristic speeds}} \\[2pt]
$f(\rho)=\rho(1-\rho)$ & Greenshields flux \\
$\lambda(\rho)=f'(\rho)=1-2\rho$ & Characteristic speed \\
$f_{0,i}$, $\lambda_{0,i}$ & Flux and characteristic speed evaluated on the
broadcast initial datum $\rho_{0,i}$ \\
$s_{(j,m),(i,n)}$ & Rankine--Hugoniot interface speed on directed edge
$(j,m)\to(i,n)$ \\
$\xi_{(i,n)} = x_i/\max(t_n,\varepsilon_t)$ & Self-similarity variable in
lifting node features \\[6pt]

\multicolumn{2}{l}{\textit{ARZ: primitive and conservative variables}} \\[2pt]
$\rho(x,t)$, $v(x,t)$ & Density and velocity \\
$\omega = v + p(\rho)$ & Lagrangian marker (1-Riemann invariant); with
$p(\rho)=\rho$, $\omega=v+\rho$ \\
$y=\rho\omega$ & Conservative momentum variable \\
$p(\rho)$, $p'(\rho)$ & Traffic pressure and its derivative \\
$\lambda_1 = v-\rho\,p'(\rho)$, $\lambda_2=v$ & First- and second-family
characteristic speeds \\
$(\rho_{0,i},\omega_{0,i})$ & Time-stacked initial marker data at cell $i$ \\
$\hat\rho_{(i,n)}^{(\ell)}$, $\hat\omega_{(i,n)}^{(\ell)}$ & Decoded probes at
layer $\ell$; $\hat v_{(i,n)}=\hat\omega_{(i,n)}-p(\hat\rho_{(i,n)})$ \\[6pt]

\multicolumn{2}{l}{\textit{Neighborhoods}} \\[2pt]
$\mathcal{N}_{(i,n)}^{(k_x,k_t)}$ & Causal space-time product stencil: nodes
$(i+p,n+q)$ with $p\in\{-k_x,\ldots,k_x\}$, $q\in\{-k_t,\ldots,0\}$ \\
$k_x$, $k_t$ & Spatial and temporal neighborhood half-widths \\
$\mathcal{A}(i,n)$ & Adjacent edges at the same time level ($m=n$, $|j-i|=1$)
\\[6pt]

\multicolumn{2}{l}{\textit{Latent field and network dimensions}} \\[2pt]
$d$ & Latent dimension \\
$L$ & Number of message-passing layers ($\ell=1,\ldots,L$) \\
$h_{(i,n)}^{(\ell)}\in\mathbb{R}^d$ & Latent vector at layer $\ell$, node
$(i,n)$ \\
$h_{(i,n)}^{(0)}$ & Initial latent field after the lifting layer \\
$h_{(i,n)}^{\mathrm{node}}$ & Node embedding from the lifting encoder \\[6pt]

\multicolumn{2}{l}{\textit{Network modules}} \\[2pt]
$\mathrm{MLP}_{\mathrm{node}}$, $\mathrm{MLP}_{\mathrm{upd}}$ &
Node encoder and latent update in the lifting layer \\
$\mathrm{MLP}_{\mathrm{adj}}$, $\mathrm{MLP}_{\mathrm{non\text{-}adj}}$ &
Adjacent- and non-adjacent-edge message MLPs \\
$\mathrm{MLP}_{\mathrm{dec}}$ & Shared decoder mapping $h_{(i,n)}^{(\ell)}$ to
physical probes \\
$\mathbf{W}\in\mathbb{R}^{d\times d}$ & Residual linear map in the latent
update \\[6pt]

\multicolumn{2}{l}{\textit{Directed edges and jumps}} \\[2pt]
$(j,m)\to(i,n)$ & Directed edge from neighbor $(j,m)$ to source $(i,n)$ \\
$r_{(j,m),(i,n)}=\operatorname{sign}(x_j-x_i)$ & Signed spatial displacement
\\
$\Delta\rho_{(j,m),(i,n)}$, $\Delta v_{(j,m),(i,n)}$,
$\Delta\omega_{(j,m),(i,n)}$ & Probe jumps across the edge (ARZ) \\
$\Delta\rho_{(j,m),(i,n)}$ & Density probe jump across the edge (LWR) \\
$\Delta x^{\mathrm{feat}}_{(j,m),(i,n)}=(x_j-x_i)/\Delta x$,
$\Delta t^{\mathrm{feat}}_{(j,m),(i,n)}=(t_m-t_n)/\Delta t$ &
Grid-normalized space--time offsets on non-adjacent edges \\[6pt]

\multicolumn{2}{l}{\textit{Messages and aggregates}} \\[2pt]
$m_{(j,m),(i,n)}^{(\ell)}$ & Message on edge $(j,m)\to(i,n)$ at layer $\ell$ \\
$M_{(i,n)}^{(\ell)}$ & Gate-normalized message aggregate at node $(i,n)$ \\
$M_{(i,n)}^{(0)}$ & Lifting-layer aggregate feeding $h_{(i,n)}^{(0)}$ \\[6pt]

\multicolumn{2}{l}{\textit{Physics gates}} \\[2pt]
$g_{(j,m),(i,n)}^{(\ell)}\in[0,1]$ & Scalar edge gate weighting a message \\
$g_{(j,n),(i,n)}^{\mathrm{up}}$ & Upwind gate (LWR; ARZ combines
$g^{\mathrm{up},1}$, $g^{\mathrm{up},2}$) \\
$g_{ij}^{\mathrm{entropy}}$, $g_{(j,n),(i,n)}^{\mathrm{ent}}$ &
Entropy gate on adjacent edges \\
$g_{ij}^{\mathrm{CFL}}$ & Temporal CFL gate on edges with $m<n$ \\
$g_{(j,m),(i,n)}^{\mathrm{lift}}$ & Gate in the lifting layer (upwind $\times$
entropy on adjacent edges, unity otherwise) \\
$\chi_{(j,n),(i,n)}^{\mathrm{up}}$ &
Hard upwind flag, $\mathbb{1}_{\{s_{(j,n),(i,n)}(x_j-x_i)<0\}}$ (LWR) \\
$\chi_{ij}^{\mathrm{entropy}}$ &
LWR entropy-violation indicator (\Cref{lwr-ent-gate}) \\
$\chi_{(j,n),(i,n)}^{\mathrm{sh}}$ &
ARZ 1-shock indicator, $\mathbb{1}_{\{\lambda_{1,L}>\lambda_{1,R}\}}$ \\
$\theta_{(j,m),(i,n)}$ & ARZ router weight between 1-wave and 2-contact
jumps (\Cref{eq:router-weight}) \\[6pt]

\multicolumn{2}{l}{\textit{Learnable gate parameters}} \\[2pt]
$\theta_\tau$, $\theta_\gamma$, $\theta_\kappa$ & Raw parameters for upwind
temperature $\tau$, entropy floor $\gamma$, and CFL decay $\kappa$ \\
$\theta_{\tau,m}$ & Per-family upwind temperature in ARZ, $m\in\{1,2\}$ \\
$\tau=\operatorname{softplus}(\theta_\tau)+\varepsilon$,
$\gamma=\operatorname{sigmoid}(\theta_\gamma)$,
$\kappa=\operatorname{softplus}(\theta_\kappa)$ & Mapped gate hyperparameters
\\[6pt]

\multicolumn{2}{l}{\textit{Constants}} \\[2pt]
$\varepsilon=10^{-6}$ & Numerical floor in gate normalization and divisions
\\
$\varepsilon_t$ & Floor in the self-similarity denominator
$\max(t_n,\varepsilon_t)$ \\
$\delta$ & Small density-jump threshold in ARZ Rankine--Hugoniot fallback
(\Cref{eq:arz-orig-s1}) \\[6pt]

\bottomrule
\end{longtable}

\section{Network design details}
\label{app:design}
This appendix collects the architectural details summarized at a high level in
\Cref{sec:scheme}: the explicit lifting-layer equations and the
PDE-specific node features, edge feature vectors, and physics gates for the LWR
and ARZ models.

\subsection{Neighborhood}
\label{app:neighborhood}
The neighborhood $\mathcal{N}_{(i,n)}^{(k_x,k_t)}$ is the collection of nodes
that the source node $(i,n)$ has edge connections with, sized by the spatial and
temporal offsets $k_x$ and $k_t$:
\begin{equation}
\label{neighborhood}
\mathcal{N}_{(i, n)}^{(k_x,k_t)}
=
\left\{
(i+p,n+q)
\;\middle|\;
p \in \{-k_x,\ldots,k_x\},\;
q \in \{-k_t,\ldots,0\},\;
x_{i+p}\in\Omega,\;
t_{n+q}\in[0,\timehor]
\right\}.
\end{equation}
It should be noted that $q\in\{-k_t,\ldots,0\}$ shows that the communication happens only with the same or earlier-time nodes. Here, $\Omega$ is the spatial domain. 


\subsection{Per-PDE design: LWR}
\label{app:lwr-design}
\subsubsection{Lifting layer}
  \label{app:lwr-lifting}
  The initial condition is broadcast across all time levels before encoding, so
  every space-time node $(i,n)$ carries the physical state of its spatial
  location at $t=0$; all lifting features below are computed from these
  broadcast states. Fixing the node $(i,n)$, our configuration uses a
  $5$-channel node input
  \begin{equation}
    \mathbf{x}^{\text{node}}_{(i,n)} \;=\;
    \bigl(\,\rho_{0,i},\; x_i,\; t_n,\; f(\rho_{0,i}),\; \lambda(\rho_{0,i})\,\bigr),
    \qquad
    f(\rho) = \rho(1-\rho), \quad \lambda(\rho) = f'(\rho) = 1 - 2\rho,
    \label{eq:lwr-node-feats}
  \end{equation}
  where $\rho_{0,i}=\rho_0(x_i)$ is the time-stacked initial datum at cell $i$ and
  $f$, $\lambda$ are the LWR flux and characteristic speed evaluated on it. A
  node-wise encoder embeds this vector into a node embedding,
  \begin{equation}
  h^{\mathrm{node}}_{(i,n)} = \mathrm{MLP}_{\mathrm{node}}\bigl(\mathbf{x}^{\text{node}}_{(i,n)}\bigr).
  \end{equation}

  The lifting is itself a gated message-passing layer over the same
  neighborhood $\mathcal{N}_{(i,n)}^{(k_x,k_t)}$ as the main message-passing
  layers. Unlike the ARZ lifting (\Cref{app:lifting}), which routes adjacent
  and non-adjacent edges through separate MLPs with distinct feature sets, the
  LWR lifting uses a \emph{single} message MLP over all stencil edges: every
  edge carries the same $8$-slot feature vector, with adjacency encoded as an
  explicit indicator and the interface slots zeroed where undefined,
  \begin{equation}
      e^{\mathrm{lift}}_{(j,m),(i,n)}
      =
      \Big(
      \Delta\rho_{(j,m),(i,n)},\,
      r_{(j,m),(i,n)},\,
      \Delta t^{\mathrm{feat}}_{(j,m),(i,n)},\,
      t_n,\,
      t_m,\,
      s_{(j,m),(i,n)},\,
      \mathbb{1}^{\mathrm{adj}}_{(j,m),(i,n)}
      \Big).
      \label{eq:lwr-lift-feats}
  \end{equation}
    Where $\Delta t^{\mathrm{feat}}_{(j,m),(i,n)} \Def (t_m-t_n)/\Delta t$.
  \begin{remark}
  In practice, we also added   $\operatorname{sign}\!\bigl(s_{(j,m),(i,n)}\bigr)$ to \Cref{eq:lwr-lift-feats}. 
  \end{remark}
  
On non-adjacent edges no
  interface orientation exists: the jump reverts to source-minus-target order,
  $\Delta\rho_{(j,m),(i,n)} = \rho_{0,j} - \rho_{0,i}$, and the interface slots close,
  $\lambda_{(j,m),(i,n)} = \operatorname{sign}(\lambda) = \mathbb{1}^{\mathrm{adj}} = 0$.
  Each feature vector is mapped to a latent message by the shared MLP,
  \begin{equation}
  m^{(0)}_{(j,m),(i,n)}
  =
  \mathrm{MLP}^{\mathrm{lift}}_{\mathrm{edge}}
    \bigl(e^{\mathrm{lift}}_{(j,m),(i,n)}\bigr).
  \label{eq:lwr-lift-msg}
  \end{equation}

  The aggregate $M^{(0)}_{(i,n)}$ is the gate-normalized sum of \Cref{eq:lwr-lift-msg} using 
  \Cref{E:aggregation}, with the scalar gate given by the LWR
  upwind--entropy gate of \Cref{lwr-up-gate,lwr-ent-gate} on adjacent edges
  and open otherwise,
  \begin{equation}
  g^{\mathrm{lift}}_{(j,m),(i,n)}
  =
  \begin{cases}
  g^{\mathrm{up}}_{(j,n),(i,n)}\; g^{\mathrm{ent}}_{(j,n),(i,n)}
    & (j,m)\in\mathcal{A}(i,n),\\[2pt]
  1 & \text{otherwise}.
  \end{cases}
  \label{eq:lwr-lift-gate}
  \end{equation}
  In fact, for the lifting layer we only calculate the gates for adjacent nodes. Non-adjacent nodes are considered to be fully communicative. 
  Finally, the node embedding and the aggregate are combined into the initial
  latent state,
  \begin{equation}
  h^{(0)}_{(i,n)}
  =
  \mathrm{MLP}_{\mathrm{upd}}
  \bigl(
  [\,h^{\mathrm{node}}_{(i,n)},\,M^{(0)}_{(i,n)}\,]
  \bigr),
  \end{equation}
  which is the input to the stack of message-passing layers, and carries a
  one-stencil receptive field.
\subsubsection{Edge feature vector design}
\label{lwr-edge-feats}
For adjacent edges, the message is computed using a dedicated adjacent-edge MLP:
\[
m_{(i,n),(j,n)}^{\mathrm{adj},(\ell)}
=
\MLP_{\mathrm{adj}}
\left(
h_{(i,n)}^{(\ell)},
h_{(j,n)}^{(\ell)},
e_{(j,n),(i,n)}^{\mathrm{adj},(\ell)}
\right), \quad j = i \pm 1
\]
and 
\[
e_{(j,n),(i,n)}^{\mathrm{adj},(\ell)}
=
\left[
s_{(j,n),(i,n)},
\operatorname{sign}(s_{(j,n),(i,n)}),
\chi_{(j,n),(i,n)}^{\mathrm{up}},
\operatorname{sign}(x_j-x_i)
\right].
\]
Here \(s_{(j,n),(i,n)}\) is the Rankine-Hugoniot interface speed, which in the case of shocks, reads
\[
s_{(j,n),(i,n)}
=
\frac{f(\hat \rho_{(j,n)})-f(\hat \rho_{(i,n)})}
{\hat \rho_{(j,n)}-\hat \rho_{(i,n)}},
\]
and if \(|\hat \rho_{(j,n)}-\hat \rho_{(i,n)}|\) is below a small numerical tolerance, we have 
\[
s_{(j,n),(i,n)}=f'(\hat \rho_{(i,n)})
\]
The quantity \(\hat \rho\) denotes the decoded state probe used to compute physics
scalars inside the message-passing layer.

The upwind indicator is defined as
\[
\chi_{(j,n),(i,n)}^{\mathrm{up}}
=\mathbb{1}_{\set{s_{(j,n),(i,n)}\cdot(x_j-x_i)<0}}.
\]
It indicates whether the node $(j,n)$ lies on the upwind side of 
node $(i,n)$. Thus, the edge features encode the local interface information needed to distinguish whether a neighboring cell is physically relevant for
the target update.

For the non-adjacent neighbors, values such as the Rankine-Hugoniot speed are not physically meaningful, so the feature vector has a simpler form:
\[
e_{(j,m), (i,n)}^{\mathrm{non-adj},(\ell)}
=
\left[
x_j-x_i,
t_m-t_n,
\operatorname{sign}(x_j-x_i)
\right].
\]
In this definition, $\operatorname{sign}(x_j - x_i)$ is included for the practical point of view. These features describe the relative space-time displacement of the source node
with respect to the target node. The non-adjacent message MLP therefore receives only topological information in addition to the two latent states. Therefore, for non-adjacent edges, the message is computed using a dedicated non-adjacent MLP:
\[
m_{(i,n),(j,n)}^{\mathrm{non-adj},(\ell)}
=
\MLP_{\mathrm{non-adj}}
\left(
h_{(i,n)}^{(\ell)},h_{(j,m)}^{(\ell)}, e_{(j,m),(i,n)}^{\mathrm{non-adj},(\ell)}
\right).
\]
Overall, this edge class distinction provides the model with two different message functions. The adjacent MLP is
specialized for local interface interactions, where hyperbolic conservation law structure is most directly expressed. The non-adjacent MLP is specialized for longer-range space-time information flow, where the relevant relation is mainly topological rather than an immediate numerical flux interface. 
\subsubsection{Upwind gate}
\label{lwr-up-gate}
The upwind gate encodes the direction of local wave propagation. For an
adjacent spatial edge $(j,n) \to (i,n)$, we estimate the local propagation speed using
the Rankine-Hugoniot formalism.

A message from $(j,n) \to (i,n)$ should be favored when $(j,n)$ lies in upwind direction of $(i,n)$. This can be expressed using the sign of the displacement from source to
target:
\[
r_{(j,n),(i,n)}=\operatorname{sign}(x_j-x_i).
\]
The upwind gate is then defined as
\[
g_{(j,n),(i,n)}^{\mathrm{up}}
=
\mathrm{sigmoid}
\left(
\frac{-s_{(j,n),(i,n)}r_{(j,n),(i,n)}}{\tau}
\right),
\qquad
\tau=\operatorname{softplus}(\theta_\tau)+\varepsilon.
\]
where, $\eps >0$ is a sufficiently small value. 

If the local wave speed points from $(j,n)$ toward $(i,n)$, then $s_{(j,n),(i,n)},r_{(j,n),(i,n)}<0$, so the sigmoid argument is positive and $g^{\mathrm{up}}_{(j,n),(i,n)}\approx 1$. If the wave propagates away from $(i,n)$, then $s_{(j,n),(i,n)},r_{(j,n),(i,n)}>0$ and the message is suppressed. The learnable temperature $\tau$ controls how sharp this selection is.
\subsubsection{Entropy gate}
\label{lwr-ent-gate}
The entropy gate is designed to attenuate messages across locally
entropy-violating discontinuities. Hyperbolic conservation laws admit weak
solutions that are not necessarily physically admissible and the entropy condition selects the physically relevant solution.

For an adjacent interface, let $\rho_{(i, n)}$ and $\rho_{(j,n)}$, $j >i$ denote the left and right
states. In the LWR PDE case considered here, the flux is $f(\rho)=\rho(1-\rho)$, which is concave. Hence a jump with $\rho_{(i, n)} <\rho_{(j, n)}$ corresponds to a shock, while
a jump with $\rho_{(i, n)}>\rho_{(j, n)}$ corresponds to a rarefaction-type configuration.

Let
\begin{equation}
\lambda_{(i, n)}=f'(\rho_{(i, n)}), \qquad \lambda_{(j, n)}=f'(\rho_{(j, n)}),
\end{equation}
and let
\begin{equation}
 \lambda_{i,j} = \frac{f(\rho_{(j, n)})-f(\rho_{(i, n)})}{\rho_{(j, n)}-\rho_{(i, n)}}
\end{equation}

denote the Rankine-Hugoniot speed. On a concave flux, an admissible Lax entropy condition satisfies
\begin{equation}
\lambda_{(i, n)} \ge \lambda_{ij} \ge \lambda_{(j, n)}.
\end{equation}
Consequently, we define the local entropy-violation indicator as 
\begin{equation}
\chi_{ij}^{\mathrm{entropy}}
=
\mathbb{1}_{\{\rho_{(i, n)} < \rho_{(j, n)}\}}
\,
\mathbb{1}_{\left\{
\lambda_{ij} > \lambda_{(i, n)} + \varepsilon
\;\;\vee\;\;
\lambda_{ij} < \lambda_{(j, n)} - \varepsilon
\right\}}, \quad \text{for} \quad \eps>0 \quad \text{sufficiently small}.
\end{equation}
In fact, $\chi_{ij}^{\mathrm{entropy}} = 1$ when the Lax entropy condition is violated. 

The entropy gate is then
\begin{equation}
g_{ij}^{\mathrm{entropy}}
= 1-\chi_{ij}^{\mathrm{entropy}}(1-\gamma),
\qquad
\gamma=\mathrm{sigmoid}(\theta_\gamma).
\end{equation}
In particular, to avoid shutting down the message passing between the nodes even in the case of entropy violation, we define a learnable parameter $\gamma \in [0, 1]$ with parameter $\theta_\gamma$. 

Equivalently,
\begin{equation}
g_{ij}^{\mathrm{entropy}}
= \begin{cases}
1,
& \text{for rarefaction-type interfaces},\\
1,
& \text{for admissible shocks}, \\
\gamma,
& \text{for entropy-violating shock-like interfaces.}
\end{cases}
\end{equation}
\subsubsection{Temporal gate}
Edges with a nonzero temporal offset ($t_n - t_{n+q}  \neq 0$), with $q$ defined in \Cref{app:neighborhood}, are weighted by the following soft
CFL-based gate. In particular, with the local characteristic speed
$\lambda_i = f'(\hat \rho_i) = 1 - 2\hat \rho_i$ evaluated on the decoded state, the CFL
number and gate are
\begin{equation}
  g^{\mathrm{CFL}}_{ij}
  = \exp\!\Bigl\{-\kappa\,\bigl(\mathrm{ReLU}(\mathrm{CFL}-1\bigr)^2\Bigr\},
  \qquad
  \mathrm{CFL}
  = \frac{\bigl(\max_{q \in \set{-k_t, \cdots, 0}}\max_{i \le k \le j} \abs{\lambda_{k,n+q}} \bigr) \Delta t}{\Delta x},
  \qquad
  \kappa = \mathrm{softplus}(\theta_\kappa),
  \label{eq:lwr-cfl-gate}
\end{equation}
so the corresponding message passing is weighted down once the CFL condition is larger than one--- the wave cannot travel the grid spacing $\Delta x$ within its temporal span---and is left open otherwise.

Unlike generic attention weights learned only from data, our gates are built
from quantities that are meaningful for scalar conservation laws---local wave
speed, upwind direction, entropy admissibility, and temporal distance. The gates
therefore act as soft inductive biases. They do not enforce a fixed numerical
scheme, but they guide the learned message passing toward physically plausible
information flow.

\subsection{Per-PDE design: ARZ} 
\label{sec:arz-design}
In this section, we explain the analogy of the LWR model in \Cref{app:lwr-design} for the ARZ model. 
\subsubsection{Lifting layer}
  \label{app:lifting}
The initial condition is broadcast across all
  time levels before encoding, so every space-time node $(i,n)$ carries the
  physical state of its spatial location at $t=0$; all lifting features below
  are computed from these broadcast states. Fixing the node $(i,n)$, our configuration uses a $7$-channel node input
  \begin{equation}
    \mathbf{x}^{\text{node}}_{(i,n)} \;=\;
    \bigl(\,\rho_{0,i},\; \omega_{0,i},\; v_{0,i},\; y_{0,i},\;
            x_i,\; t_n,\; \xi_{(i,n)}\,\bigr),
    \qquad
    y_{0,i} = \rho_{0,i}\,\omega_{0,i}, \quad
    v_{0,i} = \omega_{0,i} - p(\rho_{0,i}), \quad
    \xi_{(i,n)} = \frac{x_i}{\max(t_n,\varepsilon_t)},
    \label{eq:arz-orig-node-feats}
  \end{equation}
  where $\omega=v+p(\rho)$ is the ARZ Riemann invariant and $y=\rho\omega$ the
  conserved momentum variable (see \Cref{sec:arz}), and $\rho_{0,i}=\rho_0(x_i)$,
  $\omega_{0,i}=\omega_0(x_i)$ are the time-stacked initial data at cell $i$.
  A node-wise encoder embeds this vector into a node embedding,
  \begin{equation}
  h^{\mathrm{node}}_{(i,n)} = \mathrm{MLP}_{\mathrm{node}}\bigl(\mathbf{x}^{\text{node}}_{(i,n)}\bigr).
  \end{equation}

  The lifting is itself a gated message-passing layer over the same neighborhood $\mathcal{N}_{(i,n)}^{(k_x,k_t)}$ as the main message-passing layers,
  with the same split into \emph{adjacent} edges (nearest spatial neighbors
  $j=i\pm1$ at the same time level) and the remaining \emph{non-adjacent}
  edges. It uses its own message MLPs (superscript $\mathrm{lift}$) with
  reduced feature sets. For adjacent edges,
  \begin{equation}
      e^{\mathrm{lift,adj}}_{(j,n),(i,n)}
      =
      \Big(
      r_{(j,n),(i,n)},\,
      \Delta t^{\mathrm{feat}}_{(j,n),(i,n)},\,
      \Delta\rho_{(j,n),(i,n)},\,
      \Delta v_{(j,n),(i,n)},\,
      \Delta \omega_{(j,n),(i,n)},\,
      \theta_{(j,n),(i,n)},\,
      \lambda_{1,(j,n),(i,n)},\,
      \lambda_{2,(j,n),(i,n)}
      \Big).
      \label{eq:arz-orig-lift-adj-feats}
  \end{equation}
  Here, $r_{(j,n),(i,n)} \Def\operatorname{sign}(x_j-x_i)$ and the interface eigenvalues are defined by 
\begin{equation}
  \lambda_{1,(j,n),(i,n)} \Def v_{(j,n),(i,n)}-\rho_{(j,n),(i,n)}\,p'(\rho_{(j,n),(i,n)}),
  \qquad
  \lambda_{2,(j,n),(i,n)} \Def v_{(j,n),(i,n)}.
  \label{eq:arz-up-interface-eig}
\end{equation}
Furthermore, the jumps are denoted by
$\Delta\rho_{(j,n),(i,n)} \Def \hat\rho_{(j,n)}-\hat\rho_{(i,n)}$ (with
$\Delta v_{(j,n),(i,n)},\Delta \omega_{(j,n),(i,n)}$ defined analogously), and
$\theta_{(j,n),(i,n)}$ is the router weight, defined as
\begin{equation}
\label{eq:router-weight}
\theta_{(j,m),(i,n)} \Def \frac{\lvert \Delta \omega_{(j,m),(i,n)} \rvert}
                         {\lvert \Delta \omega_{(j,m),(i,n)} \rvert + \lvert \Delta v_{(j,m),(i,n)} \rvert + \varepsilon}
\end{equation}
for sufficiently small $\eps>0$. In addition, the entropy-violation indicator $\chi^{\mathrm{entropy}}_{(j,n),(i,n)}$, is defined in the entropy gate
\eqref{eq:m2-gent}. 

  For non-adjacent edges,
  \begin{equation}
      e^{\mathrm{lift,non-adj}}_{(j,m),(i,n)}
      =
      \Big(
      \Delta x^{\mathrm{feat}}_{(j,m),(i,n)},\,
      \Delta t^{\mathrm{feat}}_{(j,m),(i,n)},\,
      r_{(j,m),(i,n)}
      \Big), \qquad \Delta t^{\mathrm{feat}}_{(j,m),(i,n)} \Def (t_m-t_n)/\Delta t
      \label{eq:arz-orig-lift-nonadj-feats}
  \end{equation}
  i.e.\ non-adjacent lifting edges carry topological information only. Each edge feature
  vector is mapped to a latent message by the MLP of its adjacency class,
  \begin{equation}
  m^{(0)}_{(j,m),(i,n)}
  =
  \begin{cases}
  \mathrm{MLP}^{\mathrm{lift}}_{\mathrm{adj}}
    \bigl(e^{\mathrm{lift,adj}}_{(j,n),(i,n)}\bigr)
    & (j,m)\in\mathcal{A}(i,n),\\[2pt]
  \mathrm{MLP}^{\mathrm{lift}}_{\mathrm{non\text{-}adj}}
    \bigl(e^{\mathrm{lift,non\text{-}adj}}_{(j,m),(i,n)}\bigr)
    & \text{otherwise},
  \end{cases}
  \label{eq:arz-orig-lift-msg}
  \end{equation}
  where $\mathcal{A}(i,n)\subset\mathcal{N}_{(i,n)}^{(k_x,k_t)}$ denotes the adjacent edges.
  The aggregate $M^{(0)}_{(i,n)}$ is the gate-normalized sum of these messages in \Cref{eq:arz-orig-lift-msg}, which directly follows from \Cref{E:aggregation}, with $\ell = 0$.  
  The scalar gates are the upwind, and entropy on adjacent nodes which are defined later in this section (see
  \Cref{arz-up-gate,arz-ent-gate}). For the non-adjacent nodes, we consider the gates to be fully functional. More precisely, 
  otherwise,
  \begin{equation}
  g^{\mathrm{lift}}_{(j,m),(i,n)}
  =
  \begin{cases}
  g^{\mathrm{up}}_{(j,n),(i,n)}\; g^{\mathrm{ent}}_{(j,n),(i,n)},
    & (j,m)\in\mathcal{A}(i,n),\\[2pt]
  1, & \text{otherwise}.
  \end{cases}
  \label{eq:arz-orig-lift-gate}
  \end{equation}
  Finally, the node embedding and the aggregate are
  combined into the initial latent state,
  \begin{equation}
  h^{(0)}_{(i,n)}
  =
  \mathrm{MLP}_{\mathrm{upd}}
  \bigl(
  [\,h^{\mathrm{node}}_{(i,n)},\,M^{(0)}_{(i,n)}\,]
  \bigr),
  \end{equation}
  which is the input to the processor (see \Cref{fig:hypno-arch}). 



\subsubsection{Edge features}
\label{arz-edge-feats}
As in the LWR case, the physical edge quantities are evaluated on the decoded
state probe $(\hat\rho_{(i,n)},\hat \omega_{(i,n)})$, and each node pair is served by a
dedicated message MLP that receives the two latent states together with an edge
feature vector. For an adjacent edge the two nodes share a time level ($m=n$,
$j=i\pm1$), and the message is
\[
m_{(i,n),(j,n)}^{\mathrm{adj},(\ell)}
=
\MLP_{\mathrm{adj}}
\left(
h_{(i,n)}^{(\ell)},\,
h_{(j,n)}^{(\ell)},\,
e_{(j,n),(i,n)}^{\mathrm{adj},(\ell)}
\right),
\qquad j=i\pm1,
\]
with the edge feature vector
\begin{multline}
    e_{(j,n),(i,n)}^{\mathrm{adj},(\ell)}
    =
    \Big(
    r_{(j,n),(i,n)},\,
    \lambda_{1,(j,n),(i,n)},\,
    \lambda_{2,(j,n),(i,n)},\,
    \chi^{\mathrm{up},1}_{(j,n),(i,n)},\,
    \chi^{\mathrm{up},2}_{(j,n),(i,n)},\,
    \Delta\rho_{(j,n),(i,n)},\,
    \Delta v_{(j,n),(i,n)},\, \\ 
    \Delta \omega_{(j,n),(i,n)},\,
    \theta_{(j,n),(i,n)},\,
    \chi^{\mathrm{entropy}}_{(j,n),(i,n)}
    \Big),
    \label{eq:arz-orig-mp-adj-feats}
\end{multline}
  where all the elements are defined in \Cref{eq:arz-orig-lift-adj-feats} and its following lines. Furthermore $\chi^{\mathrm{up},1}_{(j,n),(i,n)}$ and $\chi^{\mathrm{up},2}_{(j,n),(i,n)}$ are defined as in \Cref{eq:arz-upwind-flags} and $\chi^{\mathrm{entropy}}_{(j, n), (i, n)}$ is defined in \Cref{eq:m2-gent}.
\begin{remark}
    In practice we have also considered $\lambda_{1,(j,n),(i,n)}\,r_{(j,n),(i,n)}$ and $ \lambda_{2,(j,n),(i,n)} r_{(j,n),(i,n)}$ as part of the edge features.
\end{remark}
The per-family upwind flags are defined by
\begin{equation}
    \chi^{\mathrm{up},1}_{(j,n),(i,n)}
    =
    \ind{ \{\lambda_{1,(j,n),(i,n)}\, r_{(j,n),(i,n)}<0 \}},
    \qquad
    \chi^{\mathrm{up},2}_{(j,n),(i,n)}
    =
    \ind{\{\lambda_{2,(j,n),(i,n)}\, r_{(j,n),(i,n)}<0 \}}.
    \label{eq:arz-upwind-flags}
\end{equation}

For a non-adjacent edge $(j,m)\to(i,n)$, interface quantities such as the
Rankine--Hugoniot speed are undefined, so the message MLP receives only
topological features,
\[
m_{(i,n),(j,m)}^{\mathrm{non-adj},(\ell)}
=
\MLP_{\mathrm{non-adj}}
\left(
h_{(i,n)}^{(\ell)},\,
h_{(j,m)}^{(\ell)},\,
e_{(j,m),(i,n)}^{\mathrm{non-adj},(\ell)}
\right),
\]
\begin{equation}
    e_{(j,m),(i,n)}^{\mathrm{non-adj},(\ell)}
    =
    \Big(
    \Delta x^{\mathrm{feat}}_{(j,m),(i,n)},\,
    \Delta t^{\mathrm{feat}}_{(j,m),(i,n)},\,
    r_{(j,m),(i,n)},\,
    \lambda^{\mathrm{spec}}_{(j,m)}
    \Big),
    \label{eq:arz-orig-mp-nonadj-feats}
\end{equation}
where $\Delta x^{\mathrm{feat}}_{(j,m),(i,n)} \Def (x_j-x_i)/\Delta x$ and
$\Delta t^{\mathrm{feat}}_{(j,m),(i,n)} \Def (t_m-t_n)/\Delta t$ are the grid-normalized (hence resolution-invariant) space--time offsets, and
$\lambda^{\mathrm{spec}}_{(j,m)}\Def \max \bigl(\lvert\lambda_{1,j,m}\rvert,\lvert\lambda_{2,j,m}\rvert \bigr)$
is the spectral radius at the neighboring node $(j,m)$.

\subsubsection{Upwind gate}
\label{arz-up-gate}
Unlike the scalar LWR case, the ARZ system carries two characteristic families,
so directionality must be resolved \emph{per family}. For an adjacent edge
$(j,n)\to(i,n)$ we evaluate each family eigenvalue at the arithmetic interface
state
\begin{equation}
  \rho_{(j,n),(i,n)} \Def \tfrac12(\hat\rho_{(i,n)}+\hat\rho_{(j,n)}),
  \qquad
  v_{(j,n),(i,n)} \Def \tfrac12(\hat v_{(i,n)}+\hat v_{(j,n)}),
\end{equation}
which represent the average of the eigenvalues of the continuous system (\Cref{eq:arz_eigenvalues}).
As in the LWR case, position of the neighbor relative to the source is captured by
$r_{(j,n),(i,n)}=\operatorname{sign}(x_j-x_i)$.

A family-$m$ message should be favored when the corresponding wave at the
interface travels from $(j,n)$ toward $(i,n)$, i.e. when
$\lambda_{m,(j,n),(i,n)}\,r_{(j,n),(i,n)}<0$. We turn this hard upwind condition
into a smooth, learnable gate per family by introducing the upwind gater in the form of 
\begin{equation}
  g^{\mathrm{up},m}_{(j,n),(i,n)}
  =
  \mathrm{sigmoid}\!\left(
  -\,\frac{\lambda_{m,(j,n),(i,n)}\,r_{(j,n),(i,n)}}{\tau_m +\varepsilon}
  \right),
  \qquad
  \tau_m=\operatorname{softplus}(\theta_{\tau,m}),
  \qquad m\in\{1,2\},
  \label{eq:arz-up-perfamily}
\end{equation}

The two per-family gates are combined into a single edge weight, so an edge is kept open if it is upstream for \emph{either}
family:
\begin{equation}
  g^{\mathrm{up}}_{(j,n),(i,n)}
  =
  1-\bigl(1-g^{\mathrm{up},1}_{(j,n),(i,n)}\bigr)\bigl(1-g^{\mathrm{up},2}_{(j,n),(i,n)}\bigr).
  \label{eq:arz-up-union}
\end{equation}
In the hard limit $\tau_m\to 0$ this reduces to the indicator flags
$\chi^{\mathrm{up},m}_{(j,n),(i,n)}=\ind{\{\lambda_{m,(j,n),(i,n)}\,r_{(j,n),(i,n)}<0\}}$
of \Cref{eq:arz-upwind-flags}, recovering classical per-characteristic upwinding.

\subsubsection{Entropy gate}
\label{arz-ent-gate}
The entropy gate is applied to the $1$-family only, as it is the only family
that should satisfy the Lax--Oleinik condition. For an adjacent edge
$(j,n)\to(i,n)$ we orient the two states by the sign of $r_{(j,n),(i,n)}$,
writing $\rho_L,\rho_R$ for the left and right states and $\lambda_{1,L}=\lambda_1(\rho_L, v_L)$,
$\lambda_{1,R}=\lambda_1(\rho_R, v_R)$ for their $1$-eigenvalues. The $1$-family
Rankine--Hugoniot speed is the secant slope of the $\rho v$ flux,
\begin{equation}
  s_{1,(j,n),(i,n)} =
  \begin{cases}
    \dfrac{\hat\rho_{(j,n)}\, \hat v_{(j,n)} - \hat\rho_{(i,n)}\, \hat v_{(i,n)}}
          {\hat\rho_{(j,n)} - \hat\rho_{(i,n)}},
      & \abs{\Delta\rho_{(j,n),(i,n)}}\ge\delta,\\[1.8ex]
    \lambda_{1,(j,n),(i,n)}, & \abs{\Delta\rho_{(j,n),(i,n)}}<\delta,
  \end{cases}
  \label{eq:arz-orig-s1}
\end{equation}
with the differential limit $\lambda_{1,(j,n),(i,n)}$ taken when the density jump
falls below a small threshold $\delta$. An admissible $1$-shock then satisfies
the Lax condition
\begin{equation}
    \lambda_{1,R} \;<\; s_{1,(j,n),(i,n)} \;<\; \lambda_{1,L}.
    \label{eq:m2-lax}
\end{equation}
Since this should hold only on $1$-family shock interfaces, we first isolate
compressive $1$-waves with the shock indicator
\begin{equation}
    \chi^{\mathrm{sh}}_{(j,n),(i,n)}
      =
    \ind{\{\lambda_{1,L}>\lambda_{1,R}\}},
    \label{arz-shock-flag}
\end{equation}
so that rarefaction-type interfaces remain fully communicative through messages. The penalty is further confined
to genuine $1$-waves by the router weight $\theta_{(j, n), (i, n)}$ as in \Cref{eq:router-weight} which approaches $0$ when the interface jump is carried by $\omega$ (a $1$-wave) and
$1$ when it is carried by $v$ (a $2$-contact); the factor
$(1-\theta_{(j,n),(i,n)})$ therefore removes the entropy penalty on
contact-dominated interfaces. The entropy gate is
\begin{align}
    g^{\mathrm{ent}}_{(j,n),(i,n)}
      &= 1 - (1-\gamma)\,\bigl(1-\theta_{(j,n),(i,n)}\bigr)\,\chi^{\mathrm{entropy}}_{(j,n),(i,n)},
    \\
    \chi^{\mathrm{entropy}}_{(j,n),(i,n)}
     & =
      \chi^{\mathrm{sh}}_{(j,n),(i,n)} \cdot
      \mathbb{1}_{\set{\,s_{1,(j,n),(i,n)} > \lambda_{1,L}\;\;\text{or}\;\;s_{1,(j,n),(i,n)} < \lambda_{1,R}}},
    \\
    \gamma &= \mathrm{sigmoid}(\theta_\gamma),
    \label{eq:m2-gent}
\end{align}
where $\chi^{\mathrm{entropy}}_{(j,n),(i,n)}\in\{0,1\}$ flags a compressive
$1$-shock whose speed violates the Lax bracket (\Cref{eq:m2-lax}), and
$\gamma=\sigma(\theta_\gamma)\in(0,1)$ is the learnable attenuation floor (with
$\theta_\gamma$ the raw parameter): admissible shocks, rarefactions, and contacts
keep weight $1$, while entropy-violating $1$-shocks are attenuated toward $\gamma$.

\subsubsection{Temporal gate}
Edges with a nonzero temporal offset ($\Delta t\neq0$) are weighted by a soft
CFL-based gate. Let
\begin{equation}
  \varrho_{(i,n)} = \max\!\bigl(\lvert\lambda_{1,i,n}\rvert,\;\lvert\lambda_{2,i,n}\rvert\bigr)
\end{equation}
be the ARZ spectral radius at the source node $(i,n)$, evaluated on the decoded
state $(\hat\rho_{(i,n)},\hat \omega_{(i,n)})$. To be physically informative, a
non-adjacent edge $(j,m)\to(i,n)$ spanning $\Delta t=t_m-t_n$ in time must allow
a wave to traverse its spatial gap $\Delta x$. The
corresponding CFL number and gate are
\begin{equation}
  \label{eq:arz-cfl-gate}
  g^{\mathrm{CFL}}_{(j,m),(i,n)}
  = \exp\Bigl\{-\kappa\,\bigl\{\mathrm{ReLU}(\mathrm{CFL}-1)\bigr\}^2\Bigr\},
  \quad   \mathrm{CFL}
  = \frac{\Bigl(\max_{q \in \set{-k_t, \cdots, 0}}\max_{i \le k \le j} \varrho_{(k,n+q)}\Bigr) \Delta t}{\Delta x},
  \quad
  \kappa = \operatorname{softplus}(\theta_\kappa),
\end{equation}
so an edge is suppressed once a wave cannot cross its spatial gap within the
temporal span ($\mathrm{CFL}>1$), and is left open otherwise; here $\theta_\kappa$
is the raw (learnable) CFL-decay parameter and
$\kappa=\operatorname{softplus}(\theta_\kappa)>0$.

\subsubsection{Composite gate}
The gates combine per edge class. An adjacent edge carries the product of the
upwind and entropy gates, $g_{(j,n),(i,n)}=g^{\mathrm{up}}_{(j,n),(i,n)}\,g^{\mathrm{ent}}_{(j,n),(i,n)}$,
while a non-adjacent space-time edge carries the CFL gate alone,
$g_{(j,m),(i,n)}=g^{\mathrm{CFL}}_{(j,m),(i,n)}$ (the upwind and entropy gates require an interface
and are undefined there). These weights feed the gate-normalized aggregation of
the message-passing layers.

\subsubsection{Network Parameters}
\Cref{tab:arz-config} lists HypNO network parameter choices for the original, mark-1, and mark-2 HypNO-ARZ models.
\begin{table*}[t]
\centering
\caption{Network Configuration Table describing training and network choices for HypNO-ARZ original, mark-1 and mark-2 models.}
\label{tab:arz-config}
\small
\resizebox{\textwidth}{!}{
\begin{tabular}{ccccccccccc}
\toprule
\textbf{Model} & \textbf{k\_x} & \textbf{k\_t} & \textbf{d\_latent} & \textbf{d\_hidden} & \textbf{n\_layers} & \textbf{activation} & \textbf{decoder\_depth} & \textbf{optimizer} & \textbf{epochs} & \textbf{batch\_size} \\
\midrule
\textbf{HypNO-ARZ} & 6 & 4 & 96 & 96 & 11 & gelu & 5 & adamw & 220 & 16 \\
\textbf{HypNO-ARZ-HLL} & 6 & 5 & 96 & 96 & 12 & gelu & 5 & adamw & 220 & 16 \\
\bottomrule
\end{tabular}
}
\end{table*}

\subsection{Choice of depth and stencil width: receptive field vs.\ domain of dependence}
\label{app:reach}
The depth $L$ and the stencil offsets $(k_x,k_t)$ are not free capacity
parameters: together with the grid they fix the model's receptive field, and
hyperbolicity imposes a hard lower bound on it. Each message-passing layer
extends the dependency of a node on the (broadcast) initial data by at most
$k_x$ cells per side, regardless of $k_t$, so after $L$ layers the prediction
at $(i,n)$ can depend on the initial condition only within the spatial reach
\begin{equation}
  \label{eq:spatial-reach}
  \abs{x_j - x_i} \;\le\; L\,k_x\,\Delta x,
\end{equation}
a rectangle in $(x,t)$ whose width is constant in time (the gated lifting
layer contributes one further stencil hop, which we conservatively ignore).
The PDE, in contrast, prescribes a domain of dependence that grows in time:
the entropy solution at $(x_i,t_n)$ depends on the initial data on the cone
$\set{x : \abs{x-x_i}\le c_{\max}\,t_n}$, where $c_{\max}=\max\abs{\lambda}$
is the fastest characteristic speed ($c_{\max}=1$ for LWR, since
$\lambda(\rho)=1-2\rho$ with $\rho\in[0,1]$, and $c_{\max}\le 1$ for ARZ with
$p(\rho)=\rho$, since $\lambda_1=v-\rho$, $\lambda_2=v$ with $v\in[0,1]$,
$\rho\in(0,1)$). For the network to be able to represent the solution up to
the final time $\timehor$, the receptive field must contain this cone at
every query node, which yields the necessary condition
\begin{equation}
  \label{eq:reach-bound}
  L\,k_x\,\Delta x \;\ge\; c_{\max}\,\timehor
  \qquad\Longleftrightarrow\qquad
  L\,k_x \;\ge\; \frac{c_{\max}\,\timehor}{\Delta x} \;=\; 64
\end{equation}
on our grids ($\Delta x = 2/n_x = 1/64$, $\timehor=1$).

This bound is sharp in practice. In earlier, under-provisioned
configurations (e.g.\ $L=9$, $k_x=5$, so $L\,k_x=45<64$) we observed a
characteristic failure mode on single-shock Riemann data: the predicted shock
track advances correctly and then \emph{freezes} at
$t_{\mathrm{freeze}} \approx L\,k_x\,\Delta x/\abs{s}$, where $s$ is the shock
speed---exactly the time at which the foot of the characteristic determining
the solution leaves the receptive field \eqref{eq:spatial-reach}. Sweeps over
Riemann initial data confirmed the predicted linear scaling of
$t_{\mathrm{freeze}}$ with $1/\abs{s}$, with slope $L\,k_x\,\Delta x$. Past
this time the model cannot, even in principle, read the initial data that
determine the solution, so no amount of training removes the freeze---only
enlarging $L\,k_x$ does.

We therefore size $L$ and $k_x$ jointly so that \eqref{eq:reach-bound} holds
with margin, and split the product between depth and width as follows.
Per-layer cost and memory scale with the number of edges per node,
$(2k_x+1)(k_t+1)$, and a single layer should not out-run the physics by too
much: the per-layer stencil has an effective signal speed
$c_{\mathrm{net}} = k_x\Delta x/(k_t\Delta t)$, and edges far outside the
physical cone ($c_{\mathrm{net}}\gg c_{\max}$) are wasted capacity that the
CFL gate (\Cref{eq:lwr-cfl-gate,eq:arz-cfl-gate}) must learn to suppress.
Conversely, the product cannot be met by depth alone, since cost grows
linearly in $L$ and very deep message-passing stacks oversmooth
\citep{rusch2023oversmoothing}. Keeping $k_x$ moderate, with
$c_{\mathrm{net}}$ a small multiple of $c_{\max}$, and setting the depth from
\eqref{eq:reach-bound} gives $L=12$, $k_x=8$ for LWR ($L\,k_x=96$, spatial
reach $1.5$, a $1.5\times$ margin) and $L=11$, $k_x=6$ for ARZ ($L\,k_x=66$;
the HLL-supervised variant in \Cref{tab:arz-config} uses $L=12$, so
$L\,k_x=72$). The margin matters because \eqref{eq:reach-bound} is necessary
rather than sufficient: initial data near the boundary of the receptive field
reach the query node only through a single chain of maximal-length hops and
are strongly attenuated.

The temporal offset $k_t$ is not constrained by initial-data access: because
the initial condition is broadcast to all time levels before encoding
(\Cref{app:lwr-lifting,app:lifting}), temporal edges serve intra-network
time-marching and information staging rather than reach, and $k_t$ is chosen
so that $c_{\mathrm{net}}$ stays close to $c_{\max}$ at fixed edge budget.

\section{Dataset construction}
\label{sec:data-construction}
In this section, we elaborate on how we construct the dataset for training and testing the proposed HypNO architecture in comparison to the existing schemes. 

\subsection{Stratified value sampling: LWR model}
\label{sec:lwr-data-families}
We consider the initial-condition families: riemann and piecewise constant (see \Cref{sec:selection_IC_LWR} for more details). In the latter family, the segment counts belong to $\{2, \cdots, p\}$, $p \ge2$.  The total sample count is selected so that every bin receives an identical
quota
\begin{equation*}
\text{Number of samples per bin} = \frac{N}{\text{\# of bins}}.
\end{equation*}
For the training set, $p = 6$ segments
give $18$ bins; with $N = 5400$ this yields $300$ samples per bin. Each bin is generated with an independent random
seed, so the bins are statistically independent and the dataset is
fully reproducible. Every sample additionally records its segment count and IC family as metadata, enabling
per-bin evaluation.

Provided that the learning pattern would be impacted by the magnitude of jumps, $\lvert \rho_L - \rho_R \rvert$ represents the left and right density values across the jumps, respectively.
The data is sampled by randomly sampling the 
\emph{jump magnitude} across discontinuities.  Given the value range $[\rho_{\min}, \rho_{\max}]$, minimum and maximum value of the solution, respectively, with the width of $\rho_{\max} - \rho_{\min}$:
\begin{enumerate}
\item
Draw a jump magnitude
      $\Delta \rho \sim \mathcal{U}\!\left(\Delta_{\min},\,
      \min(\Delta_{\max}, \rho_{\max} - \rho_{\min})\right)$,
      with $\Delta_{\min} = 0.03$ and $\Delta_{\max} = 0.95$ and $\mathcal U$ denotes the uniform distribution. The
      floor $\Delta_{\min}$ guarantees every discontinuity is
      resolvable on the grid.
\item Draw a mean level
      $\bar{\rho} \sim \mathcal{U}(\rho_{\min}, \rho_{\max})$.
\item Consider the interval
      $\bigl[\bar{\rho} - \tfrac{1}{2}\Delta \rho,\;
             \bar{\rho} + \tfrac{1}{2}\Delta \rho\bigr]$. In the case one of the endpoints of the interval exceeds the respected $\rho_{\min}$ or $\rho_{\max}$, we shift the interval such that it respects the proper values. By the construction of the interval; such a shift ensures that the interval falls into the correct range.

\item Define $\rho_L, \rho_R \coloneqq \bar \rho \pm \tfrac 12 \Delta \rho$ which are selected randomly.  
\end{enumerate}
The resulting jump $\Delta \rho = |\rho_L - \rho_R|$ is therefore uniform over
$[\Delta_{\min}, \Delta_{\max}]$ \emph{at every value level}.

\paragraph{Motivation} Drawing $\rho_L$ and $\rho_R$ as two independent
uniforms makes the jump $\Delta \rho$ \emph{triangularly} distributed,
peaked at $\Delta \rho = 0$: both weak and strong jumps are rare, and the
value-range corners (e.g.\ $\rho_L = 0.9$, $\rho_R = 0.92$) are severely
undersampled. The reparameterization in terms of
$(\bar{\rho}, \Delta \rho)$ removes this bias.

\subsubsection{Initial-condition families}
\label{sec:selection_IC_LWR}
\begin{itemize}
\item \textbf{\texttt{riemann\_stratified}.} A single discontinuity
      at a location $x_0$ drawn uniformly in the interior; the two
      states $(\rho_L, \rho_R)$ come from the definition above.
\item \textbf{\texttt{piecewise\_constant}.} The domain
      is partitioned into $p$ segments by $p-1$ random cut points.
      Segment values follow a controlled random walk: the first is
      uniform on $[\rho_{\min}, \rho_{\max}]$, and each subsequent value
      is the previous one plus a step of magnitude
      $\mathcal{U}(\Delta_{\min}, \Delta_{\max})$ and random sign,
      reflected off the range boundaries. Every interface is thus a
      visible, well-spread discontinuity.

\end{itemize}

\subsection{Stratified value sampling: ARZ model}
  \label{sec:arz-data-families}
  The ARZ model evolves a \emph{pair} of fields, so an initial condition
  consists of a density profile $\rho_0(x)$ and a velocity profile
  $v_0(x)$; the Lagrangian marker is derived as
  $w_0 = v_0 + p(\rho_0)$ and is never sampled independently. Densities
  are drawn from $[\rho_{\min}, \rho_{\max}] = [0.1, 0.9]$ --- bounded
  away from vacuum ($\rho = 0$) and from the jam density --- and
  velocities from $[v_{\min}, v_{\max}] = [0, 1]$.

  As in the LWR case, the dataset is stratified over
  (IC family)\,$\times$\,(segment count) bins with an identical quota per
  bin. We use the three families of \Cref{sec:selection_IC_ARZ} and
  segment counts $p \in \{2, 3, 5, 7, 10\}$, giving $15$ bins; with
  $N = 6000$ this yields $400$ samples per bin. (For the Riemann family
  the segment count is vacuous --- every sample has a single interface ---
  so its five bins simply enlarge the family quota.) Per-sample seeds are
  derived deterministically from a single master seed, so the dataset is
  fully reproducible, and every sample records its family and segment
  count as metadata for per-bin evaluation.

  Value sampling reuses the mean-level/jump-magnitude reparameterization
  of \Cref{sec:lwr-data-families}, applied \emph{per channel}: the
  density jump $\lvert \rho_L - \rho_R \rvert$ and the velocity jump
  $\lvert v_L - v_R \rvert$ are drawn independently, each uniform at
  every value level of its own range. Two ARZ-specific refinements are
  made for the Riemann family:
  \begin{enumerate}
  \item \emph{Co-located interfaces.} The $\rho$ and $v$ discontinuities
        share a single interface position $x_0$, drawn uniformly over the
        central $60\%$ of the domain. Sampling the two channels with
        independent jump locations would instead produce two separated
        Riemann problems, neither of which exercises the coupled wave
        structure of the system.
  \item \emph{Binned jump magnitudes.} Each jump magnitude is drawn by
        first selecting one of $B = 8$ equal-width bins covering
        $[\Delta_{\min},\, u_{\max} - u_{\min}]$ (with
        $\Delta_{\min} = 0.01\,(u_{\max} - u_{\min})$, where $u$ stands
        for the channel being sampled) uniformly at random, and then
        sampling uniformly within the bin. Compared to the LWR draw,
        this extends coverage to the full channel span and lowers the
        floor on weak jumps, while retaining a minimum magnitude so that
        every discontinuity remains resolvable on the grid.
  \end{enumerate}

  \subsubsection{Initial-condition families}
  \label{sec:selection_IC_ARZ}
  \begin{itemize}
  \item \textbf{\texttt{riemann}.} A single co-located discontinuity in
        both channels at $x_0$; the states
        $(\rho_L, v_L)$, $(\rho_R, v_R)$ come from the sampler above.
  \item \textbf{\texttt{piecewise\_constant}.} The construction of
        \Cref{sec:selection_IC_LWR} applied independently to each
        channel: random cut points and a reflected random walk of
        controlled step magnitude, over $[\rho_{\min}, \rho_{\max}]$ for
        $\rho_0$ and $[v_{\min}, v_{\max}]$ for $v_0$.
  \item \textbf{\texttt{sine\_staircase}.} A single global sine of random
        frequency and phase, quantised into $p$ equal-width
        piecewise-constant levels: the domain is split into $p$ equal
        segments and each segment takes the sine's value at its midpoint.
        The result approximates a smooth profile while remaining
        piecewise constant with finitely many discontinuities, which is
        required for the exact solver below; $p$ controls the staircase
        coarseness. Each channel receives an independent staircase.
  \end{itemize}

  \subsubsection{Reference solutions}
  \label{sec:arz-ground-truth}
  All training data uses the homogeneous system ($\tau = \infty$), so
  exact solutions are available and no finite-volume discretisation
  enters the ground truth. For the Riemann family, the exact solution of
  the ARZ Riemann problem is evaluated directly at the cell midpoints at
  each output time. For the multi-segment families, we use a
  wave-front-tracking (WFT) solver: since every initial condition is
  piecewise constant with finitely many interfaces, each interface spawns
  a local Riemann fan whose shocks and contact discontinuities are
  propagated exactly, and whose rarefactions are approximated by fans of
  small fronts with density increment $\delta = 0.1\, \Delta x$; front
  interactions are resolved exactly in order of collision time. On
  Riemann data the WFT solution agrees with the exact solver to
  $\mathcal{O}(10^{-7})$ in the maximum norm, i.e.\ to single-precision
  machine accuracy, so the ground truth is sharp: shocks and contacts
  occupy a single cell, and any multi-cell transition in the data is a
  genuine rarefaction fan rather than numerical diffusion.

  \subsubsection{Wave-structure stratification: Riemann dataset}
  \label{sec:arz-riemann-stratified}
  For the pure-Riemann dataset we stratify over the \emph{wave structure}
  of the solution rather than over raw jump magnitudes. The solution of
  an ARZ Riemann problem consists of a 1-wave (shock or rarefaction,
  across which $w$ is constant) connecting $(\rho_L, v_L)$ to an
  intermediate state $(\rho_*, v_*)$, followed by a 2-contact (across
  which $v$ is constant) connecting $(\rho_*, v_*)$ to $(\rho_R, v_R)$.
  We partition the samples over a grid of
  \[
    \underbrace{2}_{\text{1-wave type}} \times
    \underbrace{4}_{\text{1-wave strength}} \times
    \underbrace{4}_{\text{contact strength}} = 32
    \text{ cells},
  \]
  where the strength bins split $[0, \rho_{\max} - \rho_{\min}]$ into
  equal intervals for $\lvert \rho_* - \rho_L \rvert$ and
  $\lvert \rho_R - \rho_* \rvert$ respectively, and each cell receives an
  identical quota ($N = 5000$ in total). Uniform i.i.d.\ endpoint
  sampling under-covers exactly the corners this grid enforces: strong
  1-shocks with weak contacts, and vice versa.

  Within a cell, states are sampled \emph{vacuum-free by construction}:
  rather than drawing the four endpoint values directly (which produces
  a vacuum intermediate state, $\rho_* = 0$, or an over-jam state,
  $\rho_* > \rho_{\max}$, in a non-negligible fraction of draws), we draw
  the physical degrees of freedom $\rho_L$, $\rho_*$, $\rho_R \in
  [\rho_{\min}, \rho_{\max}]$ and $v_*$ directly, then recover the
  remaining values from the Riemann invariants:
   $\underbrace{4}_{\text{contact strength}} = 32
    \text{ cells},
  $
  where the strength bins split $[0, \rho_{\max} - \rho_{\min}]$ into
  equal intervals for $\lvert \rho_* - \rho_L \rvert$ and
  $\lvert \rho_R - \rho_* \rvert$ respectively, and each cell receives an
  identical quota ($N = 5000$ in total). Uniform i.i.d.\ endpoint
  sampling under-covers exactly the corners this grid enforces: strong
  1-shocks with weak contacts, and vice versa.

  The intermediate velocity $v_*$ is drawn from the subinterval
  of $[v_{\min}, v_{\max}]$ for which $v_L$ also lands in range, so both
  velocities are bounded without any post-hoc clipping that would break
  the invariant. The 1-wave type fixes the sign of $\rho_* - \rho_L$
  (shock for $\rho_L < \rho_*$, rarefaction for $\rho_L > \rho_*$), and
  the contact sign is drawn at random.
\section{LWR evaluation appendix}
This appendix collects the LWR diagnostics on $\rho$: the MAE breakdown per number of segments for HypNO-LWR, WENO5, Godunov, and FNO.
\begin{table}[h]
\centering
\caption{LWR Headline result: MAE vs Lax-Hopf exact (mean $\pm$ std). The ID subset is (ic\_type, num\_segments) cells whose ic\_type and num\_segments both appear in the training configuration; the OOD subset is the rest of the evaluation set.}
\label{tab:paper_eval_headline}
\begin{tabular}{lcccc}
\hline
Subset & HypNO-LWR & WENO5 & Godunov & FNO \\
\hline
ID & \besterr{1.39{\times}10^{-3}}{1.03{\times}10^{-3}} & \err{2.71{\times}10^{-3}}{2.15{\times}10^{-3}} & \err{4.86{\times}10^{-3}}{3.90{\times}10^{-3}} & \err{5.43{\times}10^{-3}}{3.87{\times}10^{-3}} \\
OOD & \besterr{3.16{\times}10^{-3}}{2.77{\times}10^{-3}} & \err{4.93{\times}10^{-3}}{4.29{\times}10^{-3}} & \err{7.32{\times}10^{-3}}{5.80{\times}10^{-3}} & \err{9.01{\times}10^{-3}}{7.14{\times}10^{-3}} \\
all & \besterr{2.37{\times}10^{-3}}{2.35{\times}10^{-3}} & \err{3.94{\times}10^{-3}}{3.67{\times}10^{-3}} & \err{6.23{\times}10^{-3}}{5.19{\times}10^{-3}} & \err{7.42{\times}10^{-3}}{6.18{\times}10^{-3}} \\
\hline
\end{tabular}
\end{table}

\begin{table}[h]
\centering
\caption{LWR Appendix: per-num\_segments MAE breakdown, mean $\pm$ std, pooled across all IC types. ID and OOD rows match \Cref{tab:paper_eval_headline}; the final row pools the entire evaluation set.}
\label{tab:paper_eval_appendix}
\begin{tabular}{lcccc}
\hline
n\_seg & HypNO-LWR & WENO5 & Godunov & FNO \\
\hline
2  & \besterr{5.36{\times}10^{-4}}{2.41{\times}10^{-4}} & \err{9.12{\times}10^{-4}}{5.93{\times}10^{-4}} & \err{1.64{\times}10^{-3}}{1.55{\times}10^{-3}} & \err{2.46{\times}10^{-3}}{1.85{\times}10^{-3}} \\
3  & \besterr{7.77{\times}10^{-4}}{3.70{\times}10^{-4}} & \err{1.72{\times}10^{-3}}{1.22{\times}10^{-3}} & \err{3.04{\times}10^{-3}}{2.18{\times}10^{-3}} & \err{3.81{\times}10^{-3}}{2.32{\times}10^{-3}} \\
5  & \besterr{1.28{\times}10^{-3}}{6.86{\times}10^{-4}} & \err{2.73{\times}10^{-3}}{1.67{\times}10^{-3}} & \err{5.14{\times}10^{-3}}{3.29{\times}10^{-3}} & \err{5.22{\times}10^{-3}}{2.60{\times}10^{-3}} \\
7  & \besterr{1.52{\times}10^{-3}}{8.20{\times}10^{-4}} & \err{3.07{\times}10^{-3}}{1.92{\times}10^{-3}} & \err{5.50{\times}10^{-3}}{3.72{\times}10^{-3}} & \err{6.45{\times}10^{-3}}{3.96{\times}10^{-3}} \\
8  & \besterr{1.79{\times}10^{-3}}{8.92{\times}10^{-4}} & \err{3.39{\times}10^{-3}}{1.98{\times}10^{-3}} & \err{6.20{\times}10^{-3}}{3.21{\times}10^{-3}} & \err{7.06{\times}10^{-3}}{4.98{\times}10^{-3}} \\
10 & \besterr{2.23{\times}10^{-3}}{1.26{\times}10^{-3}} & \err{4.13{\times}10^{-3}}{2.50{\times}10^{-3}} & \err{7.18{\times}10^{-3}}{4.17{\times}10^{-3}} & \err{7.58{\times}10^{-3}}{4.42{\times}10^{-3}} \\
20 & \besterr{3.85{\times}10^{-3}}{2.37{\times}10^{-3}} & \err{5.95{\times}10^{-3}}{3.95{\times}10^{-3}} & \err{8.99{\times}10^{-3}}{5.80{\times}10^{-3}} & \err{1.08{\times}10^{-2}}{6.67{\times}10^{-3}} \\
25 & \besterr{4.56{\times}10^{-3}}{2.96{\times}10^{-3}} & \err{6.48{\times}10^{-3}}{4.78{\times}10^{-3}} & \err{9.00{\times}10^{-3}}{6.68{\times}10^{-3}} & \err{1.12{\times}10^{-2}}{7.61{\times}10^{-3}} \\
30 & \besterr{4.84{\times}10^{-3}}{3.19{\times}10^{-3}} & \err{7.11{\times}10^{-3}}{5.23{\times}10^{-3}} & \err{9.37{\times}10^{-3}}{6.73{\times}10^{-3}} & \err{1.22{\times}10^{-2}}{8.56{\times}10^{-3}} \\
\hline
ID  & \besterr{1.39{\times}10^{-3}}{1.03{\times}10^{-3}} & \err{2.71{\times}10^{-3}}{2.15{\times}10^{-3}} & \err{4.86{\times}10^{-3}}{3.90{\times}10^{-3}} & \err{5.43{\times}10^{-3}}{3.87{\times}10^{-3}} \\
OOD & \besterr{3.16{\times}10^{-3}}{2.77{\times}10^{-3}} & \err{4.93{\times}10^{-3}}{4.29{\times}10^{-3}} & \err{7.32{\times}10^{-3}}{5.80{\times}10^{-3}} & \err{9.01{\times}10^{-3}}{7.14{\times}10^{-3}} \\
\hline
all & \besterr{2.37{\times}10^{-3}}{2.35{\times}10^{-3}} & \err{3.94{\times}10^{-3}}{3.67{\times}10^{-3}} & \err{6.23{\times}10^{-3}}{5.19{\times}10^{-3}} & \err{7.42{\times}10^{-3}}{6.18{\times}10^{-3}} \\
\hline
\end{tabular}
\end{table}

\FloatBarrier

\section{ARZ evaluation appendix}
This appendix collects the ARZ shock-neighborhood diagnostics on $\rho$: the
MAE in the full domain, the 1-shock band, and the 2-contact band as a function of IC complexity, followed by representative per-segment-count comparisons for full field, zoomed band, and a
single time slice, of HypNO-ARZ against the numerical baselines. The band masks are defined
in \Cref{app:shock-band}.
\setcounter{topnumber}{4}
\setcounter{bottomnumber}{4}
\setcounter{totalnumber}{8}
\renewcommand{\topfraction}{0.95}
\renewcommand{\bottomfraction}{0.95}
\renewcommand{\textfraction}{0.03}
\renewcommand{\floatpagefraction}{0.85}
\setlength{\floatsep}{6pt plus 2pt minus 2pt}
\setlength{\textfloatsep}{8pt plus 2pt minus 2pt}
\setlength{\abovecaptionskip}{4pt}

\begin{figure}[!htbp]
  \centering
  \includegraphics[width=0.9\linewidth]{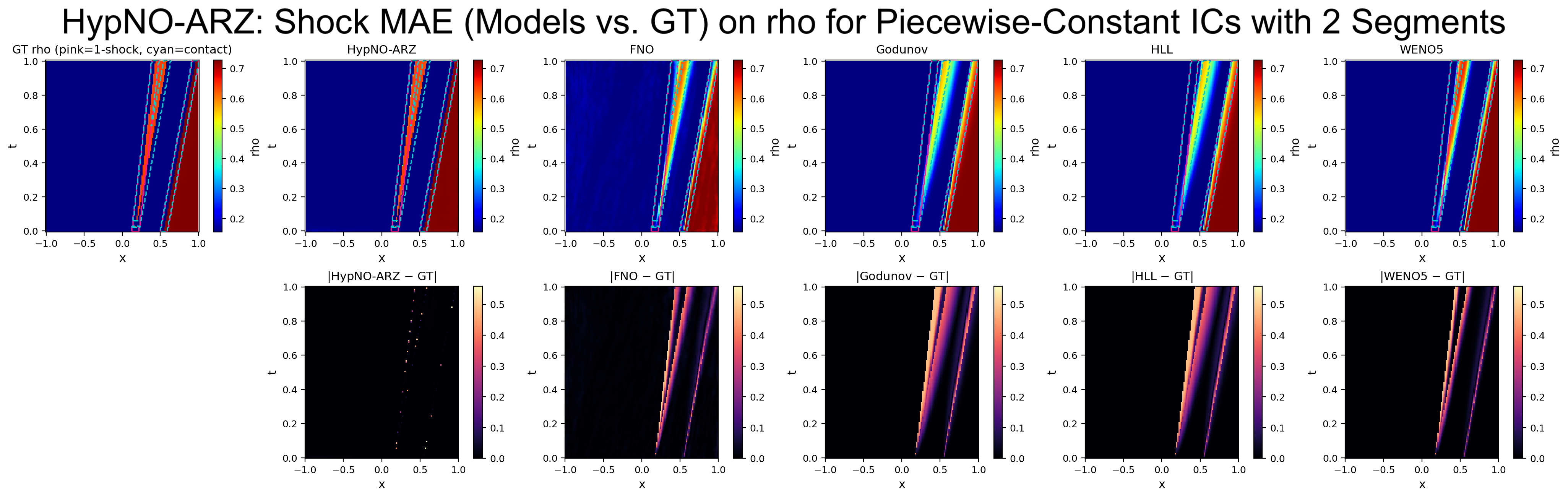}
  \caption{ARZ: Representative sample for an initial condition partitioned into $N=2$ initial discontinuity segments. Top row: $\rho$ fields with the 1-shock band (pink) and contact band (cyan dashed) outlined; bottom row: absolute error to the ground truth.}
  \label{fig:arz-shock-compare-2}
\end{figure}
\begin{figure}[!htbp]
  \centering
  \includegraphics[width=0.9\linewidth]{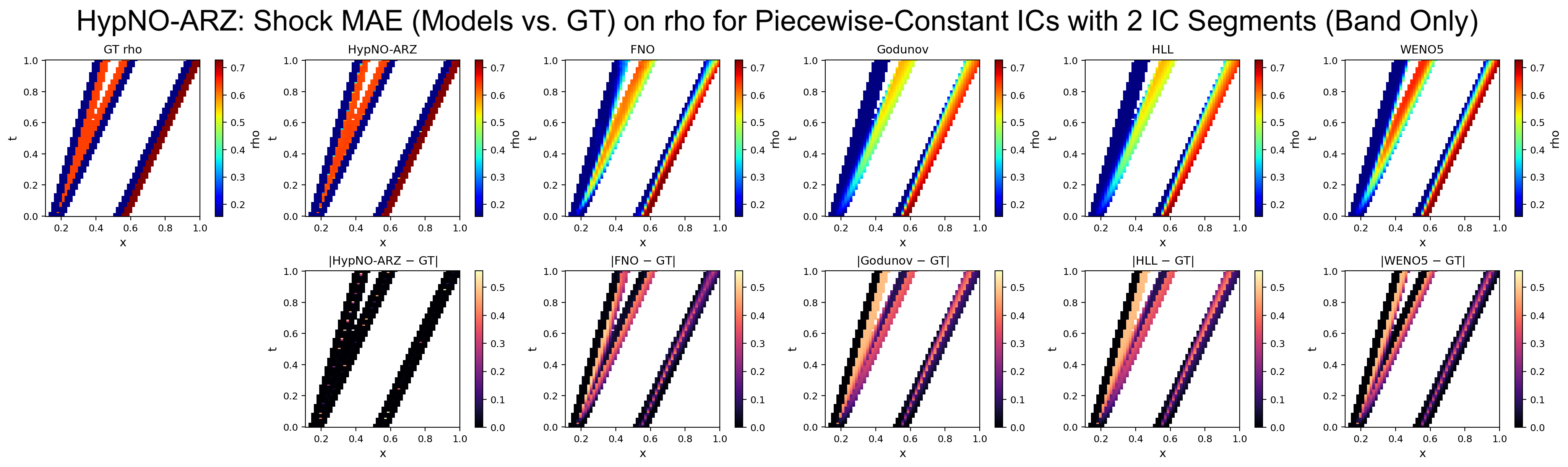}
  \caption{Same ARZ representative sample as Figure~\ref{fig:arz-shock-compare-2}, zoomed to the combined-band bounding box with non-band cells blanked.}
  \label{fig:arz-shock-zoom-2}
\end{figure}
\begin{figure}[!htbp]
  \centering
  \includegraphics[width=0.7\linewidth]{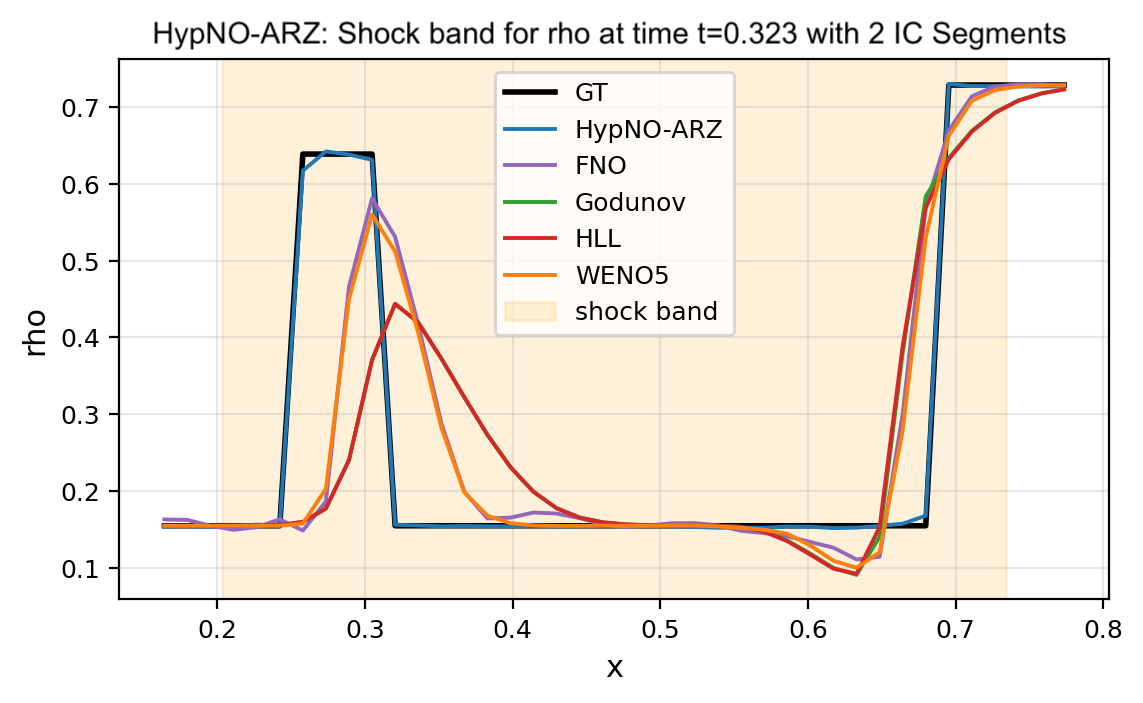}
  \caption{$\rho(x)$ at $t=0.323$ through a band row for ARZ model with $2$ initial discontinuity segments. The shaded region marks the combined shock band.}
  \label{fig:arz-shock-slice-2}
\end{figure}
\begin{figure}[!htbp]
  \centering
  \includegraphics[width=0.9\linewidth]{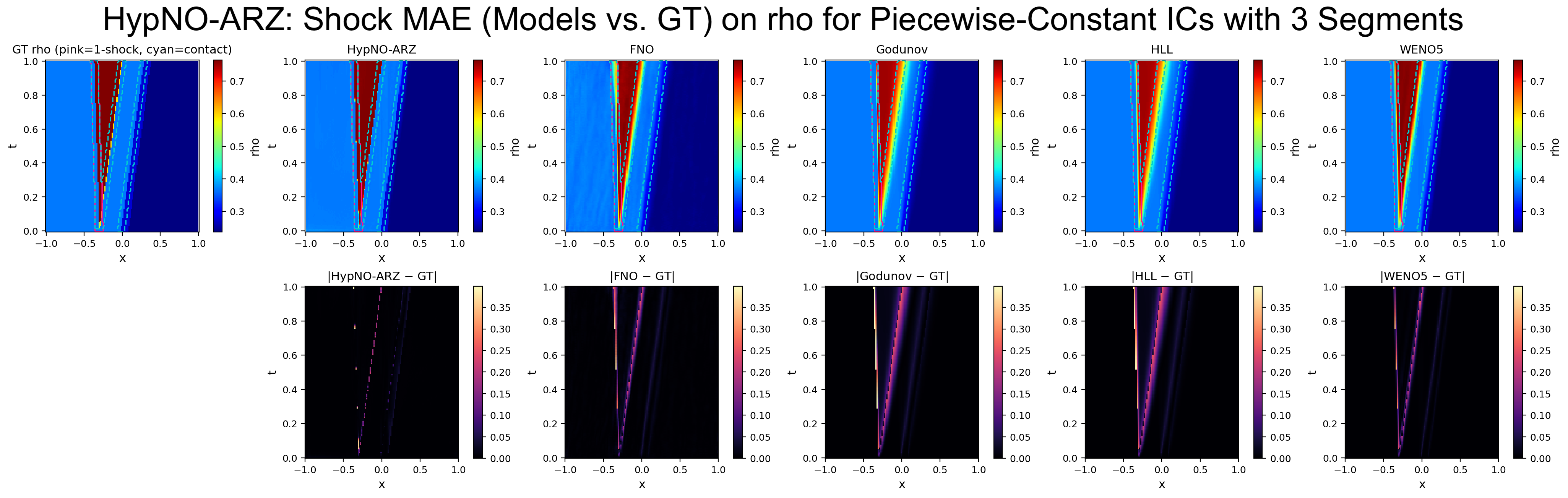}
  \caption{ARZ: Representative sample for an initial condition partitioned into $N=3$ initial discontinuity segments. Top row: $\rho$ fields with the 1-shock band (pink) and contact band (cyan dashed) outlined; bottom row: absolute error to the ground truth.}
  \label{fig:arz-shock-compare-3}
\end{figure}
\begin{figure}[!htbp]
  \centering
  \includegraphics[width=0.9\linewidth]{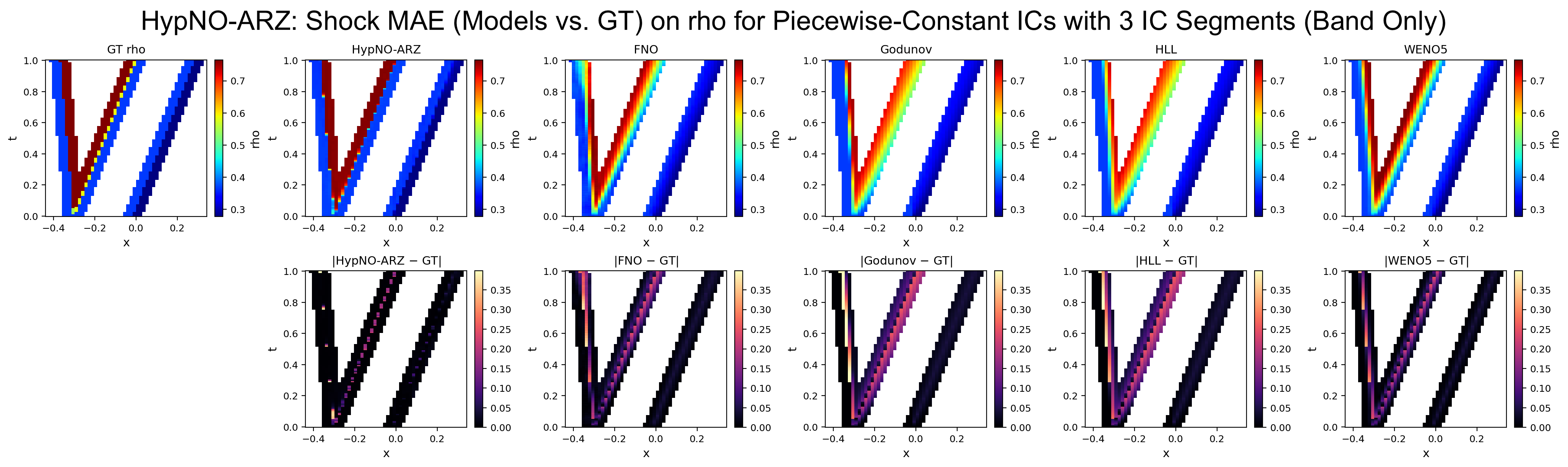}
  \caption{Same ARZ representative sample as Fig.~\ref{fig:arz-shock-compare-3}, zoomed to the combined-band bounding box with non-band cells blanked.}
  \label{fig:arz-shock-zoom-3}
\end{figure}
\begin{figure}[!htbp]
  \centering
  \includegraphics[width=0.7\linewidth]{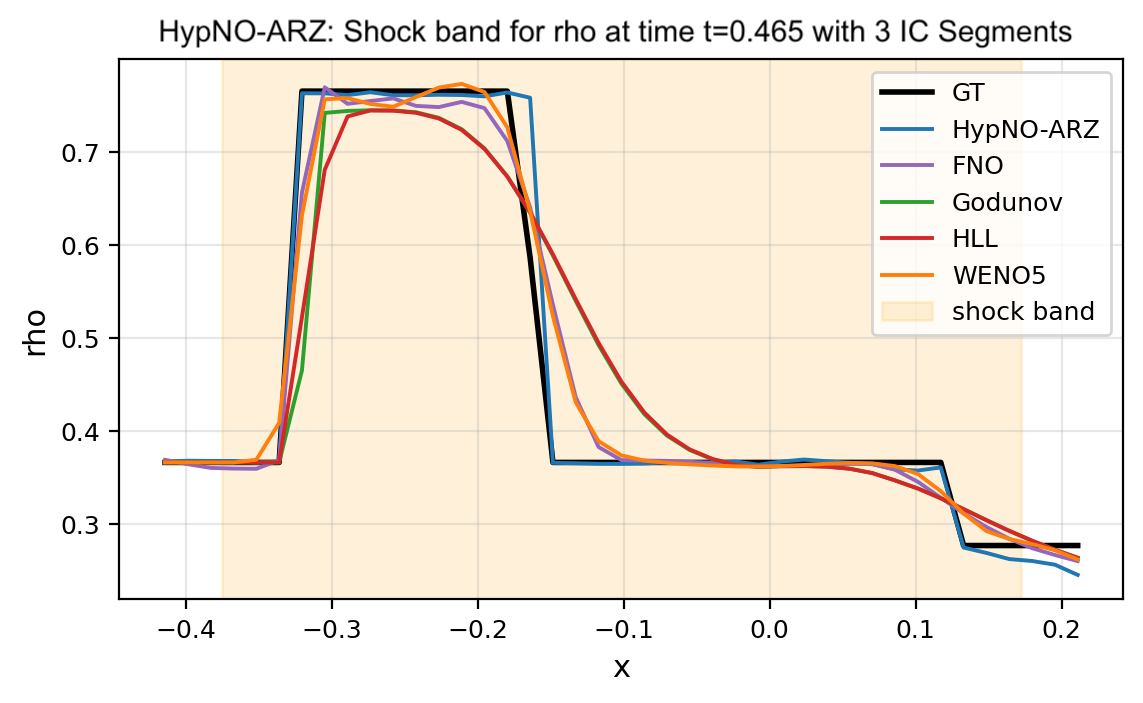}
  \caption{$\rho(x)$ at $t=0.465$ through a band row for ARZ model with $3$ initial discontinuity segments. The shaded region marks the combined shock band.}
  \label{fig:arz-shock-slice-3}
\end{figure}
\begin{figure}[!htbp]
  \centering
  \includegraphics[width=0.9\linewidth]{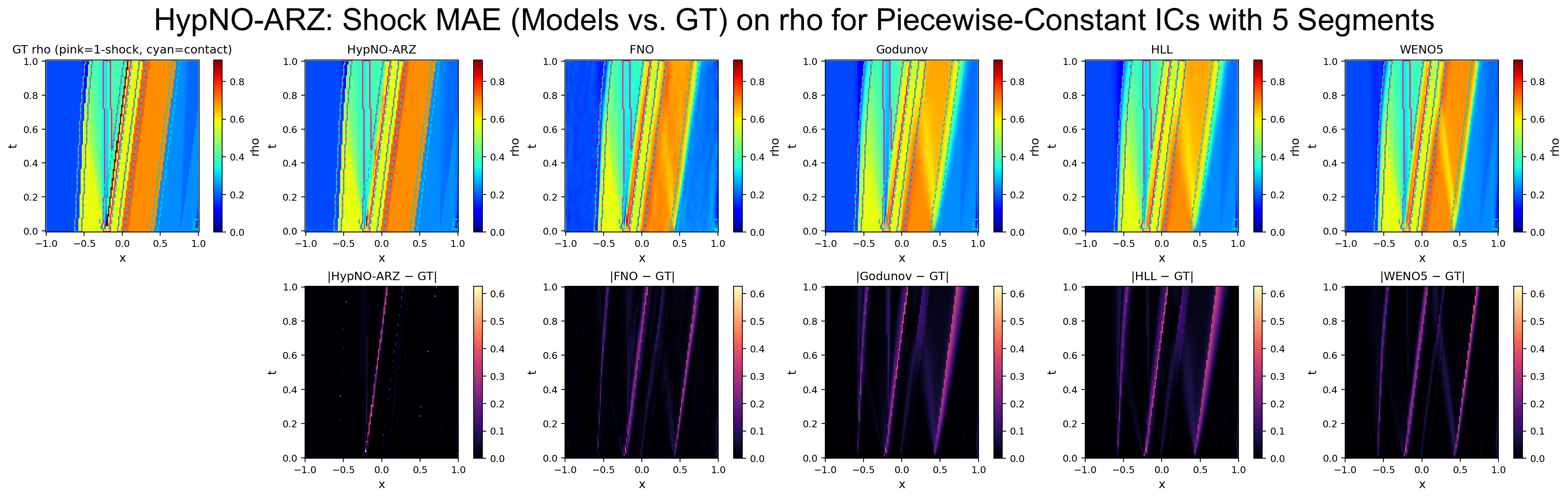}
  \caption{ARZ: Representative sample for an initial condition partitioned into $N=5$ initial discontinuity segments.  Top row: $\rho$ fields with the 1-shock band (pink) and contact band (cyan dashed) outlined; bottom row: absolute error to the ground truth.}
  \label{fig:arz-shock-compare-5}
\end{figure}
\begin{figure}[!htbp]
  \centering
  \includegraphics[width=0.9\linewidth]{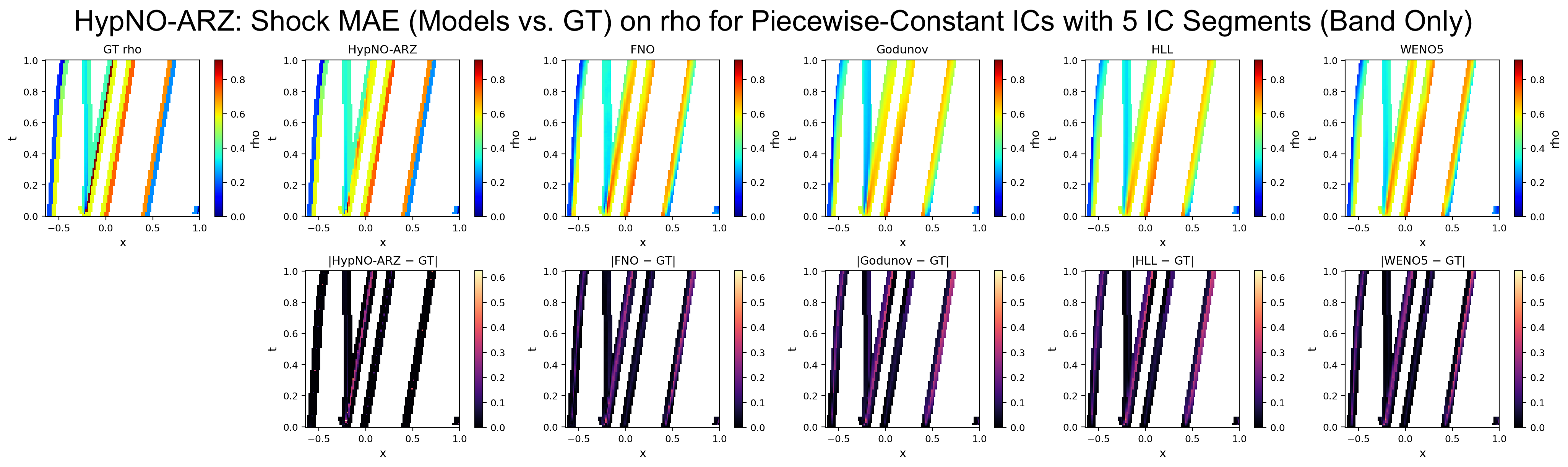}
  \caption{Same ARZ representative sample as Figure ~\ref{fig:arz-shock-compare-5}, zoomed to the combined-band bounding box with non-band cells blanked.}
  \label{fig:arz-shock-zoom-5}
\end{figure}
\begin{figure}[!htbp]
  \centering
  \includegraphics[width=0.7\linewidth]{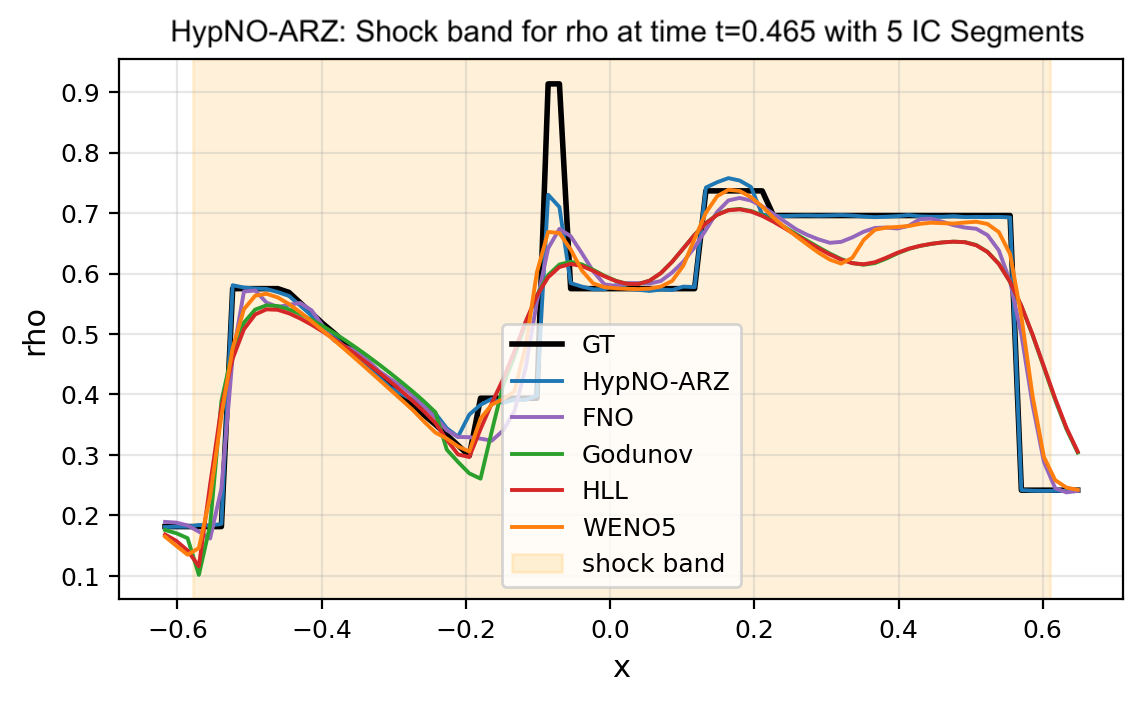}
  \caption{$\rho(x)$ at $t=0.465$ through a band row for ARZ model with $5$ initial discontinuity segments. The shaded region marks the combined shock band.}
  \label{fig:arz-shock-slice-5}
\end{figure}
\begin{figure}[!htbp]
  \centering
  \includegraphics[width=0.9\linewidth]{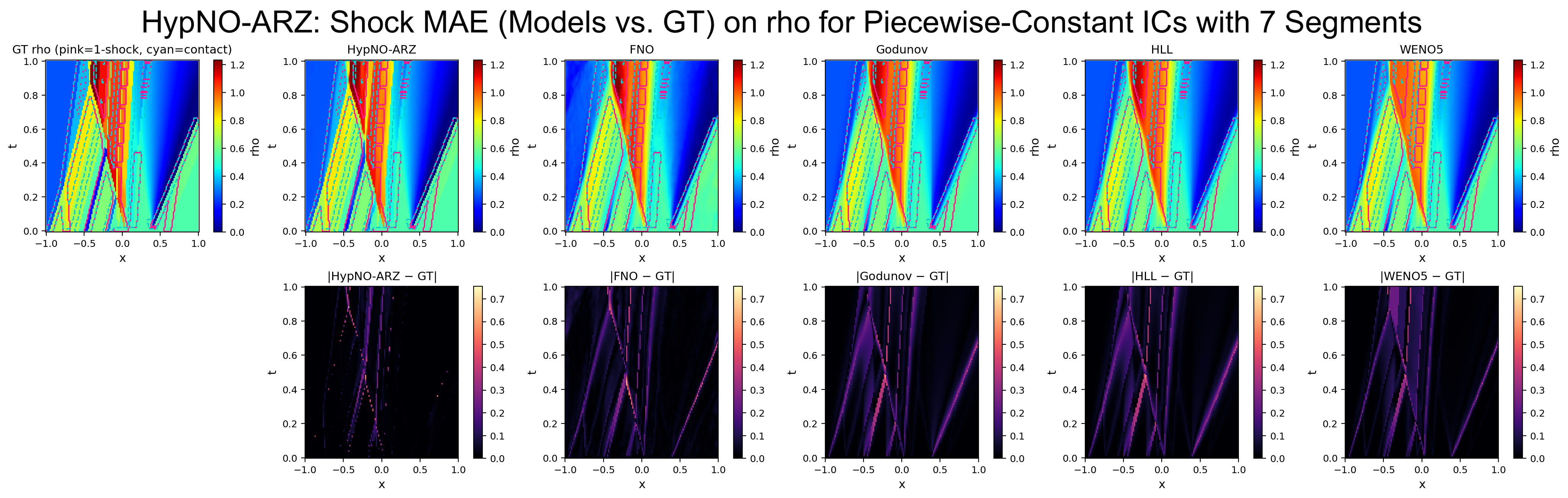}
  \caption{ARZ: Representative sample for an initial condition partitioned into $N=7$ initial discontinuity segments.  Top row: $\rho$ fields with the 1-shock band (pink) and contact band (cyan dashed) outlined; bottom row: absolute error to the ground truth.}
  \label{fig:arz-shock-compare-7}
\end{figure}
\begin{figure}[!htbp]
  \centering
  \includegraphics[width=0.9\linewidth]{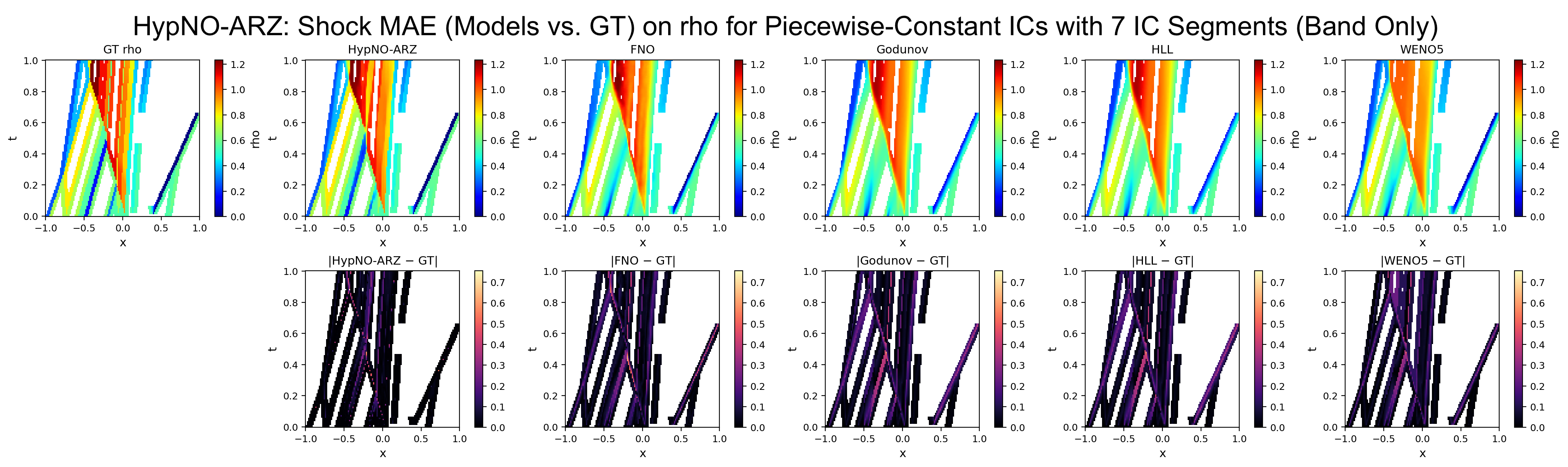}
  \caption{Same ARZ representative sample as Figure~\ref{fig:arz-shock-compare-7}, zoomed to the combined-band bounding box with non-band cells blanked.}
  \label{fig:arz-shock-zoom-7}
\end{figure}
\begin{figure}[!htbp]
  \centering
  \includegraphics[width=0.7\linewidth]{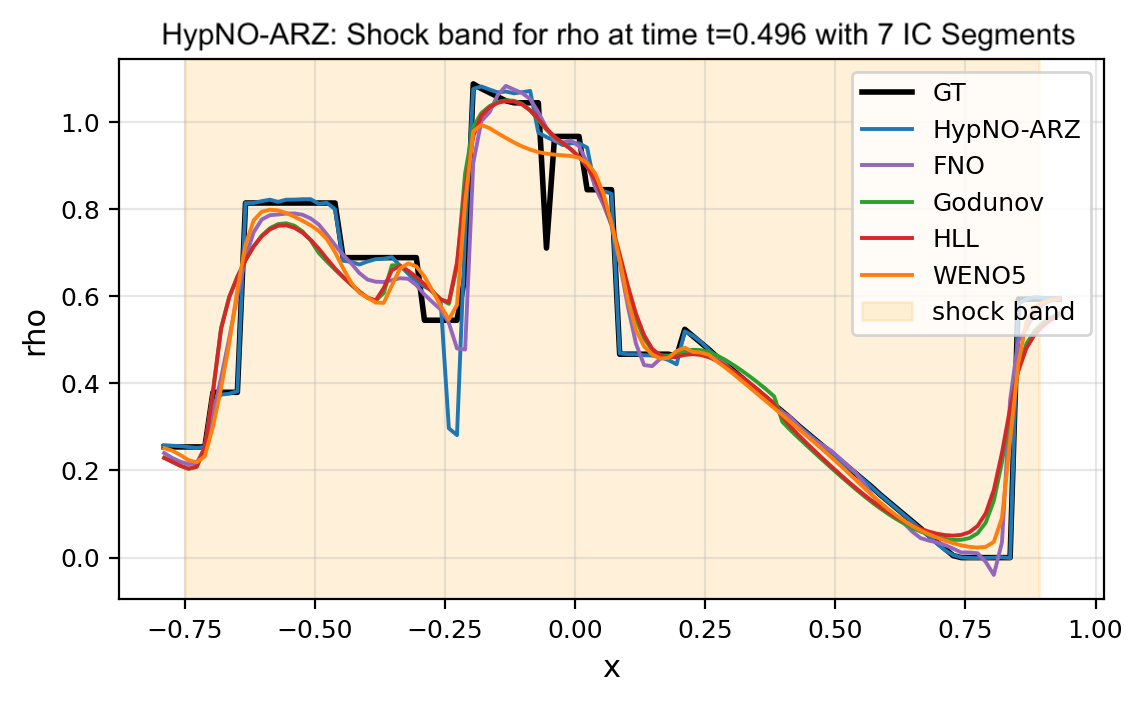}
  \caption{$\rho(x)$ at $t=0.496$ through a band row for ARZ model with $7$ initial discontinuity segments. The shaded region marks the combined shock band.}
  \label{fig:arz-shock-slice-7}
\end{figure}

\begin{figure}[!htbp]
  \centering
  \includegraphics[width=0.9\linewidth]{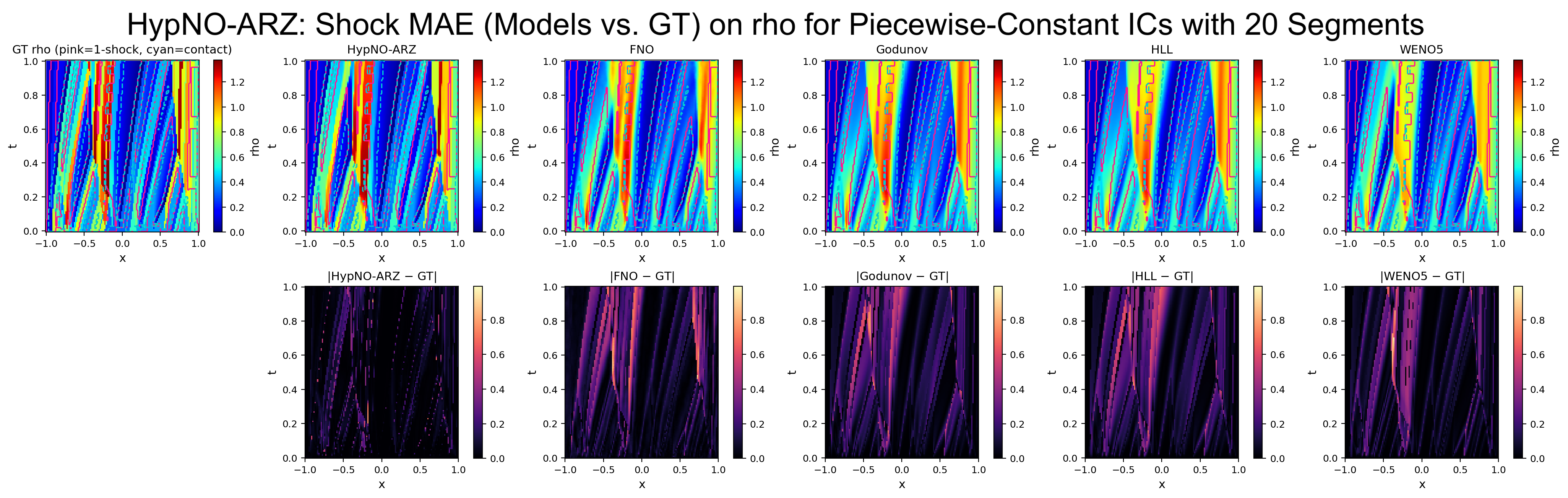}
  \caption{ARZ: Representative sample for an initial condition partitioned into $N=20$ initial discontinuity segments. Top row: $\rho$ fields with the 1-shock band (pink) and contact band (cyan dashed) outlined; bottom row: absolute error to the ground truth.}
  \label{fig:arz-shock-compare-20}
\end{figure}
\begin{figure}[!htbp]
  \centering
  \includegraphics[width=0.9\linewidth]{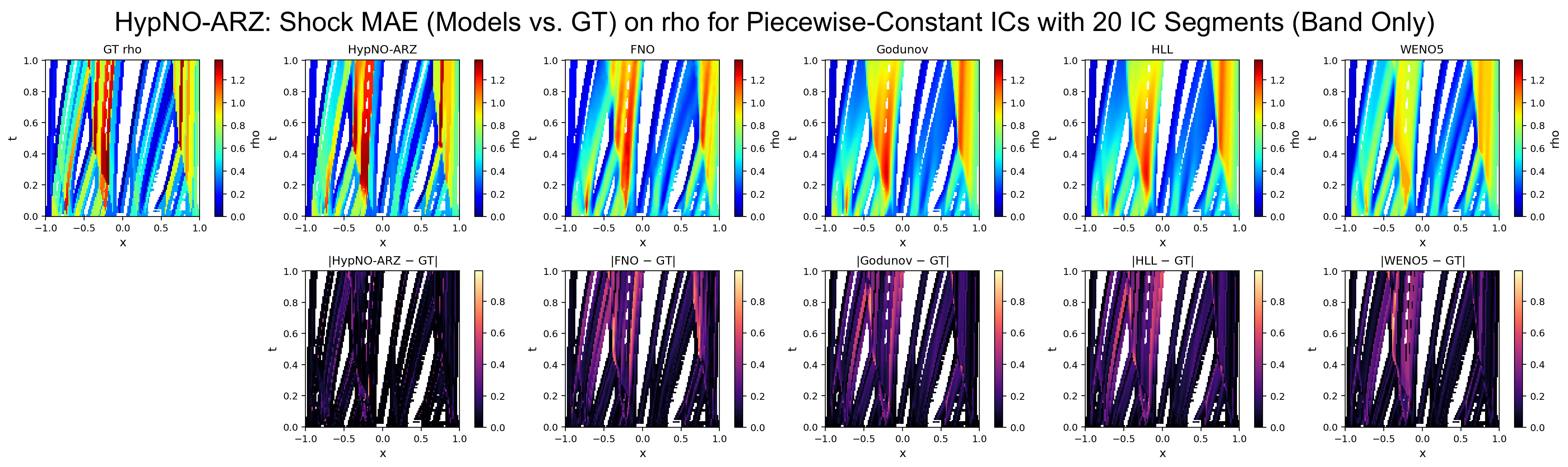}
  \caption{Same ARZ representative sample as Figure~\ref{fig:arz-shock-compare-20}, zoomed to the combined-band bounding box with non-band cells blanked.}
  \label{fig:arz-shock-zoom-20}
\end{figure}
\begin{figure}[!htbp]
  \centering
  \includegraphics[width=0.7\linewidth]{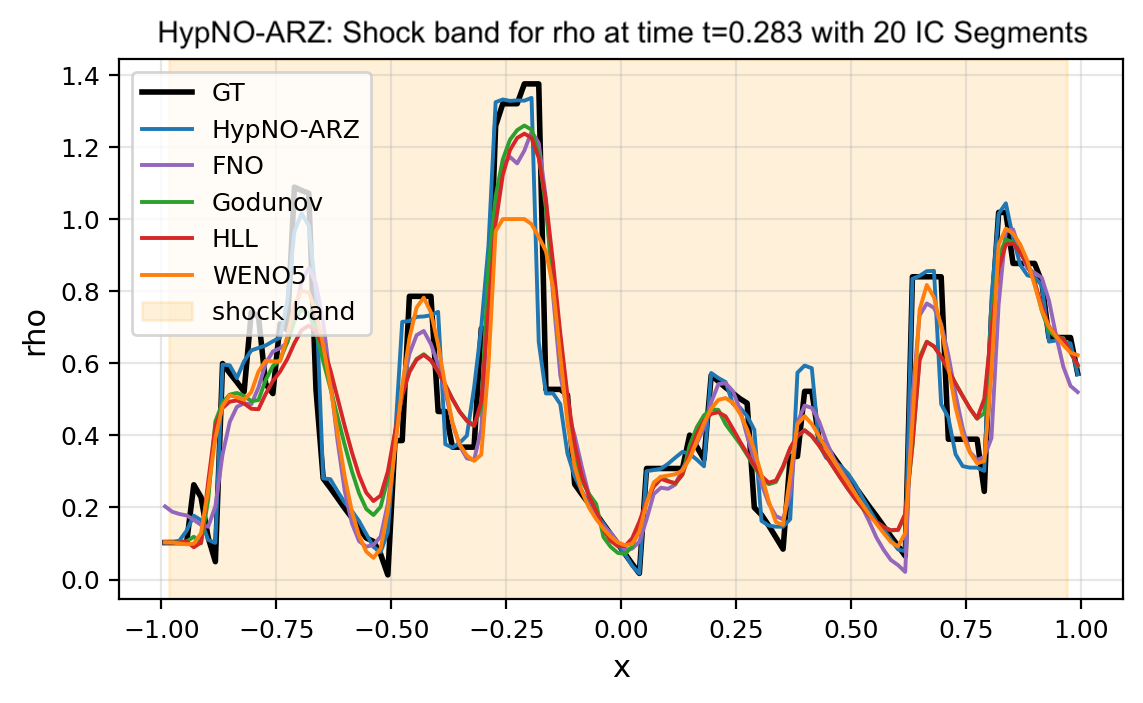}
  \caption{$\rho(x)$ at $t=0.283$ through a band row for ARZ model with $20$ initial discontinuity segments. The shaded region marks the combined shock band.}
  \label{fig:arz-shock-slice-20}
\end{figure}
\begin{figure}[!htbp]
  \centering
  \includegraphics[width=0.9\linewidth]{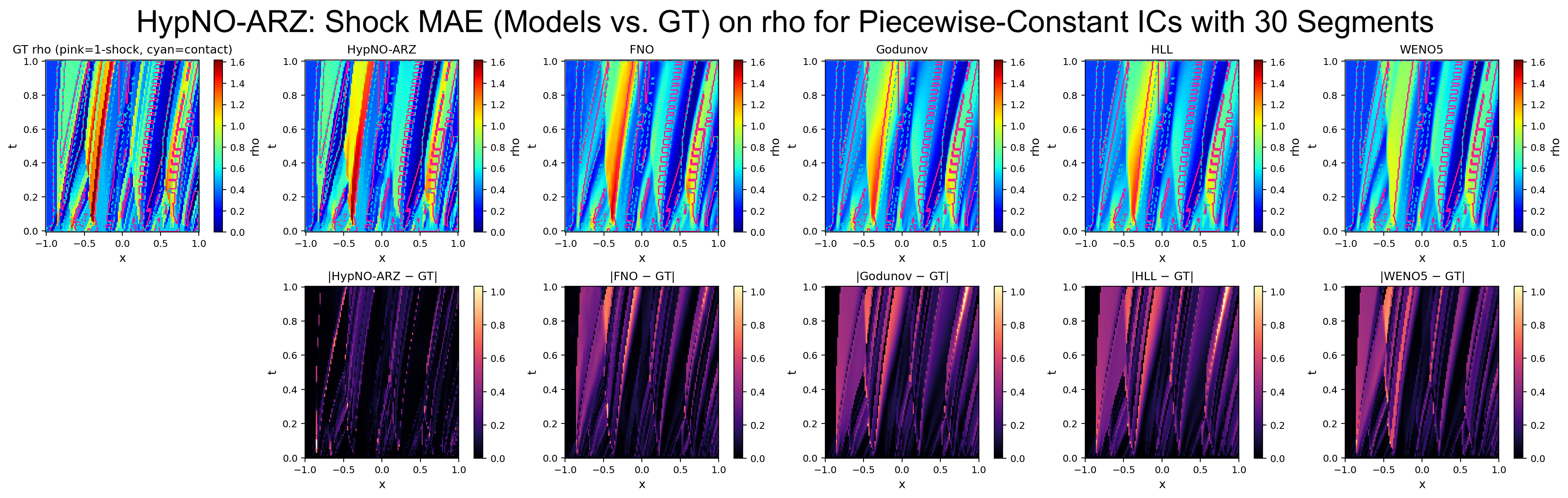}
  \caption{ARZ: Representative sample for an initial condition partitioned into $N=30$ initial discontinuity segments.  Top row: $\rho$ fields with the 1-shock band (pink) and contact band (cyan dashed) outlined; bottom row: absolute error to the ground truth.}
  \label{fig:arz-shock-compare-30}
\end{figure}
\begin{figure}[!htbp]
  \centering
  \includegraphics[width=0.9\linewidth]{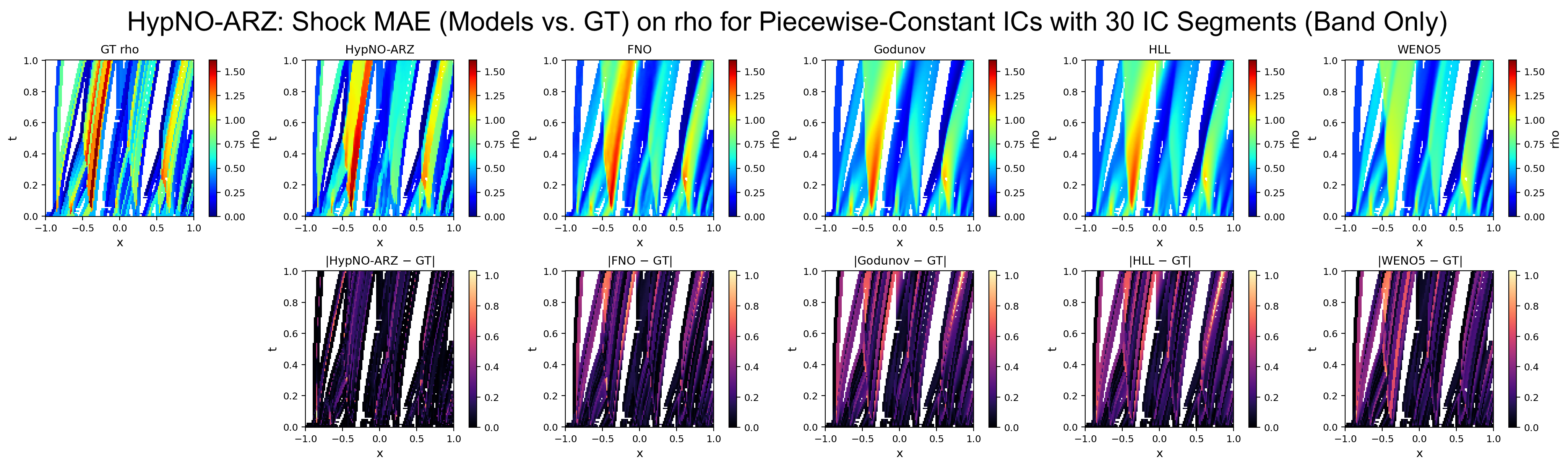}
  \caption{Same ARZ representative representative sample as Figure~\ref{fig:arz-shock-compare-30}, zoomed to the combined-band bounding box with non-band cells blanked.}
  \label{fig:arz-shock-zoom-30}
\end{figure}
\begin{figure}[!htbp]
  \centering
  \includegraphics[width=0.7\linewidth]{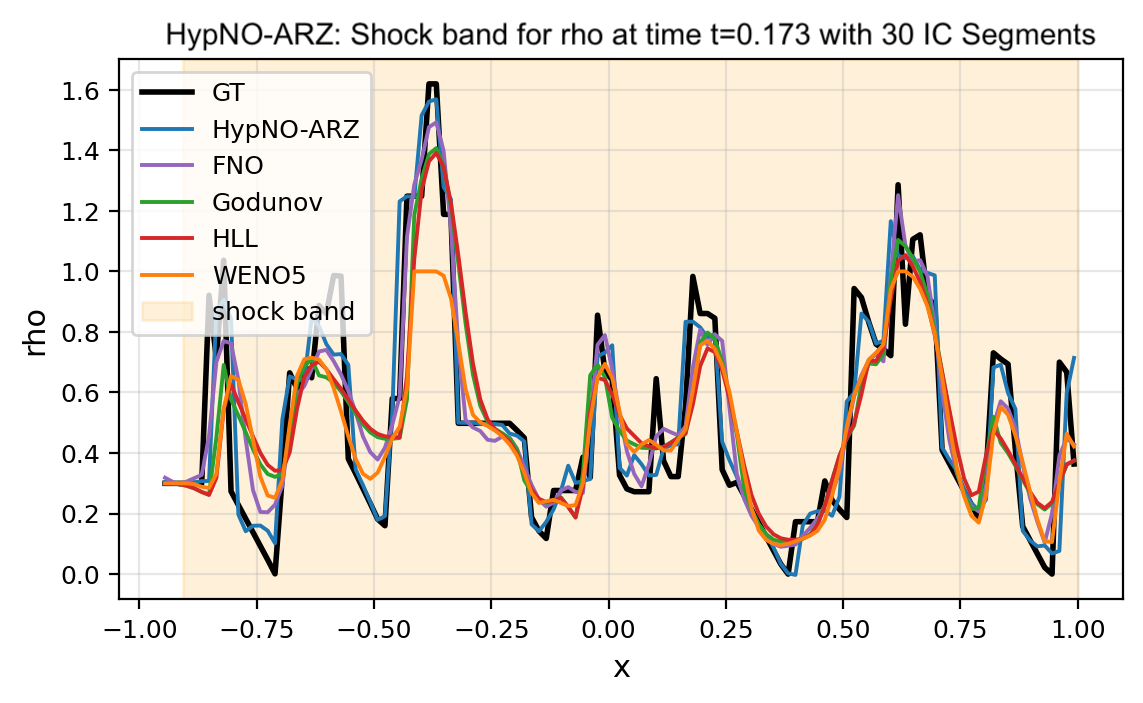}
  \caption{$\rho(x)$ at $t=0.173$ through a band row for ARZ model with $30$ initial discontinuity segments. The shaded region marks the combined shock band.}
  \label{fig:arz-shock-slice-30}
\end{figure}

\FloatBarrier
\subsection{HLL-pretrained variant (HypNO-ARZ-HLL)}
\label{app:arz-hll-eval}

We additionally evaluate the HLL-pretrained variant of the ARZ model
(\emph{HypNO-ARZ-HLL} in \Cref{tab:arz-config}): the same architecture first
trained to convergence on reference solutions generated with the diffusive HLL
scheme, and subsequently fine-tuned, with a freshly initialized optimizer, on
the wave-front-tracking dataset used by the main model
(\Cref{sec:data-construction}). The evaluation protocol is identical to
\Cref{sec:arz-mark0-eval}: the same held-out stratified evaluation set, scored
as MAE against the wave-front-tracking ground truth, with the learned FNO
baseline and the Godunov, HLL, and WENO5 numerical schemes (the \emph{HLL
column} refers to the numerical scheme, not the model).
\Cref{tab:arz_hll_paper_eval_pwc,tab:arz_hll_paper_eval_riemann} report the
per-family breakdown on $\rho$;
\Cref{tab:arz_hll_paper_eval_seg} pools the two families per segment count and
summarizes the in-distribution (ID), out-of-distribution (OOD), and overall
errors; \Cref{tab:arz_hll_paper_eval_channels} reports the same pooled
summaries per channel ($\rho$, $\omega$, $v$).

The HLL-pretrained model is the most accurate method in every cell, with an
overall $\rho$-MAE of $9.02{\times}10^{-3}$
(ID $3.99{\times}10^{-3}$, OOD $1.74{\times}10^{-2}$), matching or modestly
improving on the from-scratch HypNO-ARZ of
\Cref{tab:arz_mark0_paper_eval_piecewise_constant_stratified,tab:arz_mark0_paper_eval_riemann_stratified}
(e.g.\ $4.09{\times}10^{-2}$ vs.\ $5.66{\times}10^{-2}$ on the $30$-segment
piecewise-constant cell). Pretraining on diffusive HLL targets therefore does
not cap the accuracy reachable after fine-tuning on sharp wave-front-tracking
targets.

\begin{center}
\captionof{table}{HypNO-ARZ-HLL density error on the \texttt{piecewise\_constant}
family: mean absolute error on $\rho$ against the wave-front-tracking ground
truth, mean $\pm$ standard deviation per cell. Rows are the number of initial
segments; segment counts $\{2,3,5,7,10\}$ are in-distribution and
$\{8,20,30\}$ are out-of-distribution. Lowest mean MAE per row in bold.}
\label{tab:arz_hll_paper_eval_pwc}
\resizebox{\textwidth}{!}{%
\begin{tabular}{lccccc}
\hline
\# segs & HypNO-ARZ-HLL & FNO & Godunov & HLL & WENO5 \\
\hline
2 & \besterr{1.41{\times}10^{-3}}{4.26{\times}10^{-4}} & \err{2.26{\times}10^{-2}}{2.75{\times}10^{-2}} & \err{2.46{\times}10^{-2}}{3.91{\times}10^{-2}} & \err{2.57{\times}10^{-2}}{4.10{\times}10^{-2}} & \err{2.11{\times}10^{-2}}{3.43{\times}10^{-2}} \\
3 & \besterr{2.25{\times}10^{-3}}{8.45{\times}10^{-4}} & \err{3.50{\times}10^{-2}}{3.17{\times}10^{-2}} & \err{4.10{\times}10^{-2}}{4.54{\times}10^{-2}} & \err{4.32{\times}10^{-2}}{4.88{\times}10^{-2}} & \err{3.99{\times}10^{-2}}{4.75{\times}10^{-2}} \\
5 & \besterr{5.02{\times}10^{-3}}{4.55{\times}10^{-3}} & \err{5.31{\times}10^{-2}}{2.71{\times}10^{-2}} & \err{5.55{\times}10^{-2}}{3.31{\times}10^{-2}} & \err{5.84{\times}10^{-2}}{3.52{\times}10^{-2}} & \err{5.88{\times}10^{-2}}{4.25{\times}10^{-2}} \\
7 & \besterr{6.89{\times}10^{-3}}{3.48{\times}10^{-3}} & \err{6.10{\times}10^{-2}}{2.24{\times}10^{-2}} & \err{6.51{\times}10^{-2}}{3.00{\times}10^{-2}} & \err{6.82{\times}10^{-2}}{3.15{\times}10^{-2}} & \err{6.91{\times}10^{-2}}{3.89{\times}10^{-2}} \\
8 & \besterr{8.09{\times}10^{-3}}{4.64{\times}10^{-3}} & \err{6.50{\times}10^{-2}}{2.70{\times}10^{-2}} & \err{7.05{\times}10^{-2}}{3.51{\times}10^{-2}} & \err{7.39{\times}10^{-2}}{3.73{\times}10^{-2}} & \err{7.03{\times}10^{-2}}{4.12{\times}10^{-2}} \\
10 & \besterr{1.11{\times}10^{-2}}{6.31{\times}10^{-3}} & \err{7.54{\times}10^{-2}}{3.25{\times}10^{-2}} & \err{7.74{\times}10^{-2}}{4.41{\times}10^{-2}} & \err{8.11{\times}10^{-2}}{4.58{\times}10^{-2}} & \err{8.49{\times}10^{-2}}{4.47{\times}10^{-2}} \\
20 & \besterr{2.74{\times}10^{-2}}{1.55{\times}10^{-2}} & \err{1.04{\times}10^{-1}}{5.97{\times}10^{-2}} & \err{1.08{\times}10^{-1}}{7.69{\times}10^{-2}} & \err{1.13{\times}10^{-1}}{8.04{\times}10^{-2}} & \err{1.11{\times}10^{-1}}{6.42{\times}10^{-2}} \\
30 & \besterr{4.09{\times}10^{-2}}{2.55{\times}10^{-2}} & \err{1.25{\times}10^{-1}}{8.40{\times}10^{-2}} & \err{1.26{\times}10^{-1}}{1.02{\times}10^{-1}} & \err{1.31{\times}10^{-1}}{1.05{\times}10^{-1}} & \err{1.38{\times}10^{-1}}{8.83{\times}10^{-2}} \\
\hline
\end{tabular}}
\end{center}

\begin{center}
\captionof{table}{HypNO-ARZ-HLL density error on the \texttt{riemann} family: mean
absolute error on $\rho$ against the exact Riemann-solver ground truth, mean
$\pm$ standard deviation per cell. As in
\Cref{tab:arz_mark0_paper_eval_riemann_stratified}, rows index the stratified
evaluation bins (a single discontinuity per sample), so all cells are
in-distribution. Lowest mean MAE per row in bold.}
\label{tab:arz_hll_paper_eval_riemann}
\resizebox{\textwidth}{!}{%
\begin{tabular}{lccccc}
\hline
Bin & HypNO-ARZ-HLL & FNO & Godunov & HLL & WENO5 \\
\hline
1 & \besterr{1.28{\times}10^{-3}}{4.52{\times}10^{-4}} & \err{2.66{\times}10^{-2}}{3.94{\times}10^{-2}} & \err{2.51{\times}10^{-2}}{5.50{\times}10^{-2}} & \err{2.75{\times}10^{-2}}{5.81{\times}10^{-2}} & \err{2.47{\times}10^{-2}}{4.94{\times}10^{-2}} \\
2 & \besterr{1.26{\times}10^{-3}}{3.96{\times}10^{-4}} & \err{2.58{\times}10^{-2}}{2.53{\times}10^{-2}} & \err{2.12{\times}10^{-2}}{3.43{\times}10^{-2}} & \err{2.31{\times}10^{-2}}{3.58{\times}10^{-2}} & \err{2.32{\times}10^{-2}}{3.75{\times}10^{-2}} \\
3 & \besterr{1.35{\times}10^{-3}}{5.13{\times}10^{-4}} & \err{2.98{\times}10^{-2}}{3.78{\times}10^{-2}} & \err{2.80{\times}10^{-2}}{4.71{\times}10^{-2}} & \err{3.19{\times}10^{-2}}{5.46{\times}10^{-2}} & \err{3.51{\times}10^{-2}}{5.50{\times}10^{-2}} \\
4 & \besterr{1.32{\times}10^{-3}}{5.04{\times}10^{-4}} & \err{2.50{\times}10^{-2}}{2.69{\times}10^{-2}} & \err{2.36{\times}10^{-2}}{3.57{\times}10^{-2}} & \err{2.51{\times}10^{-2}}{4.02{\times}10^{-2}} & \err{2.46{\times}10^{-2}}{4.65{\times}10^{-2}} \\
5 & \besterr{1.28{\times}10^{-3}}{3.81{\times}10^{-4}} & \err{2.43{\times}10^{-2}}{3.15{\times}10^{-2}} & \err{2.25{\times}10^{-2}}{3.76{\times}10^{-2}} & \err{2.54{\times}10^{-2}}{4.37{\times}10^{-2}} & \err{2.41{\times}10^{-2}}{4.38{\times}10^{-2}} \\
6 & \besterr{1.30{\times}10^{-3}}{4.72{\times}10^{-4}} & \err{1.99{\times}10^{-2}}{1.78{\times}10^{-2}} & \err{1.63{\times}10^{-2}}{2.11{\times}10^{-2}} & \err{1.77{\times}10^{-2}}{2.43{\times}10^{-2}} & \err{2.11{\times}10^{-2}}{3.73{\times}10^{-2}} \\
7 & \besterr{1.28{\times}10^{-3}}{4.60{\times}10^{-4}} & \err{2.90{\times}10^{-2}}{3.40{\times}10^{-2}} & \err{2.78{\times}10^{-2}}{4.93{\times}10^{-2}} & \err{3.00{\times}10^{-2}}{5.15{\times}10^{-2}} & \err{3.12{\times}10^{-2}}{4.99{\times}10^{-2}} \\
8 & \besterr{1.24{\times}10^{-3}}{4.06{\times}10^{-4}} & \err{2.17{\times}10^{-2}}{2.57{\times}10^{-2}} & \err{2.04{\times}10^{-2}}{3.81{\times}10^{-2}} & \err{2.15{\times}10^{-2}}{3.94{\times}10^{-2}} & \err{2.25{\times}10^{-2}}{4.51{\times}10^{-2}} \\
\hline
\end{tabular}}
\end{center}

\begin{center}
\captionof{table}{HypNO-ARZ-HLL: per-num\_segments MAE on $\rho$, mean $\pm$ std,
pooled across IC families. The ID row pools the in-distribution cells
(ic\_type, num\_segments both present in training), the OOD row the rest, and
the final row the entire evaluation set. Lowest mean MAE per row in bold.}
\label{tab:arz_hll_paper_eval_seg}
\resizebox{\textwidth}{!}{%
\begin{tabular}{lccccc}
\hline
num\_segments & HypNO-ARZ-HLL & FNO & Godunov & HLL & WENO5 \\
\hline
2 & \besterr{1.37{\times}10^{-3}}{4.39{\times}10^{-4}} & \err{2.39{\times}10^{-2}}{3.20{\times}10^{-2}} & \err{2.47{\times}10^{-2}}{4.50{\times}10^{-2}} & \err{2.63{\times}10^{-2}}{4.74{\times}10^{-2}} & \err{2.23{\times}10^{-2}}{4.00{\times}10^{-2}} \\
3 & \besterr{1.92{\times}10^{-3}}{8.62{\times}10^{-4}} & \err{3.19{\times}10^{-2}}{3.00{\times}10^{-2}} & \err{3.44{\times}10^{-2}}{4.31{\times}10^{-2}} & \err{3.65{\times}10^{-2}}{4.59{\times}10^{-2}} & \err{3.43{\times}10^{-2}}{4.51{\times}10^{-2}} \\
5 & \besterr{3.79{\times}10^{-3}}{4.11{\times}10^{-3}} & \err{4.54{\times}10^{-2}}{3.29{\times}10^{-2}} & \err{4.63{\times}10^{-2}}{4.05{\times}10^{-2}} & \err{4.96{\times}10^{-2}}{4.44{\times}10^{-2}} & \err{5.09{\times}10^{-2}}{4.84{\times}10^{-2}} \\
7 & \besterr{5.03{\times}10^{-3}}{3.88{\times}10^{-3}} & \err{4.90{\times}10^{-2}}{2.94{\times}10^{-2}} & \err{5.13{\times}10^{-2}}{3.75{\times}10^{-2}} & \err{5.39{\times}10^{-2}}{4.01{\times}10^{-2}} & \err{5.43{\times}10^{-2}}{4.66{\times}10^{-2}} \\
8 & \besterr{5.82{\times}10^{-3}}{4.97{\times}10^{-3}} & \err{5.15{\times}10^{-2}}{3.44{\times}10^{-2}} & \err{5.45{\times}10^{-2}}{4.25{\times}10^{-2}} & \err{5.77{\times}10^{-2}}{4.57{\times}10^{-2}} & \err{5.49{\times}10^{-2}}{4.74{\times}10^{-2}} \\
10 & \besterr{7.85{\times}10^{-3}}{6.93{\times}10^{-3}} & \err{5.69{\times}10^{-2}}{3.86{\times}10^{-2}} & \err{5.70{\times}10^{-2}}{4.77{\times}10^{-2}} & \err{5.99{\times}10^{-2}}{4.99{\times}10^{-2}} & \err{6.36{\times}10^{-2}}{5.19{\times}10^{-2}} \\
20 & \besterr{1.87{\times}10^{-2}}{1.77{\times}10^{-2}} & \err{7.88{\times}10^{-2}}{6.33{\times}10^{-2}} & \err{8.11{\times}10^{-2}}{7.85{\times}10^{-2}} & \err{8.51{\times}10^{-2}}{8.19{\times}10^{-2}} & \err{8.42{\times}10^{-2}}{7.06{\times}10^{-2}} \\
30 & \besterr{2.77{\times}10^{-2}}{2.80{\times}10^{-2}} & \err{9.07{\times}10^{-2}}{8.55{\times}10^{-2}} & \err{9.08{\times}10^{-2}}{9.93{\times}10^{-2}} & \err{9.44{\times}10^{-2}}{1.03{\times}10^{-1}} & \err{9.93{\times}10^{-2}}{9.39{\times}10^{-2}} \\
\hline
ID & \besterr{3.99{\times}10^{-3}}{4.65{\times}10^{-3}} & \err{4.14{\times}10^{-2}}{3.49{\times}10^{-2}} & \err{4.28{\times}10^{-2}}{4.45{\times}10^{-2}} & \err{4.52{\times}10^{-2}}{4.73{\times}10^{-2}} & \err{4.51{\times}10^{-2}}{4.89{\times}10^{-2}} \\
OOD & \besterr{1.74{\times}10^{-2}}{2.13{\times}10^{-2}} & \err{7.37{\times}10^{-2}}{6.66{\times}10^{-2}} & \err{7.55{\times}10^{-2}}{7.86{\times}10^{-2}} & \err{7.91{\times}10^{-2}}{8.17{\times}10^{-2}} & \err{7.95{\times}10^{-2}}{7.54{\times}10^{-2}} \\
\hline
all & \besterr{9.02{\times}10^{-3}}{1.50{\times}10^{-2}} & \err{5.35{\times}10^{-2}}{5.16{\times}10^{-2}} & \err{5.50{\times}10^{-2}}{6.17{\times}10^{-2}} & \err{5.79{\times}10^{-2}}{6.46{\times}10^{-2}} & \err{5.80{\times}10^{-2}}{6.25{\times}10^{-2}} \\
\hline
\end{tabular}}
\end{center}

\begin{center}
\captionof{table}{HypNO-ARZ-HLL: per-channel MAE (mean $\pm$ std) on $\rho$, the
Riemann invariant $\omega$, and the velocity $v$, pooled by ID/OOD subset and
over the entire evaluation set. Lowest mean MAE per row in bold.}
\label{tab:arz_hll_paper_eval_channels}
\resizebox{\textwidth}{!}{%
\begin{tabular}{llccccc}
\hline
channel & subset & HypNO-ARZ-HLL & FNO & Godunov & HLL & WENO5 \\
\hline
$\rho$ & ID & \besterr{3.99{\times}10^{-3}}{4.65{\times}10^{-3}} & \err{4.14{\times}10^{-2}}{3.49{\times}10^{-2}} & \err{4.28{\times}10^{-2}}{4.45{\times}10^{-2}} & \err{4.52{\times}10^{-2}}{4.73{\times}10^{-2}} & \err{4.51{\times}10^{-2}}{4.89{\times}10^{-2}} \\
$\rho$ & OOD & \besterr{1.74{\times}10^{-2}}{2.13{\times}10^{-2}} & \err{7.37{\times}10^{-2}}{6.66{\times}10^{-2}} & \err{7.55{\times}10^{-2}}{7.86{\times}10^{-2}} & \err{7.91{\times}10^{-2}}{8.17{\times}10^{-2}} & \err{7.95{\times}10^{-2}}{7.54{\times}10^{-2}} \\
$\rho$ & all & \besterr{9.02{\times}10^{-3}}{1.50{\times}10^{-2}} & \err{5.35{\times}10^{-2}}{5.16{\times}10^{-2}} & \err{5.50{\times}10^{-2}}{6.17{\times}10^{-2}} & \err{5.79{\times}10^{-2}}{6.46{\times}10^{-2}} & \err{5.80{\times}10^{-2}}{6.25{\times}10^{-2}} \\
\hline
$\omega$ & ID & \besterr{3.40{\times}10^{-3}}{3.78{\times}10^{-3}} & \err{3.82{\times}10^{-2}}{3.18{\times}10^{-2}} & \err{4.12{\times}10^{-2}}{3.93{\times}10^{-2}} & \err{4.33{\times}10^{-2}}{4.11{\times}10^{-2}} & \err{3.92{\times}10^{-2}}{3.81{\times}10^{-2}} \\
$\omega$ & OOD & \besterr{1.48{\times}10^{-2}}{1.96{\times}10^{-2}} & \err{7.10{\times}10^{-2}}{6.56{\times}10^{-2}} & \err{7.62{\times}10^{-2}}{7.97{\times}10^{-2}} & \err{7.94{\times}10^{-2}}{8.25{\times}10^{-2}} & \err{7.26{\times}10^{-2}}{6.88{\times}10^{-2}} \\
$\omega$ & all & \besterr{7.66{\times}10^{-3}}{1.35{\times}10^{-2}} & \err{5.05{\times}10^{-2}}{5.00{\times}10^{-2}} & \err{5.43{\times}10^{-2}}{6.03{\times}10^{-2}} & \err{5.68{\times}10^{-2}}{6.26{\times}10^{-2}} & \err{5.17{\times}10^{-2}}{5.42{\times}10^{-2}} \\
\hline
$v$ & ID & \besterr{2.80{\times}10^{-3}}{2.03{\times}10^{-3}} & \err{3.33{\times}10^{-2}}{2.67{\times}10^{-2}} & \err{3.37{\times}10^{-2}}{3.40{\times}10^{-2}} & \err{3.47{\times}10^{-2}}{3.54{\times}10^{-2}} & \err{2.43{\times}10^{-2}}{2.62{\times}10^{-2}} \\
$v$ & OOD & \besterr{7.42{\times}10^{-3}}{6.09{\times}10^{-3}} & \err{4.97{\times}10^{-2}}{3.73{\times}10^{-2}} & \err{5.62{\times}10^{-2}}{5.70{\times}10^{-2}} & \err{5.79{\times}10^{-2}}{5.87{\times}10^{-2}} & \err{4.05{\times}10^{-2}}{4.03{\times}10^{-2}} \\
$v$ & all & \besterr{4.53{\times}10^{-3}}{4.64{\times}10^{-3}} & \err{3.94{\times}10^{-2}}{3.21{\times}10^{-2}} & \err{4.22{\times}10^{-2}}{4.54{\times}10^{-2}} & \err{4.34{\times}10^{-2}}{4.69{\times}10^{-2}} & \err{3.04{\times}10^{-2}}{3.32{\times}10^{-2}} \\
\hline
\end{tabular}}
\end{center}

\FloatBarrier
\section{Shock-neighborhood detection}
\label{app:shock-band}
The shock-neighborhood masks used in the shock evaluations are detected once on
the exact ground truth and reused, unchanged, for every method, so all methods are
scored on the identical set of cells.

\paragraph{LWR}
A grid cell $(t_k,x_i)$ is flagged as a shock cell when a signed-jump test and a
local total-variation test both pass. The signed central difference
\begin{equation}
\label{eq:shock-detector-jump}
\tfrac{1}{2}\bigl(\rho(x_{i+1}, t_k)-\rho(x_{i-1}, t_k)\bigr) \;>\; \tau,
\end{equation}
with $\tau=0.06$, encodes the Lax entropy condition $\rho_L<\rho_R$ for the concave
flux $f(\rho)=\rho(1-\rho)$, so rarefactions are rejected; the windowed total variation
\begin{equation}
\label{eq:shock-detector-tv}
\sum_{j=i-b}^{i+b}\bigl|\rho(x_{j+1}, t_k)-\rho(x_j,t_k)\bigr| \;>\; \alpha\,\tau,
\end{equation}
with half-width $b=2$ and multiplier $\alpha=1.5$, requires the flagged cell to
sit on a localized cluster of large face differences rather than an isolated
smooth-region blip. The surviving cells are dilated along $x$ by $\pm b$ cells to
form the \emph{shock neighborhood}, capturing the smeared transition region into
which the discontinuity is numerically resolved and within which any method's
error concentrates.

\paragraph{ARZ}
Because ARZ carries two wave families, two bands are detected separately on the
ground truth, and then combined. The \emph{1-shock} band is flagged on the
genuinely nonlinear field via the Lax-1 condition
$\lambda_{1,L}>\lambda_{1,R}$ on the eigenvalue
$\lambda_1=v-\rho\,p'(\rho)$ (with $p(\rho)=\rho$), where both $\lambda_1$ and
$v$ drop across the interface; the \emph{2-contact} band is flagged on $\rho$
where the density jumps but $v$ stays continuous. As in the LWR case, a
total-variation gate suppresses smooth-region blips, and the surviving cells are
dilated by $\pm2$ cells (threshold $\tau=0.06$). The same masks are reused across
all methods.



\bibliographystyle{siam}
\bibliography{refs}

@book{leveque2002fv,
  author    = {LeVeque, Randall J.},
  title     = {Finite Volume Methods for Hyperbolic Problems},
  publisher = {Cambridge University Press},
  year      = {2002}
}

@book{toro2009riemann,
  author    = {Toro, Eleuterio F.},
  title     = {Riemann Solvers and Numerical Methods for Fluid Dynamics: A Practical Introduction},
  edition   = {3},
  publisher = {Springer},
  year      = {2009}
}

@incollection{shu2009weno,
  author    = {Shu, Chi-Wang},
  title     = {High Order Weighted Essentially Nonoscillatory Schemes for Convection Dominated Problems},
  booktitle = {SIAM Review},
  volume    = {51},
  issue    = {1},
  pages     = {82--126},
  year      = {2009},
  publisher = {Society for Industrial and Applied Mathematics}
}

@article{lighthill1955lwr,
  author  = {Lighthill, M. J. and Whitham, G. B.},
  title   = {On Kinematic Waves {II}. A Theory of Traffic Flow on Long Crowded Roads},
  journal = {Proceedings of the Royal Society of London. Series A},
  volume  = {229},
  number  = {1178},
  pages   = {317--345},
  year    = {1955}
}

@article{richards1956traffic,
  author  = {Richards, Paul I.},
  title   = {Shock Waves on the Highway},
  journal = {Operations Research},
  volume  = {4},
  number  = {1},
  pages   = {42--51},
  year    = {1956}
}

@article{aw2000arz,
  author  = {Aw, A. and Rascle, M.},
  title   = {Resurrection of ``Second Order'' Models of Traffic Flow},
  journal = {SIAM Journal on Applied Mathematics},
  volume  = {60},
  number  = {3},
  pages   = {916--938},
  year    = {2000}
}

@book{front2015holden,
  author = {Helge Holden and Nils Henrik Risebro},
  year = {2015},
  title = {Front Tracking for Hyperbolic Conservation Laws},
  publisher = {Springer-Verlag Berlin Heidelberg}
}

@article{raissi2019pinns,
  author  = {Raissi, Maziar and Perdikaris, Paris and Karniadakis, George Em},
  title   = {Physics-Informed Neural Networks: A Deep Learning Framework for Solving Forward and Inverse Problems Involving Nonlinear Partial Differential Equations},
  journal = {Journal of Computational Physics},
  volume  = {378},
  pages   = {686--707},
  year    = {2019},
  doi     = {10.1016/j.jcp.2018.10.045}
}

@article{kharazmi2019vpinn,
  author  = {Kharazmi, Ehsan and Zhang, Zhongqiang and Karniadakis, George Em},
  title   = {Variational Physics-Informed Neural Networks for Solving Partial Differential Equations},
  journal = {arXiv preprint arXiv:1912.00873},
  year    = {2019}
}

@article{karniadakis2021physics,
  author  = {Karniadakis, George Em and Kevrekidis, Ioannis G. and Lu, Lu and Perdikaris, Paris and Wang, Sifan and Yang, Liu},
  title   = {Physics-Informed Machine Learning},
  journal = {Nature Reviews Physics},
  volume  = {3},
  number  = {6},
  pages   = {422--440},
  year    = {2021}
}

@misc{khodakarami2026spectralbias,
  author        = {Khodakarami, Siavash and Oommen, Vivek and Daryakenari, Nazanin Ahmadi and Beekenkamp, Maxim and Karniadakis, George Em},
  title         = {Spectral bias in physics-informed and operator learning: Analysis and mitigation guidelines},
  year          = {2026},
  eprint        = {2602.19265},
  archivePrefix = {arXiv},
  primaryClass  = {cs.LG}
}

@article{lu2021deeponet,
  author  = {Lu, Lu and Jin, Pengzhan and Karniadakis, George Em},
  title   = {DeepONet: Learning Nonlinear Operators for Identifying Differential Equations Based on the Universal Approximation Theorem of Operators},
  journal = {Nature Machine Intelligence},
  volume  = {3},
  number  = {3},
  pages   = {218--229},
  year    = {2021}
}

@inproceedings{li2021fno,
  author    = {Li, Zongyi and Kovachki, Nikola and Azizzadenesheli, Kamyar and Liu, Burigede and Bhattacharya, Kaushik and Stuart, Andrew and Anandkumar, Anima},
  title     = {Fourier Neural Operator for Parametric Partial Differential Equations},
  booktitle = {International Conference on Learning Representations},
  year      = {2021},
  pages = {1--16}
}

@article{kovachki2021neuraloperator,
  author  = {Kovachki, Nikola and Li, Zongyi and Liu, Burigede and Azizzadenesheli, Kamyar and Bhattacharya, Kaushik and Stuart, Andrew and Anandkumar, Anima},
  title   = {Neural Operator: Learning Maps Between Function Spaces With Applications to Partial Differential Equations},
  journal = {Journal of Machine Learning Research},
  volume  = {24},
  number  = {89},
  pages   = {1--97},
  year    = {2023}
}

@article{li2023pino,
  author  = {Li, Zongyi and Zheng, Hongkai and Kovachki, Nikola and Jin, David and Chen, Haoxuan and Liu, Burigede and Azizzadenesheli, Kamyar and Anandkumar, Anima},
  title   = {Physics-Informed Neural Operator for Learning Partial Differential Equations},
  journal = {ACM / IMS Journal of Data Science},
  volume  = {1},
  number  = {3},
  pages   = {1--27},
  year    = {2024}
}

@inproceedings{anandkumar2020gkn,
  author    = {Li, Zongyi and Kovachki, Nikola and Azizzadenesheli, Kamyar and Liu, Burigede and Bhattacharya, Kaushik and Stuart, Andrew and Anandkumar, Anima},
  title     = {Neural Operator: Graph Kernel Network for Partial Differential Equations},
  booktitle = {ICLR Workshop on Integration of Deep Neural Models and Differential Equations},
  year      = {2020},
  pages = {1--21}
}

@article{li2020mgno,
  author  = {Li, Zongyi and Kovachki, Nikola and Azizzadenesheli, Kamyar and Liu, Burigede and Bhattacharya, Kaushik and Stuart, Andrew and Anandkumar, Anima},
  title   = {Multipole Graph Neural Operator for Parametric Partial Differential Equations},
  journal = {arXiv preprint arXiv:2006.09535},
  year    = {2020}
}

@article{sarkar2025sp2gno,
  author  = {Sarkar, Subhankar and Chakraborty, Souvik},
  title   = {Spatio-spectral graph neural operator for solving computational mechanics problems on irregular domain and unstructured grid},
  journal = {Computer Methods in Applied Mechanics and Engineering},
  volume  = {435},
  pages   = {117--659},
  year    = {2025},
  doi     = {10.1016/j.cma.2024.117659}
}

@inproceedings{pfaff2021meshgraphnets,
  author    = {Pfaff, Tobias and Sanchez-Gonzalez, Alvaro and Battaglia, Peter},
  title     = {Learning Mesh-Based Simulation with Graph Networks},
  booktitle = {International Conference on Learning Representations},
  year      = {2021},
  pages = {1--18}
}

@inproceedings{brandstetter2022mpde,
  author    = {Brandstetter, Johannes and Worrall, Daniel E. and Welling, Max},
  title     = {Message Passing Neural {PDE} Solvers},
  booktitle = {International Conference on Learning Representations},
  year      = {2022},
  pages = {1--27}
}

@misc{velickovic2018gat,
  author        = {Veli{\v{c}}kovi{\'c}, Petar and Cucurull, Guillem and Casanova, Arantxa and Romero, Adriana and Li{\`o}, Pietro and Bengio, Yoshua},
  title         = {Graph Attention Networks},
  year          = {2018},
  eprint        = {1710.10903},
  archivePrefix = {arXiv},
  primaryClass  = {stat.ML}
}

@misc{rusch2023oversmoothing,
  author        = {Rusch, T. Konstantin and Bronstein, Michael M. and Mishra, Siddhartha},
  title         = {A Survey on Oversmoothing in Graph Neural Networks},
  year          = {2023},
  eprint        = {2303.10993},
  archivePrefix = {arXiv},
  primaryClass  = {cs.LG}
}

@inproceedings{horie2024fluxgnn,
  author    = {Horie, Masanobu and Mitsume, Naoto},
  title     = {Graph Neural {PDE} Solvers with Conservation and Similarity-Equivariance},
  booktitle = {International Conference on Machine Learning},
  pages     = {18785--18814},
  year      = {2024}
}

@inproceedings{liu2024clawno,
  author    = {Liu, Ning and Fan, Yiming and Zeng, Xianyi and Kl{\"o}wer, Milan and Zhang, Lu and Yu, Yue},
  title     = {Harnessing the Power of Neural Operators with Automatically Encoded Conservation Laws},
  booktitle = {International Conference on Machine Learning},
  year      = {2024},
  pages = {1--33}
}

@article{kim2024fluxfno,
  author  = {Kim, Taeyoung and Kang, Myungjoo},
  title   = {Approximating Numerical Fluxes Using Fourier Neural Operators for Hyperbolic Conservation Laws},
  journal = {arXiv preprint arXiv:2401.01783},
  year    = {2024}
}

@inproceedings{lichtle2026nfv,
  author    = {Lichtl{\'e}, Nathan and Canesse, Alexi and Fu, Zhe and Matin, Hossein and Delle Monache, Maria Laura and Bayen, Alexandre M.},
  title     = {{(U)NFV}: {(Un)}Supervised Neural Finite Volume Methods for Solving Hyperbolic {PDEs}},
  booktitle = {International Conference on Learning Representations (ICLR)},
  year      = {2026},
  pages = {1--23}
}

@article{patel2022tcpinn,
  author  = {Patel, Ravi G. and Manickam, Indu and Trask, Nathaniel A. and Wood, Mitchell A. and Lee, Myoungkyu and Tomas, Ignacio and Cyr, Eric C.},
  title   = {Thermodynamically consistent physics-informed neural networks for hyperbolic systems},
  journal = {Journal of Computational Physics},
  volume  = {449},
  pages   = {110--754},
  year    = {2022},
  doi     = {10.1016/j.jcp.2021.110754}
}

@article{wang2022pinnsfail,
  author  = {Wang, Sifan and Yu, Xinling and Perdikaris, Paris},
  title   = {When and why {PINNs} fail to train: A neural tangent kernel perspective},
  journal = {Journal of Computational Physics},
  volume  = {449},
  pages   = {110--768},
  year    = {2022},
  doi     = {10.1016/j.jcp.2021.110663}
}

@article{rahman2023uno,
  author  = {Rahman, Md Ashiqur and Ross, Zachary E. and Azizzadenesheli, Kamyar},
  title   = {{U-NO}: {U}-Shaped Neural Operators},
  journal = {Transactions on Machine Learning Research},
  year    = {2023},
  url     = {https://mlanthology.org/tmlr/2023/rahman2023tmlr-uno/}
}

@inproceedings{horie2022penn,
  author    = {Horie, Masanobu and Mitsume, Naoto},
  title     = {Physics-Embedded Neural Networks: Graph Neural {PDE} Solvers with Mixed Boundary Conditions},
  booktitle = {Advances in Neural Information Processing Systems},
  year      = {2022},
  pages = {1--12},
  url       = {https://openreview.net/forum?id=B3TOg-YCtzo}
}

@article{mishra2023generalization,
  author  = {Mishra, Siddhartha and Molinaro, Roberto},
  title   = {Estimates on the generalization error of physics-informed neural networks for approximating {PDEs}},
  journal = {IMA Journal of Numerical Analysis},
  volume  = {43},
  number  = {1},
  pages   = {1--43},
  year    = {2023},
  doi     = {10.1093/imanum/drab093}
}

@article{deryck2024wpinns,
  author  = {De Ryck, Tim and Mishra, Siddhartha and Molinaro, Roberto},
  title   = {w{PINNs}: Weak Physics Informed Neural Networks for Approximating Entropy Solutions of Hyperbolic Conservation Laws},
  journal = {SIAM Journal on Numerical Analysis},
  volume  = {62},
  number  = {2},
  pages   = {811--841},
  year    = {2024},
  doi     = {10.1137/22M1522504}
}

@article{baba2026hyperbolic,
  author  = {Baba, Zakaria and Bayen, Alexandre M. and Canesse, Alexi and Delle Monache, Maria Laura and Drieux, Martin and Fu, Zhe and Lichtl{\'e}, Nathan and Liu, Zihe and Matin, Hossein Nick Zinat and Piccoli, Benedetto},
  title   = {Supervised and Unsupervised Neural Network Solver for First Order Hyperbolic Nonlinear {PDEs}},
  journal = {arXiv preprint arXiv:2601.06388},
  year    = {2026},
  doi     = {10.48550/arXiv.2601.06388}
}

@article{sirignano2018dgm,
  author  = {Sirignano, Justin and Spiliopoulos, Konstantinos},
  title   = {{DGM}: A Deep Learning Algorithm for Solving Partial Differential Equations},
  journal = {Journal of Computational Physics},
  volume  = {375},
  pages   = {1339--1364},
  year    = {2018},
  doi     = {10.1016/j.jcp.2018.08.029}
}

@article{sundong2024lrnn,
  author  = {Sun, Jingbo and Dong, Suchuan and Wang, Fei},
  title   = {Local Randomized Neural Networks with Discontinuous {Galerkin} Methods for Partial Differential Equations},
  journal = {Journal of Computational and Applied Mathematics},
  volume  = {445},
  pages   = {115--830},
  year    = {2024},
  doi     = {10.1016/j.cam.2024.115830}
}

@article{mitusch2021hybridfemnn,
  author  = {Mitusch, Sebastian K. and Funke, Simon W. and Kuchta, Miroslav},
  title   = {Hybrid {FEM-NN} Models: Combining Artificial Neural Networks with the Finite Element Method},
  journal = {Journal of Computational Physics},
  volume  = {439},
  pages   = {110--651},
  year    = {2021},
  doi     = {10.1016/j.jcp.2021.110651}
}

@article{DBLP:journals/corr/HendrycksG16,
  author       = {Dan Hendrycks and
                  Kevin Gimpel},
  title        = {Bridging Nonlinearities and Stochastic Regularizers with Gaussian
                  Error Linear Units},
  journal      = {CoRR},
  volume       = {abs/1606.08415},
  year         = {2016},
  url          = {http://arxiv.org/abs/1606.08415},
  eprinttype   = {arXiv},
  eprint       = {1606.08415},
  bibsource    = {dblp computer science bibliography, https://dblp.org}
}

\end{document}